\documentclass[10pt,twocolumn]{article}

\usepackage[margin=0.85in,columnsep=0.24in]{geometry}
\usepackage[T1]{fontenc}
\usepackage[utf8]{inputenc}
\usepackage{microtype}
\usepackage{newtxtext,newtxmath}
\usepackage{abstract}
\usepackage{dblfloatfix}
\usepackage{placeins}   
\usepackage[font=small,labelfont=bf]{caption}

\usepackage{amsmath}
\usepackage{mathtools}
\usepackage{bm}
\usepackage{graphicx}
\usepackage{booktabs}
\usepackage{multirow}
\usepackage{array}
\usepackage{xcolor}
\usepackage{tikz}
\usetikzlibrary{arrows.meta,positioning,fit,backgrounds,shapes.geometric,calc}
\usepackage[authoryear,round]{natbib}
\usepackage[colorlinks=true,linkcolor=blue,citecolor=blue,urlcolor=blue]{hyperref}
\usepackage{xurl}   

\usepackage{colortbl}   
\definecolor{goodgreen}{HTML}{2ECC71}
\definecolor{badred}{HTML}{E74C3C}

\newtheorem{definition}{Definition}

\newcommand{\system}{OntoGraphRAG}

\title{When Confidence Takes the Wrong Path:\\
Diagnosing Retrieval-State Lock-In in RAG}
\author{\large Sahib Julka\\[0.25em]
\normalsize LMU University Hospital, LMU Munich\\
\normalsize Marchioninistr.\ 15, 81377 Munich}
\date{}

\begin{document}

\twocolumn[
\maketitle
\begin{onecolabstract}
The trustworthiness of a retrieval-augmented generation (RAG) system depends on
more than the answer it returns, yet many black-box uncertainty methods still
read agreement among sampled answers as confidence.
That inference fails when repeated samples condition on the same defective
retrieval state. The state may be empty, with the model falling back on
parametric memory, or populated by a coherent but wrong neighbourhood. In either
case, the answers agree because the error is stable. The problem is recognised
in deployed RAG, but it has lacked a name, a measurable signature, and a
prevalence bound.
We supply all three. We name the failure \emph{retrieval-state lock-in} and
diagnose it by separating the three objects a single confidence score
conflates: the answer surface, the retrieved evidence, and the retrieval state
itself. In an inspectable, ontology-guided knowledge-graph RAG (KG-RAG) system
across six question-answering snapshots, we measure the agreement blind spot
directly: at five samples per question, $42\%$ of KG-RAG
errors and $59\%$ of dense-retrieval errors carry zero answer dispersion, so
agreement has nothing to rank, while evidence- and retrieval-state checks
still flag most of them.
The decomposition supports an auditable decision rule: accepting an answer only
when answer, evidence, and retrieval checks all agree that it is low-risk reaches
$91.9\%$ pooled precision against a $69.7\%$ accept-all rate.
The cost is coverage: it certifies only $7.7\%$ of answers as low-risk. On the
clinical calibration domain it reaches $100\%$ precision under an automated
judge; this is an in-domain automated-label upper bound, not a clinical safety
claim, and still needs human validation.
Confidence in RAG is object-specific: when answers agree, the useful question is
which part of the pipeline to distrust.
\end{onecolabstract}
\vspace{1.0em}
]

\section{Introduction}
\label{sec:introduction}

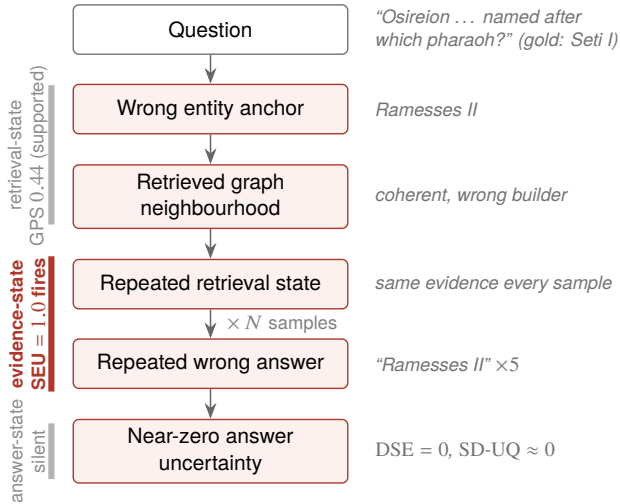
\begin{figure}[t]
\centering
\resizebox{\columnwidth}{!}{%
\begin{tikzpicture}[
  font=\footnotesize\sffamily,
  >={Stealth[length=2mm,width=1.5mm]},
  stage/.style={draw=black!55, line width=0.5pt, rounded corners=2pt,
                fill=white, minimum height=0.62cm, minimum width=3.45cm,
                align=center, inner xsep=5pt, inner ysep=3pt,
                font=\footnotesize\sffamily},
  bad/.style={stage, draw=badred!75!black, fill=badred!8},
  note/.style={font=\scriptsize\itshape\sffamily, text=black!55,
               align=left, anchor=west},
  raillab/.style={font=\scriptsize\sffamily, align=center, rotate=90},
  flow/.style={->, line width=0.55pt, black!60},
]
\node[stage] (q)    at (0,0)     {Question};
\node[bad]   (anc)  at (0,-1.0)  {Wrong entity anchor};
\node[bad]   (nbh)  at (0,-2.1)  {Retrieved graph\\neighbourhood};
\node[bad]   (ret)  at (0,-3.2)  {Repeated retrieval state};
\node[bad]   (ans)  at (0,-4.2)  {Repeated wrong answer};
\node[bad]   (unc)  at (0,-5.3)  {Near-zero answer\\uncertainty};

\draw[flow] (q) -- (anc);
\draw[flow] (anc) -- (nbh);
\draw[flow] (nbh) -- (ret);
\draw[flow] (ret) -- (ans)
  node[midway, right=2pt, font=\scriptsize\sffamily, text=black!50]
  {$\times\,N$ samples};
\draw[flow] (ans) -- (unc);

\draw[line width=2.2pt, black!30]
  (-2.00,-0.70) -- (-2.00,-2.45);
\node[raillab, text=black!45] at (-2.30,-1.58)
  {retrieval-state\\GPS $0.44$ (supported)};
\draw[line width=2.2pt, badred!75!black]
  (-2.00,-2.85) -- (-2.00,-4.55);
\node[raillab, text=badred!75!black, font=\scriptsize\bfseries\sffamily]
  at (-2.30,-3.70) {evidence-state\\SEU $=1.0$ fires};
\draw[line width=2.2pt, black!30]
  (-2.00,-4.95) -- (-2.00,-5.65);
\node[raillab, text=black!45] at (-2.30,-5.30)
  {answer-state\\silent};

\node[note] at (1.95,0)
  {``Osireion \ldots\ named after\\which pharaoh?'' (gold: Seti~I)};
\node[note] at (1.95,-1.0) {Ramesses~II};
\node[note] at (1.95,-2.1) {coherent, wrong builder};
\node[note] at (1.95,-3.2) {same evidence every sample};
\node[note] at (1.95,-4.2) {``Ramesses~II'' $\times 5$};
\node[note] at (1.95,-5.3)
  {$\mathrm{DSE}=0$, $\mathrm{SD\text{-}UQ}\approx 0$};
\end{tikzpicture}%
}
\caption{%
  \textbf{Worked presence-lock-in trace} (the HotpotQA Osireion case, also in
  Appendix Table~\ref{tab:lockin_gallery}).
  The retriever anchors to a strongly associated but wrong entity; the resulting
  neighbourhood is internally coherent, every sample receives the same evidence,
  and the repeated wrong answer carries near-zero answer-state (sampled-answer)
  uncertainty.
  The left rail shows where each diagnostic family attaches: answer-state
  agreement is silent (all samples agree), graph-path support (GPS) stays low
  because the wrong answer is still reachable in the graph, and only the
  evidence-state check fires (the retrieved passages contradict the answer).
  In the \emph{absence} variant (Section~\ref{sec:results_collapse}) the chain
  instead repeats an empty retrieval state, moving the warning to a
  retrieval-state abstention.%
}
\label{fig:lockin_trace}
\end{figure}

Ask a retrieval-augmented generation (RAG) system \citep{lewis2020rag} which
pharaoh the temple in front of the Osireion honours.
The gold answer is \emph{Seti~I}, but the retriever anchors to
\emph{Ramesses~II}, the pharaoh of the adjacent Abydos temple.
Sample the answer five times and it returns \emph{Ramesses~II} every time, with
zero disagreement and a coherent supporting graph trace
(cf.\ Figure~\ref{fig:lockin_trace}).
Agreement is perfect; the answer is wrong.
A black-box uncertainty estimator that reads agreement as confidence sees nothing
to flag. Only one signal fires: the retrieved passages contradict the confident
answer.

Confidently stable wrong answers are a recognised risk in deployed RAG, yet the
failure lacks a standard name, a measurable signature, and a prevalence
bound. This paper supplies all three.
Many black-box uncertainty estimators (semantic entropy
\citep{farquhar2024detecting,kuhn2023semantic}, SelfCheckGPT
\citep{manakul2023selfcheckgpt}, and their RAG-integrated variants
\citep{jiang2023active,su2024dragin,zimmerman2024twotieredrag}) ask whether
sampled answers agree, then read agreement as confidence.
That inference breaks down when retrieval keeps returning the same defective
state: the answer is stable for the wrong reason, and drawing more samples cannot
recover the missing signal.

\paragraph{Retrieval-state lock-in.}
We call this failure mode \textbf{retrieval-state lock-in}: the retrieval state
is degenerate and near-identical across repeated samples, so resampling cannot
surface the error.
Two variants matter. In \emph{absence} lock-in, retrieval repeatedly returns no
usable graph state, so the model answers from parametric memory. In
\emph{presence} lock-in (the Osireion case), retrieval repeatedly returns a
coherent but wrong graph neighbourhood. Both count as lock-in because the fixed
retrieval state, empty or wrong, is what defeats sampling.
We call the observable signature a \emph{silent error}: a wrong answer with zero
observed answer-state dispersion (all $N$ samples in one semantic cluster).
Silence is a signature, not proof: parametric overconfidence, benchmark
ambiguity, and answer normalisation can all leave the same footprint, so
silent-error rates are an upper bound on lock-in prevalence.
We then decompose those silent errors by retrieval-side mechanism, empty versus
demonstrably wrong route, to separate lock-in from look-alike confounders and
show the signature is not predominantly artefactual
(Section~\ref{sec:results_collapse}).

Lock-in is easier to diagnose in knowledge-graph-augmented RAG (KG-RAG), where a
symbolic retriever can repeatedly anchor to the same entity and traverse the
same relation path.
Knowledge graphs can improve multi-hop or provenance-sensitive tasks
\citep{edge2024graphrag,gutierrez2024hipporag,hu2025grag,shen2025gear,zhu2025kg2rag,wu2025medgraphrag},
though recent comparisons find graph retrieval task-dependent and often
complementary to dense retrieval
\citep{zhang2025ragvsgraphrag,xiang2025whengraphs,chen2025comparingraggraphrag,peng2024graphragsurvey}.
The symptom is not limited to graphs: any system that repeatedly returns a
concentrated evidence set can produce stable wrong answers, and the risk
compounds in agentic pipelines where successive queries reinforce a locked-in
state.
KG-RAG is useful here because its retrieval state is visible: matched entities,
triples, paths, and anchors make the mechanism observable.
Our experiments use \system{},\footnote{\url{https://github.com/julka01/OntoGraphRAG}}
an ontology-guided KG-RAG framework with entity-first routing, dense fallback,
and graph-state logging.

\paragraph{Three distinct uncertainty objects.}
We argue that RAG confidence has three distinct objects:
\begin{enumerate}
  \item \textbf{Answer-state uncertainty} measures variation across sampled
  answers.
  \item \textbf{Evidence-state uncertainty} measures whether the retrieved
  passages locally support the generated answer.
  \item \textbf{Retrieval-state uncertainty} measures whether the retrieval
  state itself supplies graph support.
\end{enumerate}
Each attaches to a different pipeline object and maps to a practical trust
question: calibrated confidence, faithfulness to evidence, and auditable
provenance.

\paragraph{Contributions.}
First, we \textbf{measure how prevalent the answer-state blind spot is} across
five QA families under deployable policies.\footnote{PubMedQA, RealMedQA,
HotpotQA, 2WikiMultiHopQA, and MuSiQue; HotpotQA contributes two snapshots
(bundle and FullWiki).}
At $N{=}5$ the silent-error footprint pools to $42\%$ of adaptive-KG and $59\%$
of dense errors ($8$--$55\%$ by dataset).

Second, we introduce a \textbf{diagnostic decomposition} separating
answer-state, evidence-state, and retrieval-state uncertainty.
Their representative scores are weakly correlated (pooled Spearman $\rho=0.03$
between answer- and evidence-state), so they carry complementary signal.

Third, we introduce \textbf{SD-UQ}, a low-cost question-conditioned
embedding-dispersion score within the answer-state family: a practical
convenience, competitive in these low-sample runs, rather than a central claim.

Fourth, we \textbf{show the decomposition is actionable}: a conjunctive rule
that certifies an answer only when answer-, evidence-, and retrieval-state
checks all agree that it is low-risk reaches $91.9\%$ pooled precision at
$7.7\%$ coverage ($81.6\%$
out-of-calibration), against a $69.7\%$ accept-all rate.

We compare dense RAG, adaptive KG-RAG, and a strict graph-only stress test
across clinical, biomedical, and open-domain multi-hop QA snapshots.
We do not claim KG-RAG is more accurate: it shows no statistically detectable
accuracy gap with dense retrieval (paired McNemar, all $p \geq 0.12$). Its value
here is to expose the failure states answer-only uncertainty cannot see.

\paragraph{Scope.}
The decomposition is graph-native: it reads \system{}'s symbolic retrieval
state to separate absence from presence, though the silent-error phenomenon
itself is general (and larger under dense retrieval).
The strict stress test changes several pipeline components at once, so it probes
a composite worst-case regime rather than isolating one causal variable.

\section{Related Work}
\label{sec:related_work}

\subsection{Answer-State Uncertainty for LLMs}

A large part of LLM uncertainty estimation stays in output space: if sampled
answers agree, the answer is treated as more likely to be correct.
Semantic entropy clusters sampled responses into meaning-equivalent groups and
computes entropy over the resulting distribution
\citep{kuhn2023semantic,farquhar2024detecting}; related methods approximate or
substitute for that idea by predicting it from hidden states
\citep{kossen2024sep}, measuring cross-sample contradiction
\citep{manakul2023selfcheckgpt}, eliciting verbalised confidence
\citep{kadavath2022language,tian2023justask,lin2022teaching}, or scoring
agreement across sampled reasoning paths
\citep{wang2023selfconsistency}.
White-box variants such as hidden-state probes \citep{kossen2024sep} and
induction-aware entropy gating \citep{bazarova2026intrygue} sit outside the
black-box setting studied here. They are orthogonal to the retrieval-state view:
where hidden states are available, they could be fused with these diagnostics
rather than treated as competitors.

A second line replaces hard semantic clustering with response-embedding
geometry: von~Neumann entropy on a similarity kernel
\citep{nikitin2024kernellanguageentropy}, concentration and dispersion in
embedding space \citep{qiu2024semanticdensity,li2025semanticvolume}, and
spectral decomposition of representation uncertainty
\citep{walha2025finegrained}.
For this paper they are the nearest answer-state baselines. Their focus,
however, is answer variation itself, not the case where retrieval stabilises
around an explicit graph trace.

\subsection{Uncertainty in RAG}

In RAG, uncertainty is asked to do several jobs. Adaptive retrieval methods use
confidence or entropy to decide when to retrieve
\citep{jiang2023active,su2024dragin,yao2025seakr,liu2025ctrla,zubkova2025sugar,moskvoretskii2025adaptive},
linking to selective prediction and abstention
\citep{geifman2017selective,tomani2024abstention}, conformal NLP
\citep{campos2024conformalnlp}, and trustworthiness work that treats reliability
as a pipeline property \citep{sun2024trustllm,ni2025trustworthyrag}.
Most of this work uses answer-side gates: sampled-answer uncertainty decides
whether retrieval, critique, or abstention is needed. That is precisely the
signal lock-in can silence. These methods sit naturally alongside the evidence-
and retrieval-state diagnostics studied here, rather than replacing them.
Conformal methods are adjacent for the same reason: they read coverage
from a single nonconformity score, whereas our conjunctive rule asks three
families to agree and targets auditable abstention rather than coverage
calibration.

Two neighbouring lines come closer to the evidence- and retrieval-state arms.
\citet{perezbeltrachini2025passageutility} predict evidence quality directly with
a lightweight passage-utility model that approximates sampling-based uncertainty
without sampling. Self-reflective methods (Self-RAG, corrective RAG
\citep{asai2024selfrag,yan2024crag}) let the model decide when to retrieve,
critique, or abstain through control tokens or fine-tuning, but need white-box
access. The present design instead stays black-box, a constraint that matters in
clinical and biomedical RAG, where recent reviews report heavy use of
proprietary models and unresolved evaluation and governance gaps
\citep{amugongo2025healthcareragreview}.

The closest neighbours sharpen the claim.
An axiomatic analysis shows that standard uncertainty estimators cannot reliably
assess correctness in RAG and proposes a calibration framework
\citep{soudani2025axiomaticrag}; this paper supplies the empirical counterpart,
locating where answer-state agreement collapses and which retrieval- and
evidence-state signals still respond.
FRANQ separates factuality from faithfulness to retrieved context, but operates
on flat-text RAG and does not expose three-way graph-side decomposition
\citep{fadeeva2025franq}.
SURE-RAG scores evidence-set sufficiency for selective answering
\citep{qiu2026surerag}, posing the evidence-state question effectively but not
assessing whether retrieval has concentrated around a wrong symbolic trace.
SURE-RAG and FRANQ are better read as more specialised evidence-state diagnostics
for flat-text RAG. We use SEU as a lightweight black-box representative, leaving
substitution with learned sufficiency models to future work.
R2C perturbs the retrieval--reasoning loop so that dispersion reflects both
retriever and generator uncertainty \citep{soudani2025uqretrievalreasoning}; it
is complementary, measuring movement under perturbation, while lock-in concerns
unperturbed retrieval that barely moves at all.

Broader retrieval-stability and context-selection work makes the same background
point: retrieval composition matters for reliability. Long-context and
context-balancing
methods ask which spans to keep \citep{li2024uncertaintyrag,voloshyn2026lrag},
noise-aware calibration studies how irrelevant context affects confidence
\citep{liu2026nacc}, and retriever benchmarks study how much information the
retrieved set carries \citep{zheng2026revisitingrag}.
Our claim is narrower: when repeated samples see the same defective state,
answer variance can collapse even if the generator is sampled.
On the graph side, Ca2KG studies overconfidence in KG-RAG through counterfactual
prompting, but it does not report a structured multi-family diagnostic that
separates retrieval-state from evidence-state uncertainty. Nor does it provide a
prevalence estimate, a route-level absence/presence decomposition, or an audit rule
\citep{ren2026ca2kg}, precisely the diagnostics this paper adds.
BRINK shows that KG-RAG systems can fall back on parametric memory when graphs
are incomplete \citep{zhou2025brink}.
Our \emph{absence} variant largely operationalises that phenomenon at retrieval
time: the strict-clinical silent errors are empty-retrieval rows answered from
parametric memory (Section~\ref{sec:results_collapse}). The additional case is
\emph{presence} lock-in, where the system retrieves a populated but
wrong-and-coherent neighbourhood. The three-family framing then separates the two
empirically. Table~\ref{tab:positioning} places this contribution against the
five closest neighbours; it should be read compositionally, not as a contest.

\begin{table*}[t]
\centering
\small
\setlength{\tabcolsep}{2.8pt}
\caption{%
  \textbf{Positioning against the closest related work.}
  \checkmark{}: direct object of study; $\sim$: partial or indirect; --: not
  addressed.
  Among these five recent neighbours, only the present study jointly inspects
  the answer, evidence, and retrieval state (the three left columns).
  The two right columns, set off by the rule, are context rather than comparison
  axes: \emph{Trace exposed} reflects the inspectable KG-RAG substrate, and
  \emph{Lock-in studied} is definitional, since the concept is introduced here.%
}
\label{tab:positioning}
\begin{tabular}{@{}lccc|cc@{}}
\toprule
& Answer & Evidence & Retrieval & Trace & Lock-in \\
& state & state & state & exposed & studied \\
\midrule
FRANQ~\citep{fadeeva2025franq} & \checkmark & \checkmark & -- & -- & -- \\
SURE-RAG~\citep{qiu2026surerag} & $\sim$ & \checkmark & $\sim$ & -- & -- \\
R2C~\citep{soudani2025uqretrievalreasoning} & \checkmark & $\sim$ & $\sim$ & -- & -- \\
Ca2KG~\citep{ren2026ca2kg} & \checkmark & $\sim$ & $\sim$ & $\sim$ & -- \\
BRINK~\citep{zhou2025brink} & -- & -- & $\sim$ & $\sim$ & -- \\
This paper & \checkmark & \checkmark & \checkmark & \checkmark & \checkmark \\
\bottomrule
\end{tabular}
\end{table*}
Three further literatures set the remaining boundary conditions.
First, RAG evaluation frameworks such as RAGAs and ARES already separate context
relevance and faithfulness from answer quality
\citep{es2024ragas,saadfalcon2024ares}; their context-relevance/faithfulness
split motivates the analogous evidence-state check here (SEU,
Table~\ref{tab:metric_taxonomy}). We tie that separation to explicit graph
retrieval state, where answer agreement and evidence support can decouple.
Second, external selective-RAG systems, including passage-utility predictors,
FRANQ, SURE-RAG, R2C, and ARES-style frameworks, target flat-text RAG, so a like-for-like
comparison would mean reimplementing them inside the KG-RAG pipeline; because the
silent-error phenomenon is larger under dense retrieval
(Section~\ref{sec:results_silent}), our own dense-RAG runs are the natural substrate
where such a baseline could be compared.
Third, knowledge-conflict work models the tension between retrieved evidence and
the model prior through source-aware synthesis or KG-mediated reconciliation
\citep{wu2024clasheval,wang2025astuterag,zhang2025faithfulrag,liu2025truthfulrag}
and through attribution and citation-faithfulness checks that catch unsupported
LLM references, including in medicine
\citep{wallat2025faithfulness,li2024citationenhanced,wu2024medicalreferenceeval}.
The present study asks which of these signals stay useful when retrieval is
visible and partially stabilised.

\subsection{KG-RAG and Graph-Side Uncertainty}

Graph-augmented retrieval has become a prominent alternative to flat-text RAG.
Systems such as RoG, Think-on-Graph, HippoRAG, SubgraphRAG, GraphRAG, StructGPT,
G-Retriever, and GNN-RAG use graphs to make multi-hop reasoning more explicit and
auditable
\citep{luo2024rog,sun2024tog,ma2025tog2,gutierrez2024hipporag,
li2025subgraphrag,edge2024graphrag,jiang2023structgpt,he2024gretriever,
mavromatis2025gnnrag}, with benefits concentrated on multi-hop retrieval, bridge
preservation, networked evidence, or stronger provenance
\citep{edge2024graphrag,gutierrez2024hipporag,shen2025gear,
hu2025grag,zhu2025kg2rag,wu2025medgraphrag}; topology-aware retrieval adds
selection by graph proximity and structural role rather than similarity alone
\citep{wang2024taar}.
This study treats the graph as a source-linked summary of passages, where
triples, paths, and relation anchors trace back to the text that licensed them.
Comparative work correspondingly finds graph and dense retrievers often
complementary rather than globally ordered
\citep{zhang2025ragvsgraphrag,xiang2025whengraphs,peng2024graphragsurvey,
chen2025comparingraggraphrag}.

Less developed is the graph-side confidence question: whether the retrieved
graph state itself, including matched entities, traversal paths, and relation
anchors, supports the generated answer. That question is different from checking
answer variation or text-evidence consistency.
Ca2KG \citep{ren2026ca2kg} and BRINK \citep{zhou2025brink}, both discussed above,
are the nearest graph-side efforts, but neither does this, while a separate
literature models uncertainty inside knowledge graphs themselves through
confidence-weighted triples or distribution shift in KG embeddings
\citep{takahashi2025uncertaintyawarekg,lee2025decomposinguncertaintykge,zhu2025certainty}.
Once a hybrid KG-RAG system exposes all three objects, the diagnostic problem is
to score answer variation, evidence support, and graph support without pretending
they are the same signal.

\section{Background and Problem Formulation}
\label{sec:background}

\subsection{Sampling-Based Uncertainty in RAG}
\label{sec:uq_background}

A RAG model does not draw answers in a vacuum: each sample is conditioned on
whatever context the retriever returns. When retrieval keeps returning the same
context, resampling explores decoder variation but not evidence variation, so
agreement reports decoder stability, not correctness.

Let $q$ be a question, $c$ a retrieved context bundle, and $r$ a model response.
Sampling-based uncertainty estimators operate on repeated draws from
$p(r \mid q)$. In RAG, the response distribution is mediated by retrieval:
\begin{equation}
  p(r \mid q)
  = \sum_{c \in \mathcal{C}(q)} p(r \mid q, c)\, p(c \mid q),
  \label{eq:marginal}
\end{equation}
where $\mathcal{C}(q)$ is the finite set of candidate context bundles.
$p(c \mid q)$ can be sharply peaked (near-degenerate, the lock-in regime) or
spread across several bundles.

\subsection{Why KG-RAG Changes the Interpretation}
\label{sec:determinism}

KG-RAG changes what disagreement, and agreement, can mean.
Vanilla RAG returns a ranked list of passages. KG-RAG returns a structured state:
anchors, relation labels, triples, and paths whose names and types can be
inspected rather than inferred after the fact.
In the entity-first regime, retrieval is mediated by entity matching and graph
expansion over a fixed graph $\mathcal{G}$.
When that process is highly stabilised, one context dominates:
\begin{equation}
  c^\star = \arg\max_{c \in \mathcal{C}(q)} p(c \mid q).
\end{equation}
As the residual mass $1 - p(c^\star \mid q)$ shrinks, the marginal collapses:
\begin{equation}
  p(r \mid q) \approx p(r \mid q, c^\star),
  \label{eq:collapsed}
\end{equation}
so repeated samples mostly expose decoder variability under fixed evidence
(Appendix~\ref{sec:appendix_residual_bound} gives the total-variation bound).
This is the regime of interest, and also the dangerous one. Once retrieval has
stabilised, low answer variance can mean good evidence, but it can also mean the
same retrieval-side mistake has been repeated on every sample.
In the deployed policy, iterative decomposition and dense fallback keep the
residual from fully collapsing, so Equation~\eqref{eq:collapsed} is a stylised
diagnostic lens rather than a full description of every call.
Write $s$ for the graph retrieval state: the matched entities, paths, and
anchors behind a retrieved context.
The empirical counterpart is read from the saved route logs: a silent error
(Definition~\ref{def:silent}) is classified by its retrieval route as
absence-compatible (an empty route) or presence-compatible (a populated route
that does not reach the gold answer entity), per Definition~\ref{def:lockin}.

\begin{definition}[Silent error]
\label{def:silent}
A wrong answer with zero observed answer-state dispersion under the sampling
budget: all $N$ samples fall in one semantic cluster (DSE $=0$) and the
embedding-dispersion score SD-UQ is at its floor (both answer-dispersion scores
are defined below, Section~\ref{sec:gen_metrics}).
The floor is implementation-defined (encoder- and $\varepsilon$-dependent), so
silent-error rates are reported as an upper bound.
\end{definition}
\begin{definition}[Retrieval-state lock-in]
\label{def:lockin}
A silent error for which the repeated samples condition on the same
\emph{defective} retrieval state $s$, either empty (absence) or a coherent but
wrong neighbourhood (presence).
``Defective'' is operationalised from route logs as an empty route, or a
populated route whose neighbourhood does not contain the gold answer entity
where checkable; rows with missing route metadata are reported separately.
\end{definition}
Both variants are retrieval-state defects. This matters because parametric
overconfidence \emph{despite} adequate retrieval can look similar at the answer
surface, but it is not lock-in.

\subsection{A Three-Family Analysis Frame for KG-RAG}
\label{sec:decomposition}


\begin{table}[t]
\centering
\small
\setlength{\tabcolsep}{5pt}
\renewcommand{\arraystretch}{1.35}
\caption{%
  \textbf{Uncertainty as an audit of three pipeline objects, not one scalar.}
  Answer-state metrics see only the sampled answers; evidence-state metrics
  compare the answer against the retrieved passages; retrieval-state diagnostics
  inspect the graph trace itself (anchors, triples, paths, and abstentions).
  The paper reports them separately because they answer different questions.
  This table defines the objects; Table~\ref{tab:failure_modes} states how each
  behaves across retrieval regimes, and Table~\ref{tab:failure_taxonomy} maps
  their joint outcomes into a $2{\times}2$ failure taxonomy.%
}
\label{tab:decomposition}
\begin{tabular}{@{}p{0.19\linewidth}p{0.27\linewidth}p{0.42\linewidth}@{}}
\toprule
\textbf{Observed object} & \textbf{Signal family} & \textbf{Diagnostic question} \\
\midrule
Answer samples
&
Answer-state uncertainty
&
Do repeated generations disagree, or has the answer surface become
artificially stable? \\
\addlinespace[0.25em]
Retrieved passages
&
Evidence-state uncertainty
&
Does the retrieved text locally entail, contradict, or fail to support the
generated answer? \\
\addlinespace[0.25em]
Graph trace
&
Retrieval-state uncertainty
&
Which anchors, typed triples, and paths made the answer reachable, and where
does the graph abstain? \\
\bottomrule
\end{tabular}

\vspace{0.45em}
\begin{minipage}{0.96\linewidth}
\footnotesize
\textit{The three families are not independent predictors of correctness.}
They are three logged views of the same KG-RAG run: generated text, retrieved
evidence, and symbolic retrieval state.
\end{minipage}
\end{table}

We use three families because a KG-RAG run exposes three different objects: the
answer, the assembled evidence, and the retrieval state
(Table~\ref{tab:decomposition}). Recall the graph retrieval state $s$; let $c$
be the textual evidence and $r$ the answer. The pipeline factorises:
\begin{equation}
  p(r, c, s \mid q) = p(s \mid q)\, p(c \mid q, s)\, p(r \mid q, c, s).
\end{equation}
Answer-state uncertainty probes variability in $p(r \mid q, c, s)$;
evidence-state measures ask whether $c$ is locally consistent with $r$;
retrieval-state measures ask whether $s$ supplies coherent graph support.
A system can produce low-variance answers from the wrong subgraph, retrieve a
plausible path whose passages do not entail the answer, or retrieve good
evidence while the answer remains unstable.

\section{System Description}
\label{sec:system}

The experimental platform is \system{}, an open-source framework that exposes
both vanilla RAG and KG-RAG under a shared interface.
The two systems use the same corpora, embeddings, and generation model; in the
adaptive dense-vs-KG comparison they differ only in retrieval and evidence
organisation.

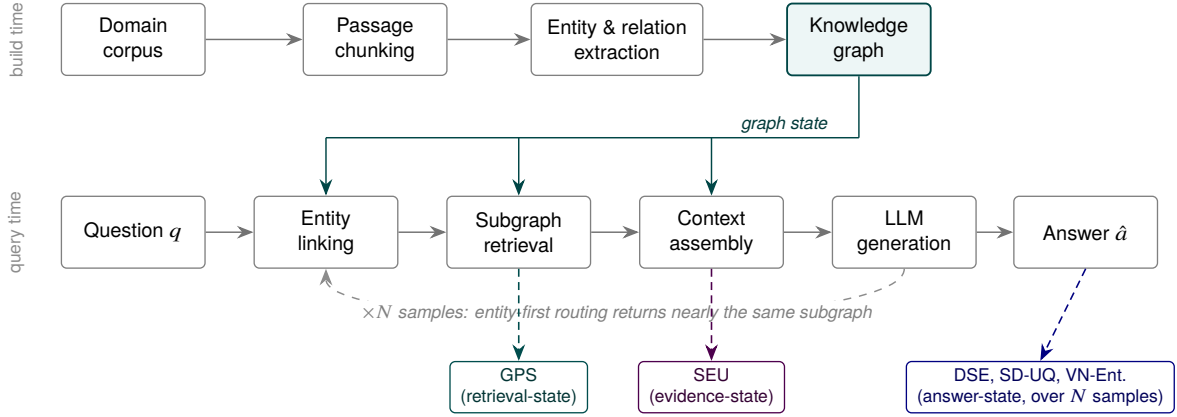
\begin{figure*}[t]
\centering
\begin{tikzpicture}[
  font=\footnotesize\sffamily,
  >={Stealth[length=2.2mm,width=1.7mm]},
  stage/.style={draw=black!50, line width=0.5pt, rounded corners=2pt,
                fill=white, minimum height=0.95cm, minimum width=1.9cm,
                align=center, inner xsep=6pt, inner ysep=4pt},
  kgstage/.style={stage, draw=teal!55!black, line width=0.7pt, fill=teal!7},
  flow/.style={->, line width=0.55pt, black!55},
  rail/.style={teal!55!black, line width=0.55pt},
  rowlab/.style={font=\scriptsize\sffamily, text=black!45},
]

\node[stage]   (corpus)  at (0,2.55)   {Domain\\corpus};
\node[stage]   (chunk)   at (3.2,2.55) {Passage\\chunking};
\node[stage]   (extract) at (6.4,2.55) {Entity \& relation\\extraction};
\node[kgstage] (kg)      at (9.6,2.55) {Knowledge\\graph};

\node[stage] (q)    at (0,0)     {Question $q$};
\node[stage] (link) at (2.55,0)  {Entity\\linking};
\node[stage] (sub)  at (5.1,0)   {Subgraph\\retrieval};
\node[stage] (ctx)  at (7.65,0)  {Context\\assembly};
\node[stage] (llm)  at (10.2,0)  {LLM\\generation};
\node[stage] (ans)  at (12.6,0)  {Answer $\hat{a}$};

\node[rowlab, rotate=90, anchor=center] at (-1.55,2.55) {build time};
\node[rowlab, rotate=90, anchor=center] at (-1.55,0)    {query time};

\draw[flow] (corpus) -- (chunk);
\draw[flow] (chunk)  -- (extract);
\draw[flow] (extract) -- (kg);
\draw[flow] (q)    -- (link);
\draw[flow] (link) -- (sub);
\draw[flow] (sub)  -- (ctx);
\draw[flow] (ctx)  -- (llm);
\draw[flow] (llm)  -- (ans);

\coordinate (raily) at (0,1.25);
\draw[rail] (kg.south) -- (kg.south |- raily);
\draw[rail] (link.north |- raily) -- (kg.south |- raily);
\draw[->, rail] (link.north |- raily) -- (link.north);
\draw[->, rail] (sub.north  |- raily) -- (sub.north);
\draw[->, rail] (ctx.north  |- raily) -- (ctx.north);
\node[font=\scriptsize\itshape\sffamily, text=teal!45!black,
      above, inner sep=1.5pt] at (8.62,1.25) {graph state};

\draw[dashed, line width=0.55pt, black!45, ->]
  (llm.south) .. controls +(0,-0.85) and +(0,-0.85) .. (link.south);
\node[font=\scriptsize\itshape\sffamily, text=black!50, fill=white,
      inner sep=1.5pt] at (6.4,-1.07)
  {$\times N$ samples: entity-first routing returns nearly the same subgraph};

\tikzset{tap/.style={draw, line width=0.5pt, rounded corners=2pt,
                     align=center, font=\scriptsize\sffamily,
                     inner xsep=4pt, inner ysep=2.5pt, fill=white},
         probe/.style={densely dashed, line width=0.6pt, ->}}
\node[tap, draw=teal!60!black, text=teal!40!black] (tgps) at (5.1,-2.05)
  {GPS\\(retrieval-state)};
\node[tap, draw=violet!60!black, text=violet!40!black] (tseu) at (7.65,-2.05)
  {SEU\\(evidence-state)};
\node[tap, draw=blue!50!black, text=blue!35!black] (tans) at (12.0,-2.05)
  {DSE, SD-UQ, VN-Ent.\\(answer-state, over $N$ samples)};
\draw[probe, teal!60!black]   (sub.south) -- (tgps.north);
\draw[probe, violet!60!black] (ctx.south) -- (tseu.north);
\draw[probe, blue!50!black]   (ans.south) -- (tans.north);

\end{tikzpicture}
\caption{%
  \textbf{\system{} pipeline with the three measurement taps.}
  \emph{Build time} (top): a domain corpus is chunked and converted via
  ontology-guided extraction into a typed knowledge graph with chunk-level
  provenance.
  \emph{Query time} (bottom): the question is linked to seed entities, and the
  graph state conditions retrieval at three points (entity linking, subgraph
  traversal, and context assembly) before the LLM produces the final answer.
	  The dashed probes mark where each diagnostic family attaches: GPS scores
	  graph support in risk orientation, SEU scores the assembled evidence against
	  the answer, and the answer-state estimators see only the $N$ sampled answers.
	  When repeated samples re-enter retrieval but entity-first routing returns
	  nearly the same subgraph, answer-state uncertainty mainly reads decoder
	  variation under almost-fixed evidence (Section~\ref{sec:metrics}).%
}
\label{fig:pipeline}
\end{figure*}

\subsection{\system{} Overview}
\label{sec:ontographrag_overview}

\system{} uses graph retrieval to add structure without discarding dense
evidence (Figure~\ref{fig:pipeline}).
The graph turns retrieval commitments into visible objects: recognised entities,
accepted relations, triples, paths, and the passages that licence them.
During KG construction, passages are converted into entity and relation nodes
with ontology-guided typing, anchor grounding, confidence scores, provenance
links, and contradiction flags when a schema is available.
During retrieval, the system routes the question through matched entities and
local neighbourhoods, falling back to dense retrieval and retriever-first
expansion when the anchor is weak.
The resulting context contains both textual passages and explicit graph paths.
That is what makes the three-level audit possible: answer state, evidence state,
and retrieval state.
\system{} logs anchors, paths, and supporting passages with the final answer, so
the provenance trace is part of the retrieval output rather than a private
prompt-construction detail.

\subsection{Vanilla RAG}
\label{sec:vanilla_rag}

Vanilla RAG performs direct vector retrieval over chunk embeddings.
For each question, it retrieves the top-$k$ chunks above a similarity
threshold and optionally appends adjacent chunks to capture answers split
across chunk boundaries.
The LLM receives flat text only. There are no explicit entities, relations, or
multi-hop paths, and hence no graph-side object on which to define
retrieval-state diagnostics.

\subsection{KG-RAG}
\label{sec:kg_rag}

KG-RAG is a routed hybrid pipeline rather than a single graph walk.

\paragraph{Stage 1: seed selection.}
The preferred path is entity-first anchoring: the system identifies seed
entities with symbolic matching and embedding lookup over short mentions.
If no reliable anchor is found, the system does not immediately reduce to
vector-only retrieval; instead it can route to the retriever-first graph pass of
Stage 3, bypassing the entity-first expansion of Stage 2.

\paragraph{Stage 2: local graph expansion and scoring.}
Starting from the seed entities, the system traverses the graph up to a
dataset-specific hop limit, producing an explicit retrieval state of linked
chunks, entities, relations, and readable traversal paths.
Traversal is provenance-aware: on question-bundle datasets, paths and edges are
restricted to the current question's local evidence rather than allowed to
borrow support from other questions in the same dataset graph.
Chunks are scored by hop distance and local diffusion over the retrieved
entity subgraph.

\paragraph{Stage 3 (fallback): retriever-first graph expansion when anchoring is weak.}
When entity-first retrieval is weak or empty, the system first retrieves dense
text chunks, extracts their linked entities, expands the graph from those
passage-derived seeds, and re-scores the resulting context with a combined
vector-plus-graph signal.

\paragraph{Stage 4: evidence organisation.}
The final prompt is organised rather than simply concatenated.
Retrieved graph paths are grouped with overlapping supporting passages into
explicit reasoning chains, with remaining passages listed separately as
additional evidence.
The generator sees both structured multi-hop support and the local text that
grounds each chain.

\paragraph{Diagnostic role.}
\system{} makes the retrieval state explicit rather than tacit. The graph is not
an oracle: its triples and paths may be helpful or wrong. Their value here is
that they can be inspected. Logged anchors, paths, and routes turn retrieval from
an internal variable into a queryable object, richer for diagnosis than a list of
nearest-neighbour chunks alone.

\paragraph{Retrieval stabilisation.}
The diagnostic problem appears when the retrieval state is stable across
repeated calls.
Retriever-first expansion, vector fallback, and iterative decomposition can
weaken that stabilisation when the graph anchor is brittle, so the experiments
compare diagnostics across retrieval regimes.
Entity-first retrieval can be especially prone to stabilising: once the system
commits to an anchor entity, the subsequent graph expansion is largely
determined, so the same neighbourhood, right or wrong, tends to recur across
samples.

\section{Uncertainty Measures}
\label{sec:metrics}

\begin{table}[t]
\centering
\footnotesize
\caption{%
  \textbf{Quick reference for the four scores used in the results.}
  All are oriented so that \emph{higher means more risk}: a high value flags a
  potentially untrustworthy answer, and a low score is low-risk and passes the
  audit gate.
  Full definitions are below; this box is a reading aid for
  Sections~\ref{sec:results} onward.%
}
\label{tab:metric_glossary}
\setlength{\tabcolsep}{4pt}
\begin{tabular}{@{}llp{3.5cm}@{}}
\toprule
Score & Object & Plain meaning (higher $=$ riskier) \\
\midrule
DSE   & answer    & sampled answers disagree \\
SD-UQ & answer    & sampled answers are dispersed in embedding space \\
SEU   & evidence  & retrieved text contradicts the answer \\
GPS   & retrieval & answer entity is weakly supported in the graph \\
\bottomrule
\end{tabular}
\end{table}

Table~\ref{tab:metric_glossary} orients the four headline scores used in the
results. The full suite contains eight diagnostic measures, but the organising
unit is not the measure; it is the object observed in a QA run: the sampled
answer surface, the retrieved evidence, or the retrieval state
(Table~\ref{tab:metric_taxonomy}).
The eight measures should therefore not be read as eight independent estimates
of one hidden confidence variable.
Some dependencies are built in: the P(True)-style proxy is monotone in discrete
semantic entropy (DSE) under black-box sampling, and VN-Entropy and SD-UQ are
both geometric statistics of the same response matrix.
For that reason, the headline tables use one representative per object:
SD-UQ for answer state (introduced here), SEU for evidence state, and GPS for
retrieval state.
DSE, P(True), SelfCheckGPT, SRE-UQ, and VN-Entropy remain useful answer-side
controls and ablations (Table~\ref{tab:metric_taxonomy});
Section~\ref{sec:results} specifies which appears in each table.
The answer-state scores mostly come from standalone LLM uncertainty estimation
\citep{kuhn2023semantic,manakul2023selfcheckgpt,kadavath2022language,moskvoretskii2025adaptive}.
Let $\bm{r} = \{r_1,\ldots,r_N\}$ denote $N$ responses sampled from the
LLM for a question $q$ with retrieved context $c$.
Let $\bm{v}_i \in \mathbb{R}^d$ denote the $\ell_2$-normalised embedding of
$r_i$, and $\bm{V} = [\bm{v}_1\mid\cdots\mid\bm{v}_N]^\top \in \mathbb{R}^{N\times d}$
the stacked response embedding matrix.
All estimators return a non-negative scalar: GPS and SEU are bounded in
$[0,1]$, DSE and VN-Entropy have maximum $\log N$, and SRE-UQ and SD-UQ are
unbounded.

\begin{table*}[t]
\centering
\small
\caption{%
  \textbf{Taxonomy of the eight uncertainty measures.}
  The first six rows are answer-state estimators; the last two are the
  evidence-state (SEU) and retrieval-state (GPS) diagnostics.
  Measured per-question costs are reported in Appendix
  Table~\ref{tab:compute_cost}.
  $\dagger$: metric or KG-RAG operationalisation defined here.
  $\ddagger$: prior estimator first benchmarked here for KG-RAG.%
}
\label{tab:metric_taxonomy}
\begin{tabular}{@{}lllp{4.5cm}p{2.5cm}p{2.5cm}@{}}
\toprule
\textbf{Object family} & \textbf{Subtype} & \textbf{Measure} & \textbf{Formal sketch} & \textbf{Required inputs} & \textbf{Cost at inference} \\
\midrule
\multirow{6}{*}{Answer-state}
  & Entropy & DSE
  & $-\sum_k \hat{p}_k \log \hat{p}_k$, with $\hat{p}_k = |C_k|/N$
  & sampled answers, NLI
  & $N$ samples; $O(N^2)$ NLI calls \\
  & Calibration & $\mathrm{P(True)}$
  & $1 - |\{i:\mathrm{cl}(r_i)=\mathrm{cl}(r_1)\}|/N$; black-box P(True)-style proxy
  & sampled answers, NLI
  & $N$ samples; shares DSE clustering \\
  & Similarity & SelfCheckGPT
  & $\frac{1}{|\mathcal{P}|}\sum_{(i,j)\in\mathcal{P}}
      [\mathrm{NLI}(r_i,r_j)=\textsc{contradiction}]$
  & sampled answers, NLI
  & $N$ samples; $O(N^2)$ NLI calls \\
  & Perturbation & SRE-UQ
  & $\frac{1}{M}\sum_{i\in\mathcal{T}}|\Delta_i|$; perturbation sensitivity of the kernel mean embedding
  & sampled answer embeddings
  & $N$ samples; embeddings only \\
  & Geometric & VN-Entropy$^{\ddagger}$
  & $-\operatorname{tr}(\bm{\rho}\log\bm{\rho})$, with $\bm{\rho}=\bm{V}\bm{V}^\top/N$
  & sampled answer embeddings
  & $N$ samples; embeddings only \\
  & Geometric & SD-UQ$^\dagger$
  & $\exp\!\bigl(\frac{1}{k}\sum_{i=1}^k \log(\eta_i+\varepsilon)\bigr)$ on question-orthogonal residuals
  & question embedding, sampled answer embeddings
  & $N$ samples; embeddings only \\
\midrule
Evidence-state & NLI support & SEU$^\dagger$
  & $\frac{1 - (n_E-n_C)/K}{2}$; entailment--contradiction deficit over retrieved chunks
  & retrieved chunks, answer, NLI
  & 1 sample; $K$ NLI calls \\
\midrule
Retrieval-state & graph support & GPS$^\dagger$
  & $\displaystyle 1 -
     \frac{\sum_{e:\mathrm{reach}_e} w_e \gamma^{|L_e-\hat L(q)|}}
          {\sum_e w_e}$; soft-linked, depth-matched answer-entity support
  & retrieved KG, entity embeddings, linked question/answer entities
  & 1 sample; per-entity graph queries, no LLM calls \\
\bottomrule
\end{tabular}
\end{table*}

\subsection{Answer-State Uncertainty Estimators}
\label{sec:gen_metrics}

\paragraph{Borrowed answer-state estimators.}
The five additional answer-state controls are standard estimators, used here
without modification; their formal definitions are deferred to
Appendix~\ref{sec:appendix_answer_metrics}.
In brief, DSE is the count-weighted black-box semantic entropy of
\citet{kuhn2023semantic}; the P(True) proxy \citep{kadavath2022language} is risk-scored as one
minus the cluster-agreement fraction, monotone in DSE under black-box sampling;
SelfCheckGPT \citep{manakul2023selfcheckgpt} is the NLI-contradiction rate over
sampled response pairs; SRE-UQ \citep{vipulanandan2026sreuq} is a
perturbation-sensitivity statistic of the response-embedding distribution; and
VN-Entropy \citep{nikitin2024kernellanguageentropy} is the von~Neumann entropy
of the normalised response-embedding Gram matrix.
Only SD-UQ, the answer-state score introduced here, is defined in full below.

\paragraph{SD-UQ (introduced here).}
SD-UQ is a question-conditioned embedding-dispersion statistic, distinct from
\citet{qiu2024semanticdensity}'s Semantic Density.
It is low when sampled answers point in nearly the same direction, which is the
signature of a stabilised (and possibly locked-in) answer surface.
Two design choices separate it from VN-Entropy, the other geometric score.
First, SD-UQ projects out the question direction, so answers that merely restate
the question register as low-dispersion.
Second, it summarises the residual spectrum by the geometric mean of the
top-$k$ singular values, rather than by the von~Neumann entropy of the full Gram
matrix, making collapse onto a single residual mode easier to see.
Given the unit-norm question embedding $\hat{\bm{q}}\in\mathbb{R}^d$, define
the orthogonal projector
\begin{equation}
  \bm{P}_{q}^{\perp}
  = \bm{I} - \hat{\bm{q}}\hat{\bm{q}}^\top.
\end{equation}
The projected response matrix is
\begin{equation}
  \bm{V}_{\perp} = \bm{V}\bm{P}_{q}^{\perp}.
\end{equation}
The thin SVD of the centred projected matrix is
\begin{equation}
  \frac{\bm{H}\bm{V}_{\perp}}{\sqrt{N}}
  = \bm{U}\bm{\Sigma}\bm{W}^\top,
\end{equation}
where $\bm{H} = \bm{I} - \frac{1}{N}\mathbf{1}\mathbf{1}^\top$ is the
row-centering matrix.
Let $\eta_1 \geq \cdots \geq \eta_k$ be the top-$k$ singular values on
the diagonal of $\bm{\Sigma}$.
\begin{equation}
  \mathrm{SD\text{-}UQ}(\bm{r})
  \;=\;
  \exp\!\left(\frac{1}{k}\sum_{i=1}^{k} \log(\eta_i + \varepsilon)\right),
  \label{eq:sd_uq}
\end{equation}
where $k = \min(N{-}1,\, k_{\max})$ with $k_{\max}=8$ (so $k=4$ at the reported
$N=5$ budget) and $\varepsilon = 10^{-12}$ is a numerical stability constant.
Because all embeddings are $\ell_2$-normalised and the encoder and sampling
budget are fixed, SD-UQ is used as an operationally comparable within-run ranking
score rather than an encoder-invariant uncertainty estimate; at the reported
$N=5$ budget only $k=4$ residual modes enter the geometric mean, so it is a
low-resolution ranking diagnostic, not a high-resolution distribution estimate.
SD-UQ needs only question and answer embeddings; this keeps its marginal cost
low.

\subsection{Retrieval-State Support Diagnostics}
\label{sec:struct_metrics}

The retrieval-state family (Section~\ref{sec:introduction}) asks whether the
retrieved graph contains a usable path from the question entities to the
asserted answer entity.
It matters when $p(c \mid q) \approx \delta_{c^*}$
(Section~\ref{sec:determinism}): repeated samples then see the same context,
so answer-state agreement can be high for both correct and wrong answers.
GPS is used here as a graph-support diagnostic in risk orientation, where higher
means weaker support. It is not a truth label or entailment test. If retrieval
anchors to a wrong but coherent neighbourhood that supports the generated
answer, GPS rates it low-risk; this is the behaviour behind its weakness on the
open-domain bridge tasks (Section~\ref{sec:results_structural}).

Let $\mathcal{E}_q$ and $\mathcal{E}_a$ denote the sets of KG entities linked
to the question $q$ and answer $a$, respectively, filtered to
$\mathcal{E}_a \setminus \mathcal{E}_q$ to exclude trivial self-loops.
Question entities are linked by surface and fuzzy name matching.
Answer entities are linked by the same surface matcher \emph{plus} an
embedding-based soft matcher: candidate answer spans are embedded with the
retrieval encoder and linked to any entity whose name embedding has cosine
similarity at least $\tau$ (the entity-linking threshold, calibrated to
$\tau = 0.60$ and distinct from the SD-UQ numerical constant $\varepsilon$) to
some span.
Each linked answer entity carries a link weight $w_e$: the matched cosine for
soft links and $w_e = 1$ for surface matches.

For GPS, answer entity extraction is restricted to the \emph{primary answer span}:
the first sentence of the response, truncated to 150 characters, which
prevents later explanatory text from inflating the reachable entity count.
If no usable answer entity remains after filtering and soft linking, GPS
abstains rather than returning a spurious support score.

\paragraph{Graph Path Support (GPS; retrieval-state diagnostic).}
GPS scores how strongly the answer is reachable from the question entities
within the retrieved KG.
Each linked answer entity contributes distance-weighted support
$\gamma^{|L_e-\hat L(q)|}$, where $L_e$ is the shortest qualifying path
length (under the same edge-confidence and provenance filters used at
retrieval time), $\hat L(q)$ is the expected reasoning depth for the question,
and unreachable entities contribute zero:
\begin{equation}
  \mathrm{GPS}(q, a) \;=\;
  1 -
  \frac{\sum_{e \in \mathcal{E}_a} w_e\, \gamma^{|L_e-\hat L(q)|}\,
        \mathbb{1}[e\ \mathrm{reachable}]}
       {\sum_{e \in \mathcal{E}_a} w_e}.
\end{equation}
$\mathrm{GPS} = 0$ when every linked answer entity is reachable at the
expected reasoning depth (full structural support, low uncertainty);
$\mathrm{GPS} = 1$ when none are reachable within the configured hop limit (no
structural support, high uncertainty); with $\gamma = 1$ the score reduces to the
unweighted unreachable-entity fraction.
The linking threshold and decay were calibrated on the RealMedQA development
run ($\tau = 0.60$, $\gamma = 0.4$) and applied frozen elsewhere; a
$5\times5$ sweep over $\tau$ and $\gamma$
(Appendix~\ref{sec:appendix_gps_sensitivity}) shows the held-out AUROCs are
stable across the grid, so the reported numbers are not an artefact of the
calibrated cell.
For 2WikiMultiHopQA and MuSiQue, $\hat L(q)$ is the logged per-question hop
count; for HotpotQA variants it is the nominal two-hop depth; for RealMedQA it
is $1$, which makes GPS identical to the calibrated one-hop-decay score on
the clinical domain.
When no answer entities are found in the primary span (or when the
question is a direct-choice comparison), the estimator abstains and the row is
excluded from retrieval-state AUROC while being counted in the
linking-failure rate (Section~\ref{sec:grounding}).

\subsection{Evidence-State Uncertainty Measures}
\label{sec:grounding_metrics}

Evidence-state scores instead test whether the retrieved passages support the
answer, so they can remain informative when sampled answers agree.
These are local answer--evidence consistency scores, not truth labels.

\paragraph{Support Entailment Uncertainty (SEU; evidence-support score).}
SEU is the normalised entailment--contradiction deficit over retrieved chunks,
close to answer-support scoring in RAGAs~\citep{es2024ragas}.

Let $c_1,\ldots,c_K$ denote the $K$ retrieved chunks and $a$ the generated answer.
Each chunk is classified by an NLI model
(\texttt{microsoft/deberta-large-mnli}; all model versions are listed in
Appendix~\ref{sec:appendix_reproducibility}):
$\ell_k \in \{E, N, C\}$ (entailment, neutral, contradiction) where
$E = \mathrm{NLI}(c_k \Rightarrow a)$.
Define the support score $s = (n_E - n_C)/K \in [-1, 1]$,
where $n_E = |\{k: \ell_k=E\}|$ and $n_C = |\{k: \ell_k=C\}|$.
\begin{equation}
  \mathrm{SEU}(q, a, \mathbf{c}) = \frac{1 - s}{2} \in [0, 1].
\end{equation}
$\mathrm{SEU} = 0.0$ when all chunks entail the answer (maximum support,
lowest uncertainty), $\mathrm{SEU} = 0.5$ when chunks are all neutral
(undefined support), $\mathrm{SEU} = 1.0$ when all chunks contradict the
answer (maximum conflict, highest uncertainty).
Neutral chunks create a plateau at $0.5$, especially when a
general-domain NLI model declines to commit on specialist biomedical passages
(Section~\ref{sec:results_grounding}).
Because NLI neutrality can reflect model uncertainty rather than true absence of
support, SEU is read as a local support heuristic, not a calibrated evidence
score.
Unlike answer-state scores, SEU needs only one generation.

\paragraph{Faithfulness to structured provenance.}
SEU does not prove that the decoder followed the retrieved graph path.
A stricter path-faithfulness diagnostic would align answer claims to entities,
relation labels, and relation-anchor text on the retrieved paths.
The runs log those ingredients, but not sentence-level claim-to-path alignment
for every answer; path faithfulness is therefore left as a future
evidence-state diagnostic.

\subsection{Entity-Linking Quality}
\label{sec:grounding}

GPS requires matched question entities and at least one linkable answer entity.
Soft answer linking reduces abstention relative to surface matching
(Section~\ref{sec:results_structural}); yes/no answers remain a predictable
failure case.
For backward compatibility the scalar API emits a $0.5$ sentinel, but AUROC,
expected calibration error (ECE) \citep{guo2017calibration}, and
precision-at-$k$ drop sentinel rows using the recorded null reason.

Entity-linking success rate $g \in [0, 1]$ measures
the fraction of content-bearing query tokens covered by matched entity names:
\begin{equation}
  g(q) = \min\!\left(1,\; \frac{|\hat{\mathcal{T}}_q^{\mathrm{cov}}|}{|\hat{\mathcal{T}}_q|}\right),
\end{equation}
where $\hat{\mathcal{T}}_q$ is the set of distinct alphanumeric content tokens
of length $\geq 4$ in $q$ (excluding stopwords), and
$\hat{\mathcal{T}}_q^{\mathrm{cov}} \subseteq \hat{\mathcal{T}}_q$ is the
subset covered by at least one matched entity name (a token $t$ is covered
when it appears as a sub-token of an entity name, or the entity name contains
$t$ as a substring).
This token-coverage score is a lightweight routing and stratification
heuristic, not an uncertainty estimator.


\begin{table*}[t]
\centering
\small
\setlength{\tabcolsep}{3pt}
\renewcommand{\arraystretch}{1.2}
\caption{%
  \textbf{Expected diagnostic behaviour across KG-RAG retrieval regimes.}
  Symbols: \checkmark~informative; $\sim$~partially informative;
  $\downarrow$~compressed and potentially silent; $\oslash$~abstains;
  \texttimes~locally consistent with a wrong state.
  Each cell states what a family can observe, not a ranking;
  Table~\ref{tab:decomposition} defines the families and
  Table~\ref{tab:failure_taxonomy} gives the resulting $2{\times}2$
  taxonomy.%
}
\label{tab:failure_modes}
\begin{tabular}{@{}p{0.15\linewidth}p{0.27\linewidth}p{0.13\linewidth}p{0.11\linewidth}p{0.11\linewidth}p{0.15\linewidth}@{}}
\toprule
\textbf{Retrieval regime}
&
\textbf{State of the context}
&
\textbf{Answer-state}
&
\textbf{SEU}
&
\textbf{GPS}
&
\textbf{Reading} \\
\midrule
Anchor failure
&
Entity linking fails; system falls back to text retrieval.
&
$\sim$~variation may persist
&
$\sim$~fallback passages
&
$\oslash$~abstains
&
Absence of a graph object is itself diagnostic. \\
\addlinespace[0.25em]
Stabilised retrieval
&
Same or near-same graph neighbourhood across samples.
&
$\downarrow$~compressed
&
\checkmark~support / contradiction
&
\checkmark~consistency
&
Agreement should be read against evidence and graph. \\
\addlinespace[0.25em]
Wrong coherent anchor
&
Anchoring succeeds on the wrong neighbourhood; paths locally coherent.
&
$\downarrow$~same wrong answer repeats
&
$\sim$~can warn
&
\texttimes~supports wrong state
&
Hardest lock-in case; trace inspection needed. \\
\addlinespace[0.25em]
KG incompleteness
&
Plausible anchor, but bridges or relations missing from the graph.
&
$\sim$~fallback restores variation
&
\checkmark~if text remains
&
$\oslash$/$\sim$~reduced
&
Coverage or construction error, not decoder uncertainty. \\
\bottomrule
\end{tabular}
\end{table*}

Table~\ref{tab:failure_modes} is the behaviour map for the three
families across KG-RAG retrieval regimes: it states what each family can observe
as retrieval moves from anchor failure to stabilised, wrong-coherent, and
incomplete-graph states, including where a family goes silent or abstains.

\section{Experimental Setup}
\label{sec:experimental_setup}

\subsection{Datasets}

The suite contains six fixed evaluation snapshots spanning three QA domains:
biomedical control, clinical QA, and open-domain multi-hop QA
(Table~\ref{tab:datasets}).
It deliberately mixes corpus contracts: per-question bundles, per-question
source documents, and shared corpora.
The primary graph object is a passage-provenance KG built from the retrieval
corpus itself, since the question is whether \system{} preserves dense-RAG
answer quality while exposing facts, passages, relations, and retrieval state.
PubMedQA and MuSiQue use $n=100$; RealMedQA uses its full evaluable set
($n=230$); HotpotQA, HotpotQA FullWiki, and 2WikiMultiHopQA use fixed
$n=250$ subsets.
PubMedQA (yes/no) is a boundary control: its label-space answers expose no
linkable answer entity, so GPS is undefined and the audit rule selects nothing,
marking the regime where the retrieval-state arm of the decomposition does not
apply while dense retrieval is already strong.

\begin{table*}[t]
\centering
\small
\caption{Evaluation datasets, corpus contracts, KG scale, and primary roles.
$h$ is the KG traversal depth; \emph{corpus} denotes the retrieval contract;
$n$ is the evaluated subset size.
$|E|$ and $|R|$ count entity nodes and typed entity--entity relations in the
dataset-scoped KG, queried from the persistent Neo4j stores used in the
reported runs.
\label{tab:datasets}}
\setlength{\tabcolsep}{3pt}
\begin{tabular}{@{}lllclrrrp{2.7cm}@{}}
\toprule
Dataset & Domain & Answer type & $h$ & Corpus & $n$ & $|E|$ & $|R|$ & Primary role \\
\midrule
PubMedQA
  & Biomedical & Yes/No/Maybe & 2 & abstract & 100 & 2{,}354 & 3{,}023 & Binary control; strong dense baseline \\
RealMedQA
  & Clinical & Free-text & 2 & shared (143) & 230 & 537 & 359 & Clinical shared-corpus grounding \\
HotpotQA \citep{yang2018hotpotqa}
  & Wikipedia & Free-text & 2 & bundle & 250 & 13{,}546 & 12{,}355 & Open-domain bridge diagnostic \\
HotpotQA FullWiki \citep{yang2018hotpotqa}
  & Wikipedia & Free-text & 2 & shared subset & 250 & 13{,}498 & 12{,}012 & Shared-corpus bridge stress test \\
2WikiMultiHopQA \citep{ho2020wikimultihopqa}
  & Wikipedia & Free-text & 2 & bundle & 250 & 8{,}328 & 7{,}187 & Bridge and comparison analysis \\
MuSiQue \citep{trivedi2022musique}
  & Wikipedia & Free-text & 4 & bundle & 100 & 15{,}440 & 24{,}858 & Hard multi-hop stress test \\
\bottomrule
\end{tabular}
\end{table*}

\textsc{RealMedQA} is the main shared-corpus clinical setting and
\textsc{HotpotQA FullWiki} a controlled Wikipedia shared-corpus stress
snapshot; the rest are closed source-document or bundle settings.
The setup is conservative for KG-RAG: dense retrieval sees small,
benchmark-filtered candidate sets and provides a demanding accuracy baseline.
That curation flatters dense retrieval, and the accuracy comparisons should be
read with that in mind: a dilution probe in
Appendix~\ref{sec:appendix_dilution} shows gold-passage recall falling sharply as
the candidate corpus grows toward deployment scale.
\textsc{PubMedQA} is retained as a control; the open-domain datasets
stress bridge preservation, shared-corpus evidence selection, comparison
questions, and deeper compositional chains.

Throughout, hop count refers to the reasoning chain implied by the
question, not the number of passages supplied by the benchmark.
Explicit decomposition metadata are used where available, and dataset-specific
conventions otherwise.
Hop-stratified analyses are reported as diagnostics rather than the main
table.

\paragraph{Scope.}
The experiments are fixed-subset diagnostic runs, not leaderboard submissions
or web-scale throughput benchmarks.
They support the paper's coarse family-level claims; fine-grained ranking of
neighbouring AUROC or AUREC values would require larger multi-seed sweeps.

\subsection{Model and Sampling}

All answer-generation experiments use GPT-4o-mini
\citep{openai2024gpt4omini}.
For answer-state uncertainty, $N=5$ responses are sampled at temperature
$T=1.0$.
Fixed evaluation subsets are drawn with seed $42$; model versions and the
per-run subset seeds are listed in Appendix~\ref{sec:appendix_reproducibility}
(Table~\ref{tab:run_artifacts}).
We use $N=5$ because it is the low-budget setting that a deployed API user can
afford per question, and the regime in which lock-in's structural ceiling on
answer-state scores is most binding.
This separates an empirical claim from a mechanistic one. \emph{Empirically},
at $N=5$ a large fraction of wrong answers show no observed answer dispersion
(Section~\ref{sec:results_silent}). \emph{Mechanistically}, a genuinely fixed
defective retrieval state yields a fixed answer distribution, so answer-only
uncertainty cannot rank those errors at any $N$; larger $N$ would soften the
empirical footprint but not the mechanism.
These are low-budget black-box diagnostics: the API provides no reliable token
likelihoods, so entropy-style scores use sampled responses rather than decoder
probabilities, and the DSE/P(True)/VN-Entropy cluster and eigen-spectra are
necessarily coarse at $N=5$ (the reported scores are operational ranking
diagnostics rather than high-resolution distribution estimates).

KG construction is largely deterministic: entity and relation extraction use
temperature $0.0$, and benchmark passages are ingested one at a time with
chunk-level provenance.
Extraction is ontology-guided on the biomedical and clinical sets (PubMedQA,
RealMedQA), which supply a domain type schema, and schema-free on the
open-domain sets; the lock-in diagnostics do not depend on the schema.
The studied regime is stabilised entity-first retrieval with explicit
fallback.

\paragraph{Scope of causal claims.}
The comparison holds the corpus, embeddings, and generation model fixed, but it
does not fully isolate graph structure from retrieval stabilisation, fallback
routing, prompt organisation, or provenance grouping
(Section~\ref{sec:limitations}).
The runs report mean pairwise chunk overlap across the $N=5$ calls
(Appendix Table~\ref{tab:retrieval_overlap}); an archived 2WikiMultiHopQA
trace also supports the stability split in
Table~\ref{tab:within_adaptive_stability}.
The headline runs do not retain per-question entity- or path-overlap statistics
for every dataset, but a dedicated graph-state diversity run on HotpotQA-FullWiki
logs the full per-sample seed-entity, path, subgraph, and chunk overlap and is
analysed in Appendix~\ref{sec:appendix_diversity}.

\subsection{Correctness Labels}
\label{sec:correctness}

Correctness labels are binary.
Label-space tasks are normalised to their answer contract
(\texttt{yes}/\texttt{no}/\texttt{maybe}).
Free-text and factoid tasks use a reference-based semantic judge given the
question, gold answer, aliases where available, and model response; by default
the judge is GPT-4o-mini
(Appendix~\ref{sec:appendix_reproducibility}).
The judge performs benchmark answer normalisation, not open-ended clinical
diagnosis: it decides reference-grounded semantic equivalence, for which
exact match would undercount aliases, paraphrases, and short explanatory
answers.
Same-model judging is a known limitation
\citep{zheng2023judging,panickssery2024llm}, so the labels are practical
reference-grounded judgements rather than an oracle.
An independent re-judge with a different model family (Llama-3.3-70B) on the
saved answers of every free-text run agrees with the original labels on
$92$--$99\%$ of answers
($\kappa = 0.52$--$0.97$, the low end being the near-ceiling-accuracy
RealMedQA adaptive run where few wrong answers make $\kappa$ unstable) and
leaves every central contrast unchanged (Section~\ref{sec:limitations}).

Two conventions matter.
Provider-side failures are recorded explicitly, so results report both
\textbf{raw accuracy} and \textbf{clean accuracy}, the latter excluding
generation failures (Appendix Table~\ref{tab:generation_failures}).
For free-text datasets, prompts elicit short explanatory answers rather than
extractive spans; semantic correctness is therefore the headline answer metric,
with EM/F1 retained only in the artefacts.

\subsection{Evaluation Metrics}

Each uncertainty score is evaluated mainly by \textbf{AUROC}: how well the score
ranks incorrect above correct answers.
\textbf{AUREC} (area under the risk-excess-coverage curve; lower is better) is
reported only for the dense-side selective-prediction frontier
(Appendix~\ref{sec:appendix_dense_frontier})
\citep{geifman2017selective}, not as a parallel metric in the main AUROC tables.
Raw accuracy, clean accuracy, retrieval overlap, and per-metric compute time
are also logged; the marginal cost of each diagnostic, including SEU's
per-chunk NLI calls and GPS, is reported in
Appendix Table~\ref{tab:compute_cost}.
All main figures report point estimates from one fixed subset per dataset.
With mixed sample sizes ($n=100$, $n=230$, $n=250$), small AUROC gaps are
descriptive unless they recur across signal families or match the qualitative
traces.
No multiple-comparison correction is applied because the heatmap is not used
for family-wise significance testing.
The two primary headline contrasts instead carry paired-bootstrap intervals.
The largest (the adaptive-versus-strict SD-UQ collapse of $+0.52$ AUROC at
$n=196$) is far from zero. Gaps near $0.05$ are not treated as significant: at
these sample sizes the bootstrap cannot reliably resolve effects that small, so
the intervals are used descriptively rather than as a formal power analysis.
Appendix Tables~\ref{tab:run_artifacts}, \ref{tab:generation_failures},
\ref{tab:effective_denominators}, \ref{tab:hotpot_fullwiki_stress}, and
\ref{tab:headline_ci_output} document run configurations, answered and failed
counts, GPS usable counts, the HotpotQA FullWiki snapshot, and bootstrap
intervals.
Retrieval-state AUROC with fewer than roughly fifty usable rows is treated as
trace-level evidence.
GPS AUROC is always conditional on rows for which GPS is defined; linking
failures and other abstentions are reported through usable/answered
denominators and audit-rule coverage, not hidden inside the AUROC.

The headline tables report the family representatives of
Section~\ref{sec:metrics} (SD-UQ, SEU, GPS), with DSE, SRE-UQ, and VN-Entropy as
within-family controls; the P(True) proxy is omitted as monotone in DSE, and
SelfCheckGPT is kept in the artefacts.
GPS values are computed by post-hoc replay on saved answer logs and persistent
KGs without rerunning answer generation or document retrieval; primary GPS
numbers reuse stored linked entities and path lengths where available, while the
coverage-raising and gold-reachability replays recompute entity alignment and
path support against the persistent KG
(Appendix~\ref{sec:appendix_reproducibility}).

\subsection{Retrieval and KG Configuration}

Both systems use \texttt{all-MiniLM-L6-v2}.
Vanilla RAG performs direct vector retrieval over chunks.
KG-RAG performs entity-first retrieval, graph expansion, and context assembly
from graph-linked chunks, paths, and entities, with dense retrieval and
retriever-first graph expansion as fallbacks when entity anchoring is weak.
Each KG-RAG response records its route
(\texttt{entity\_first}, \texttt{rfge}, or \texttt{semantic\_only}) and route
reason.

Dataset-scoped KGs are built passage-wise, preserving question and passage
provenance and avoiding artificial cross-question chunking.
Each artefact retains chunk identifiers, passage titles where available,
relation paths, route labels, and relation-anchor text when supplied by the KG
extractor.
Dense retrieval can log passages and scores; the KG layer additionally exposes
matched entities, typed triples, graph paths, relation anchors, and abstention
events, so the KG trace is structured provenance.
Traversal depth is dataset-specific: $h=2$ for PubMedQA, RealMedQA,
HotpotQA, HotpotQA FullWiki, and 2Wiki; $h=4$ for MuSiQue.
During iterative decomposition, each sub-question uses a bounded local graph
expansion cap so that overall chain length and per-step search breadth are not
confounded.

\section{Results}
\label{sec:results}

The results follow the lock-in mechanism, not a leaderboard.
We first check whether KG-RAG shows any detectable accuracy gap against dense
retrieval (Section~\ref{sec:results_accuracy}), then measure how often answers go
silently wrong under deployable policies
(Section~\ref{sec:results_silent}), show that answer-state ranking
collapses once retrieval is forced to concentrate
(Sections~\ref{sec:results_output}--\ref{sec:results_collapse}), and
finally test whether evidence- and retrieval-state signals fill the gap,
distilling the decomposition into a conjunctive audit rule
(Sections~\ref{sec:results_grounding}--\ref{sec:results_composite}).
Table~\ref{tab:results_snapshot} gives the running summary.

\begin{table*}[t]
\centering
\footnotesize
\caption{%
  \textbf{KG-RAG shows no statistically detectable accuracy gap with dense
  retrieval, while exposing per-family diagnostics dense retrieval cannot.}
  Accuracy is clean semantic accuracy (provider failures excluded).
  The displayed $n$ counts answered KG rows; $\Delta$ is KG $-$ dense accuracy.
  Dense and KG columns report clean accuracy on each system's own answered
  rows. Paired deltas and McNemar tests use rows answered by both systems
  (Appendix Table~\ref{tab:mcnemar_accuracy}).
  GPS is the graph-support diagnostic (lower is stronger support) with its
  usable/answered denominator; the per-family answer-, evidence-, and
  retrieval-state AUROCs are in Figure~\ref{fig:adaptive_kg_auroc_heatmap}.
  The \emph{Main reading} column summarises each row; intervals and denominators
  are in Appendix~\ref{sec:appendix_reproducibility}.
  GPS was calibrated on RealMedQA ($\tau{=}0.60$, $\gamma{=}0.40$) and applied
  frozen elsewhere; the Main reading column flags where GPS is in its
  calibration domain, near chance, or weak.%
}
\label{tab:results_snapshot}
\setlength{\tabcolsep}{3.2pt}
\begin{tabular}{@{}lcccclp{3.4cm}@{}}
\toprule
Dataset & $n$ & Dense acc. & KG acc. & $\Delta$ & GPS (usable) & Main reading \\
\midrule
PubMedQA & 100 & 0.750 & 0.730 & $-0.020$ & -- &
Binary control; GPS abstains (no answer entities). \\
RealMedQA & 223 & 0.950 & 0.946 & $-0.004$ & 0.76 (195/223) &
Clinical near-match; GPS calibration domain. \\
HotpotQA & 238 & 0.655 & 0.605 & $-0.050$ & 0.51 (158/238) &
Dense stronger; KG adds anchors and paths. \\
HotpotQA FullWiki & 218 & 0.721 & 0.665 & $-0.056$ & 0.54 (185/218) &
Shared-corpus stress; GPS near chance. \\
2WikiMHQA & 244 & 0.712 & 0.689 & $-0.023$ & 0.38 (182/244) &
Near-match; GPS weak on bridges. \\
MuSiQue & 94 & 0.478 & 0.394 & $-0.084$ & 0.68 (83/94) &
Hard multi-hop; auditable, not competitive. \\
\bottomrule
\end{tabular}
\end{table*}

\begin{figure*}[t]
\centering
\includegraphics[width=\linewidth]{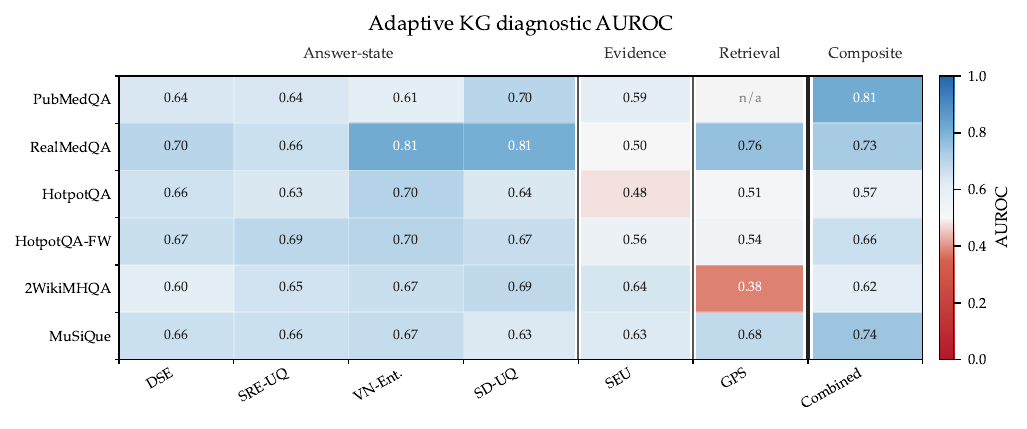}
\caption{%
  \textbf{Answer-state scores rank errors most consistently; GPS discriminates
  only in its calibration domain (RealMedQA).}
  Adaptive KG-side diagnostic AUROC across the six snapshots, grouped by family;
  incorrect answers are the positive (risk) class and the diverging colour scale
  is neutral at chance ($0.50$), red below, blue above.
  Vertical rules separate the answer-, evidence-, and retrieval-state families;
  the Combined column is the mean of within-dataset percentile ranks of SD-UQ,
  SEU, and GPS-risk (abstentions imputed high-risk, so it differs from the
  usable-row GPS column). PubMedQA GPS is undefined (yes/no answers expose no
  entities).
  Thin-denominator cells (notably MuSiQue) carry wide intervals and are read
  qualitatively; denominators and bootstrap $95\%$ CIs are in Appendix
  Tables~\ref{tab:effective_denominators} and~\ref{tab:headline_ci_output}.%
}
\label{fig:adaptive_kg_auroc_heatmap}
\end{figure*}

\subsection{Accuracy: no statistically detectable KG--dense gap at these sample sizes}
\label{sec:results_accuracy}

KG-RAG point estimates sit slightly below dense retrieval on all six snapshots
($0.004$ to $0.084$), but no per-snapshot difference is statistically reliable
(paired McNemar, all $p \geq 0.12$; Appendix Table~\ref{tab:mcnemar_accuracy}).
This is a no-detectable-gap result at these sample sizes, not an equivalence
claim.
The comparison is deliberately conservative: we do not claim KG-RAG is more
accurate, only that it exposes answer-, evidence-, and retrieval-state
diagnostics the dense baseline never logs.
Comparable accuracy makes the comparison fair, but it is not the contribution.
The strict graph-only stress test forces retrieval to concentrate, a synthetic
composite worst-case probe rather than a controlled ablation: it disables dense
fallback, augmentation, reranking, and decomposition together, so the collapse is
not attributed to graph structure alone.
It is a diagnostic contrast, not evidence that removing any one component would
cause the same failure.

\begin{table}[t]
\centering
\small
\setlength{\tabcolsep}{6pt}
\renewcommand{\arraystretch}{1.25}
\caption{%
  \textbf{Failure taxonomy for retrieval-augmented generation.}
  Retrieval-state lock-in (top right) is the focal regime: the retrieval state
  is defective while answer-state uncertainty is low.
  Absence lock-in is flagged by retrieval-state abstention; presence lock-in,
  a coherent wrong neighbourhood, remains the hard case.
  At $N{=}5$, $42\%$ of adaptive-KG errors fall in this silent cell
  ($59\%$ dense, $84\%$ strict).%
}
\label{tab:failure_taxonomy}
\begin{tabular}{@{}l >{\raggedright\arraybackslash}p{0.30\linewidth}
                     >{\raggedright\arraybackslash}p{0.30\linewidth}@{}}
\toprule
& \textbf{Retrieval state correct}
& \textbf{Retrieval state wrong / missing} \\
\midrule
\textbf{Low answer-state}  & \cellcolor{blue!12}\textbf{Certified low-risk}
                           & \cellcolor{badred!15}\textbf{Retrieval-state lock-in} \\
\textbf{uncertainty}       & \cellcolor{blue!12}all three families concur
                           & \cellcolor{badred!15}answer-state silent; absence variant flagged by retrieval-state abstention, presence variant open \emph{(this paper)} \\
\addlinespace[0.3em]
\textbf{High answer-state} & \cellcolor{yellow!12}\textbf{Ordinary uncertainty}
                           & \cellcolor{black!6}\textbf{Retrieval failure} \\
\textbf{uncertainty}       & \cellcolor{yellow!12}answer-state flags it
                           & \cellcolor{black!6}answer-state flags it; retrieval-state abstains or shows no support \\
\bottomrule
\end{tabular}
\end{table}

\subsection{Finding 1: Answer-state uncertainty ranks errors but is silent on a large slice}
\label{sec:results_silent}

\emph{Answer-state uncertainty ranks errors well in ordinary adaptive runs, yet
a non-trivial share of wrong answers carry no answer-state signal at all.}
Table~\ref{tab:silent_failures} reports the \emph{silent-failure rate}: the
fraction of wrong answers with zero answer-state uncertainty (DSE $=0$ and
SD-UQ at its numerical floor), so no within-question disagreement signal remains.
The blind spot belongs to the whole family: identical samples contain no signal
for DSE, SD-UQ, VN-Entropy, or any answer-dispersion variant.
The footprint is strongly dataset-dependent: $8\%$ on the clinical domain
(RealMedQA) to $55\%$ on 2WikiMultiHopQA, pooling to $42\%$ (adaptive KG) and
$59\%$ (dense).

The strict probe reaches $84\%$, but that figure is not comparable to the
deployable-policy rates: it comes from the multi-component stress test above, so
its elevation reflects engineered absence of retrieval, not graph structure
alone.

Dense retrieval produces \emph{more} silent errors than adaptive KG-RAG, so the
phenomenon is not graph-specific. Nor is it a low-budget artefact: an $N{=}20$
probe still leaves $68\%$ of dense 2WikiMultiHopQA errors strictly silent
(Appendix~\ref{sec:appendix_silent_sensitivity}).

The KG-specific risk is \emph{auditable wrongness}: a silent error arrives with
matched entities, typed paths, and provenance anchors, making the system look
structurally auditable while it is wrong.

\begin{table}[t]
\centering
\footnotesize
\caption{%
  \textbf{Silent-failure rates with Wilson 95\% intervals.}
  Fraction of wrong answers with zero answer-state uncertainty
  (DSE $=0$ and SD-UQ at its numerical floor), per retrieval policy.
  $n$ is the number of wrong answered rows.%
}
\label{tab:silent_failures}
\setlength{\tabcolsep}{2.1pt}
\begin{tabular}{@{}lccc@{}}
\toprule
Dataset & Dense & Adaptive KG & Strict KG \\
\midrule
PubMedQA    & .84 [.65,.94] (25) & .67 [.48,.81] (27) & --        \\
RealMedQA   & .09 [.02,.38] (11) & .08 [.01,.35] (12) & .73 [.61,.82] (66) \\
HotpotQA    & .59 [.48,.69] (78) & .46 [.36,.56] (94) & --        \\
HotpotQA-FW & .45 [.33,.58] (58) & .18 [.11,.28] (73) & --        \\
2WikiMHQA   & .79 [.67,.87] (61) & .55 [.44,.66] (76) & .90 [.84,.94] (145)\\
MuSiQue     & .51 [.37,.65] (47) & .42 [.30,.55] (57) & --        \\
\midrule
Pooled      & .59 [.53,.65] (280) & .42 [.36,.47] (339) & .84 [.79,.89] (211) \\
\bottomrule
\end{tabular}
\end{table}

At the operational $N{=}5$ budget, a method that only observes sampled-answer
disagreement can recall at most $41\%$ of dense errors, $58\%$ of adaptive
KG errors, and $16\%$ of strict-stress-test errors; the remaining wrong
answers provide no within-question disagreement signal to rank, whatever
weighting it applies to the samples.
That missing observable motivates evidence-state and retrieval-state
diagnostics.

\subsection{Answer-state metrics are strong but not sufficient}
\label{sec:results_output}

The next question is how much answer-state estimators still contribute when
retrieval remains adaptive.
Of the eight measures (Table~\ref{tab:metric_taxonomy}), the additional
answer-side controls do not outperform SD-UQ in this low-sample regime, so the
narrative tracks the three family representatives (SD-UQ, SEU, GPS) and the
appendix carries the rest.
In the adaptive runs they remain the strongest default: the AUROC heatmap
(Figure~\ref{fig:adaptive_kg_auroc_heatmap}) shows that answer-side scores have
the clearest overall association with error, especially when retrieval and
decoding still produce meaningful answer variation.
When graph routing stabilises the context around a wrong entity or path, answer
disagreement can be low even for incorrect answers; the score then measures the
smoothness of the answer surface rather than the reliability of the retrieved
state.
Answer-state uncertainty remains a strong default, but it is not a lock-in
diagnostic.

Within that family, the embedding-geometric scores VN-Entropy and SD-UQ are the
most dependable answer-side baselines in these low-sample runs, matching or
beating the imported semantic-entropy and perturbation scores on every KG
snapshot but MuSiQue (per-dataset AUROCs and intervals in
Appendix Table~\ref{tab:headline_ci_output}).
They differ mainly in how they degrade as retrieval concentrates:
hard-clustering scores such as DSE lose resolution and drift toward chance once
overlap rises, whereas SD-UQ retains discriminative power by first projecting
out the question direction and then measuring the residual spread among answers
(cf.\ Appendix Table~\ref{tab:within_adaptive_stability}).
This robustness costs little: SD-UQ needs only
question and answer embeddings, with no log-probabilities, NLI calls, or model
internals.
One caveat matters for this comparison: SD-UQ and VN-Entropy read the
geometry of an external embedding space that the cluster- and NLI-based scores
(DSE, SelfCheckGPT) do not use, so the answer-state contrast is not strictly
like-for-like.

\subsection{Finding 2: Strict graph-only stress collapses answer-state ranking}
\label{sec:results_collapse}

\emph{Forcing retrieval to concentrate drives wrong answers into the
low-dispersion corner, so answer-state ranking degrades to or below chance.}
The sharpest stress-test signature appears in the strict graph-only regime,
where answer-state uncertainty becomes a poor guide to correctness.
That answer dispersion falls under forced concentration is expected; the
informative part is what the collapse is made of, which the route
decomposition below recovers.

\begin{figure*}[t]
\centering
\includegraphics[width=\linewidth]{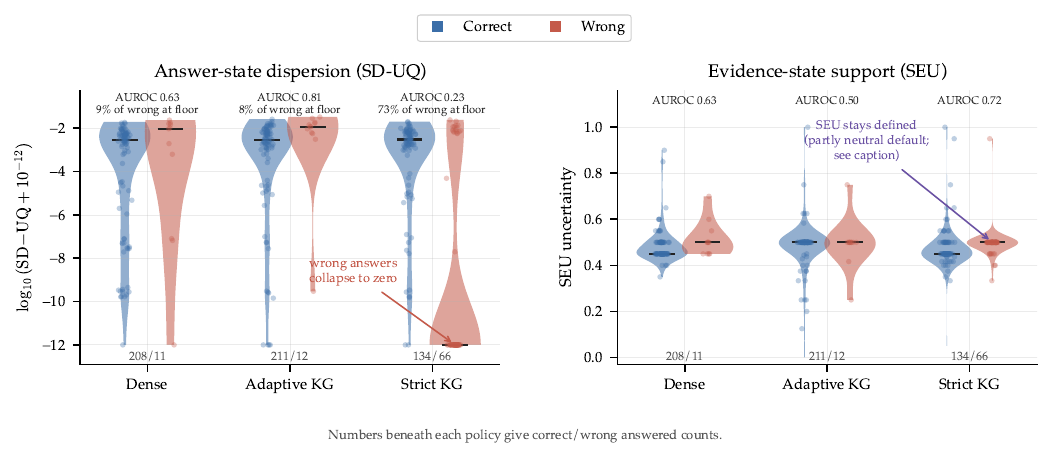}
\caption{%
  \textbf{Under strict retrieval the answer surface collapses while evidence
  support stays separated (RealMedQA: dense, adaptive, strict graph-only).}
  Violin/strip distributions split by correctness, with AUROC per panel.
  $73\%$ of wrong answers fall to the SD-UQ floor under strict retrieval
  ($48/66$ vs.\ $1/12$ adaptive; Fisher exact $p = 4\times10^{-5}$); SEU stays
  separated, but mainly because empty-retrieval wrong rows take its neutral
  default (Table~\ref{tab:silent_accounting}).
  This clinical collapse is driven by empty-retrieval (absence) rows, distinct
  from the populated-route, presence-compatible silent errors that dominate the
  adaptive-policy footprint.
  Counts beneath each policy are correct/wrong answered rows.%
}
\label{fig:context_collapse_ablation}
\end{figure*}

The stress test moves errors into the low-dispersion corner: $43/196$ RealMedQA
and $87/242$ 2WikiMultiHopQA questions shift from correct-and-adaptive to
wrong-and-silent under strict retrieval (Appendix
Figure~\ref{fig:lockin_migration}).
Clean KG accuracy falls to $0.670$ and chunk overlap rises from $0.558$ to
$0.661$ (cf.\ Appendix Table~\ref{tab:retrieval_overlap}).
Answer-state ranking collapses: DSE AUROC $0.463$, VN-Entropy $0.248$, SD-UQ
$0.233$, at or below chance, actively misleading
(Figure~\ref{fig:context_collapse_ablation};
Appendix Table~\ref{tab:realmedqa_strict_entity}).
The $0.233$ reflects engineered absence of retrieval, not a generic failure on
populated lock-in (where ranking stays healthy).
The adaptive-to-strict SD-UQ gap on the paired subset is $+0.52$ AUROC
(bootstrap 95\%~CI $[0.28, 0.73]$, $n=196$).
SEU ($0.721$) and GPS ($0.746$) still rank errors on this strict run (GPS in its
calibration domain); the contrast holds under an independent Llama-3.3-70B judge
($\kappa = 0.89$).

A verbalised-confidence probe recovers much of the clinical mass (AUROC $0.89$
vs.\ SD-UQ's $0.233$) at one call, but sits near chance on 2WikiMultiHopQA and
returns no auditable provenance (Appendix~\ref{sec:appendix_verbalized}).
We keep SD-UQ as the answer-state representative for consistency with the
adaptive tables; the verbalised probe dominates it in the strict regime but not
on the multi-hop tasks the decomposition targets.

\paragraph{Route decomposition.}
The collapse decomposes into three components.

\emph{First, it is not a chunk-level artefact.}
An interventional dose-response reduces chunk overlap from $0.67$ to $0.29$,
but leaves SD-UQ AUROC flat ($0.18$) and the silent-error count unchanged
(Appendix~\ref{sec:appendix_doseresponse}).

\emph{Second, route logging shows where the silent state lives.}
Every silent error in the strict clinical run is an empty-retrieval row
($48/48$ at $n=230$): strict anchoring found no usable context, the same empty
state recurred across all five samples, and the generator answered unanimously
from parametric memory.
The nearest explanation is the removal of fallback and vector augmentation in
this stress run, but the design still changes multiple components and does not
isolate one as the cause.
On the populated-route rows, answer-state ranking is healthy (SD-UQ AUROC
$0.79$).
Much of SEU's apparent survival is mechanical: empty-retrieval wrong rows
receive the all-neutral default $0.5$, which ranks above confidently supported
correct rows; populated-row SEU is only $0.56$.
This is \emph{absence} lock-in, whose simplest reliable flag is the
empty-route/abstention observable, not answer dispersion or evidence
entailment.

\emph{Third, the strict 2WikiMultiHopQA run decomposes the same way but leaves
a remainder.}
Of its $130$ silent errors, $110$ are empty-retrieval, while $20$ ride
populated entity-first routes into wrong-but-coherent neighbourhoods of the
Osireion type (\emph{presence} lock-in).
Populated-row SD-UQ stays degraded there ($0.65$).
Presence lock-in is the harder case: the answer surface and the route
observable are both silent, leaving only evidence-level checks.
It is also a setting where SEU itself proved unreliable (the strict 2-hop
slice), which is why the composite recommendation does not reduce to any single
surviving family.

\paragraph{Adaptive-policy footprint.}
Under the deployable adaptive policy the silent errors are mostly
presence-like.
Of the $141$ pooled adaptive silent errors, $89$ carry recorded route metadata,
all on populated routes with none on the empty-retrieval route, so the
deployable-policy footprint is not an empty-retrieval artefact.
The logging gap does not carry that claim: the remaining $52$ of the $141$
silent errors have no route metadata, so their absence/presence mix cannot be
read from the saved logs.
But even under the worst case for our reading (all $52$ unlogged rows being
empty-retrieval), the populated-route share is still at least
$89/141 \approx 63\%$, so populated-route, presence-compatible silent errors
dominate the deployable-policy footprint under any imputation.

Decomposing those $141$ adaptive silent errors by family, $134$ ($95\%$) carry
at least one non-answer flag and only $7$ ($5\%$) remain unflagged. The two
non-answer families overlap on the flagged rows: $67$ show an evidence-state
contradiction ($\mathrm{SEU}>0.5$) and $127$ weak or absent graph support
(GPS $>0.5$ or abstention).

A manual audit of these seven (Appendix~\ref{sec:appendix_residue}) finds
mostly benchmark-ambiguous questions, wrong-answer-slot errors, and parametric
conflations rather than confirmable deep presence lock-in whose evidence
\emph{entails} the wrong answer (the Baltic Cup type of
Table~\ref{tab:lockin_gallery}); chunk text is re-retrievable from the KG, but
confirming that deceptive type needs the cross-passage entailment and
path-faithfulness replay described below, not run for this log-only audit.
This decomposition helps separate lock-in from look-alike confounders:
those $134$ flagged rows carry a concrete evidence- or retrieval-state mechanism,
while the residue is confounders rather than confirmable lock-in, which is why
the silent-error rate is reported throughout as an upper bound on lock-in
prevalence.

Flagging the cases where all three diagnostic arms fail, with SEU and GPS both
near zero on a wrong answer, requires two checks this log-only analysis does not run:
cross-passage contradiction mining over the re-retrieved chunk text, and
relation-level path faithfulness to check that the connecting path uses the
relation the question asked for rather than a locally coherent but wrong one.
Both are concrete, retrieval-replay-only extensions, not new theory.
Table~\ref{tab:silent_accounting} makes the accounting explicit.

\begin{table*}[t]
\centering
\footnotesize
\caption{%
  \textbf{Silent-error accounting from saved logs.}
  Buckets are mutually exclusive and apply only to wrong answered rows.
  ``Max answer recall'' is the structural upper bound for any
  answer-dispersion-only detector at the $N{=}5$ budget: silent wrong answers
  contain no sampled-answer disagreement signal.
  Empty denotes absence lock-in; Pop.+SEU and Pop.+GPS are populated-route
  silent errors flagged by evidence or graph-support signals.
  Unflagged is the residue with no non-answer-family flag.%
}
\label{tab:silent_accounting}
\setlength{\tabcolsep}{3pt}
\begin{tabular}{@{}lrrrrrrrr@{}}
\toprule
Run & Wrong & Silent & Max answer recall & Empty & Pop.+SEU & Pop.+GPS & Route-unk.+flag & Unflagged \\
\midrule
Adaptive KG pooled & 339 & 141 & .58 & 0 & 40 & 44 & 50 & 7 \\
Strict KG pooled   & 211 & 178 & .16 & 158 & 13 & 6 & 0 & 1 \\
RealMedQA strict   & 66  & 48  & .27 & 48  & 0  & 0 & 0 & 0 \\
2WikiMHQA strict   & 145 & 130 & .10 & 110 & 13 & 6 & 0 & 1 \\
\bottomrule
\end{tabular}
\\[2pt]
{\footnotesize Pop.\ buckets are route-known; because GPS can be defined without
a recorded route label, $2$ of the Adaptive-KG Unflagged rows are route-unknown,
so the $89$ route-known / $52$ route-unknown split in
Section~\ref{sec:results_collapse} is not a direct column sum.}
\end{table*}
Which signal survives depends on the variant: route or abstention
observables for absence, and evidence-level checks, imperfectly, for presence.

\paragraph{Cross-dataset corroboration.}
The same pattern appears on the 2WikiMultiHopQA $n=250$ run
(cf.\ Appendix Table~\ref{tab:twowiki_hopwise_diagnostics}).
The adaptive KG policy stays close to dense retrieval overall ($0.689$ vs.\
$0.712$) and matches it on the 4-hop slice.
On the strict 4-hop slice, accuracy is zero and both SD-UQ and VN-Entropy
collapse to near-zero: the answer surface is stable precisely when the retrieved
graph state has become wrong enough to make every answer fail.
The collapse signature is the \emph{joint} collapse of SD-UQ and VN-Entropy
together with failed answers, not low SD-UQ alone (the dense 4-hop row has low
SD-UQ but healthy VN-Entropy and high accuracy).

Per-question support comes from a dedicated diversity run on HotpotQA-FullWiki
($n=221$, $74$ wrong; Appendix Table~\ref{tab:diversity_run}).
Across $N{=}5$ samples the retrieval state is highly stable (seed-entity Jaccard
$0.99$; path/subgraph/chunk overlap $0.58$--$0.65$), and among wrong answers
more overlap predicts lower dispersion ($\rho = -0.30$/$-0.23$/$-0.24$, all
$p<0.05$; the 2WikiMHQA trace is consistent: $\rho = -0.32$, $p=0.01$).
These analyses support the mechanism without identifying it causally
(Appendix~\ref{sec:appendix_diversity}, \ref{sec:appendix_concentration};
Appendix Figure~\ref{fig:concentration_facets}).

\paragraph{Boundary case (HotpotQA).}
HotpotQA supplies a useful counterweight: dense retrieval leads on accuracy
($0.655$ vs.\ $0.605$) while answer-state AUROCs stay healthy on both systems
and GPS is only weakly discriminative ($0.51$/$0.54$; Appendix
Table~\ref{tab:hotpot_fullwiki_stress}).
These runs bound the claim: when dense retrieval already has a compact,
well-indexed candidate set, graph organisation adds provenance but does not
improve answer quality or retrieval-state ranking.

\paragraph{Selective prediction.}
Figure~\ref{fig:selective_prediction} reframes the results as selective
prediction: if a diagnostic score is high, should the system answer, broaden
retrieval, or send the case to review?
Under adaptive runs, answer-side scores give the cleanest rejection signal;
under strict lock-in, SEU and GPS must be read as provenance checks rather than
weaker versions of answer-state entropy.
On the strict 4-hop slice ($n=58$) every answer is wrong, so AUROC is undefined.

\subsection{Finding 3: Evidence-state helps unevenly; retrieval-state is auditable but brittle}
\label{sec:results_grounding}

\emph{Once the answer surface goes silent, evidence-state and retrieval-state
signals recover part of the missing footprint, but not the hardest
presence-lock-in cases.}
When the answer surface goes silent, the evidence-side question is whether the
retrieved text betrays the error. This matters most in high-accuracy or near-match
settings, where the deployment question is whether the answer can be justified by
retrieved evidence.

SEU's signal is dataset-dependent.
It reaches $0.721$ AUROC on the strict RealMedQA stress test, but much of that
comes from the all-neutral default over empty-retrieval rows rather than
genuine entailment (Section~\ref{sec:results_collapse}).
On the strict 2WikiMultiHopQA 2-hop slice it falls below chance.
Bridge questions retrieve passages about the wrong-but-coherent anchor, and
those passages weakly \emph{entail} the wrong answer from that neighbourhood
rather than contradicting it.
On the adaptive RealMedQA run, $74\%$ of rows hit the all-neutral $0.5$ plateau
because no chunk entails or contradicts the answer, giving a chance-level AUROC.
Evidence support helps in some regimes, but its utility varies
with the retrieval policy and the NLI model's domain competence.

A domain-NLI ablation on PubMedQA (a proxy for the clinical plateau, not a
direct RealMedQA rerun; Appendix~\ref{sec:appendix_seu_nli_ablation}) tests
whether the neutral plateau is an NLI-domain artefact.
We recompute SEU with a biomedical-aware LLM-NLI judge
(\texttt{gpt-4o-mini}) in place of the paper's \texttt{deberta-large-mnli}.
AUROC rises from $0.55$ to $0.64$ on the dense run and from $0.60$ to $0.63$
on the KG run, so part of the plateau is indeed a general-domain-NLI
artefact.
The improvement is modest, however, and the neutral rate does not fall under
the LLM judge; it rises.
The plateau is not \emph{entirely} an artefact: the retriever
genuinely returns evidence that neither entails nor contradicts many answers.
Sentence-level entailment scoring is the plausible refinement, but it requires
re-running retrieval rather than replaying logs.
The selective-prediction curves in Figure~\ref{fig:selective_prediction}
and the fixed-coverage operating points in Appendix
Table~\ref{tab:operating_points} show the same regime switch as an abstention
decision.

\begin{figure*}[t]
\centering
\includegraphics[width=\linewidth]{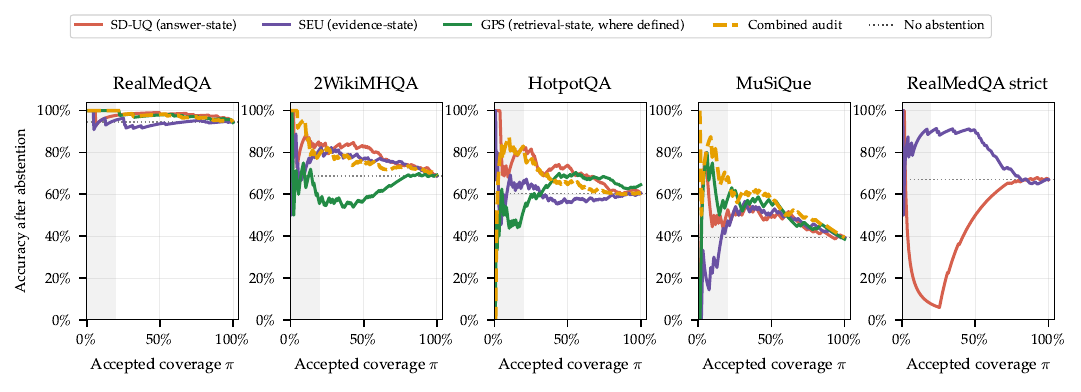}
\caption{%
  \textbf{The useful diagnostic changes with the retrieval regime.}
  Selective prediction with different KG-side diagnostic families: panels are the
  four adaptive datasets (RealMedQA, 2WikiMultiHopQA, HotpotQA, MuSiQue) and the
  RealMedQA strict stress test (rightmost).
  In the strict stress test, SD-UQ gating \emph{inverts}, accepting the silent
  wrong mass first, while SEU gating climbs toward $90\%$.
  Curves sort answered KG-RAG queries by increasing uncertainty and report the
  accuracy retained at each coverage level; the dotted line is no-abstention
  accuracy and the shaded band marks the noisy $<20\%$-coverage region.
  GPS and the combined audit are omitted from the strict panel because GPS comes
  from post-hoc replay rather than row-level logs.%
}
\label{fig:selective_prediction}
\end{figure*}
Under the adaptive policy, gating on SD-UQ alone is the strongest simple
abstention rule.
Under the strict clinical stress test that rule becomes uninformative: $0.338$
error at $80\%$ coverage against a $0.330$ no-abstention base.
SEU gating instead cuts the residual error to $0.269$, an $18\%$ relative
reduction in the regime where the answer surface is silent.
A single global ranking hides this regime-dependent switch.

\paragraph{Retrieval-state diagnostics: calibrated but brittle.}
\label{sec:results_structural}
If evidence-state (SEU) is hit-or-miss, retrieval-state (GPS) offers a
different, graph-specific lens, but its utility depends entirely on whether the
retrieval policy exposes the gold entity.
Soft linking fixes coverage, but it cannot fix ranking; the held-out AUROCs
remain poor.
GPS behaviour is domain-dependent.
On the clinical shared-corpus setting where its two hyperparameters were
calibrated, it is a useful error ranker ($0.76$ adaptive, $0.75$ strict).
On the held-out suites transfer is uneven: MuSiQue reaches $0.68$ with
per-question depth, while HotpotQA, FullWiki, and 2WikiMultiHopQA hover near or
below chance (exact AUROCs in the breakdown below).
A sensitivity replay over $\tau\in\{0.50,\ldots,0.70\}$ and
$\gamma\in\{0.2,\ldots,1.0\}$ leaves the held-out open-domain GPS AUROCs
essentially unchanged (SD $\leq 0.018$ for HotpotQA, FullWiki,
2WikiMultiHopQA, and MuSiQue; Appendix Table~\ref{tab:gps_sensitivity}).
The weak 2WikiMultiHopQA ranking is not a threshold/decay artefact.
GPS scores graph support for the generated answer, not correctness.
It can rate a wrong-but-coherent neighbourhood as low risk.
In the Baltic Cup case of Table~\ref{tab:lockin_gallery}, the wrong entity is
reachable, GPS is $0.00$, SEU is $0.00$, and the answer surface has near-zero
dispersion.
The decomposition makes that failure visible but does not solve it.
Retrieval-state AUROC is computed only on rows with linkable answer
entities, with linking failures reported as abstentions
(Table~\ref{tab:results_snapshot}).

Even where the scalar fails, GPS still provides an auditable retrieval-state
trace of matched entities, paths, and reachability on every dataset where it is
defined, which is useful for provenance regardless of ranking quality.
The held-out weakness is a property of the retrieval policy: on these suites the
policy often leaves the gold entity outside the retrieved subgraph even when the
answer is correct, so GPS-risk and correctness decouple (we test this prediction
in Appendix Table~\ref{tab:gold_reachability}).
GPS is most reliable as an abstention record, usable denominator, and trace-level
support check; it is one scalar instantiation of the retrieval-state class
(anchors, paths, routes, abstention), not the family itself.
A fixed-bin reliability check makes the same point without AUROC:
RealMedQA error rates rise from $0/39$ in the lowest GPS-risk bin to $6/42$
in the highest, but the open-domain bins are not monotone, especially on
2WikiMultiHopQA (cf.\ Appendix Table~\ref{tab:gps_reliability_bins}).

Soft answer-entity linking reduces the abstention cost (e.g.\ $185/218$ on
HotpotQA FullWiki, up from $69$ under surface matching); PubMedQA still abstains
completely because binary answers expose no answer entities.

GPS can mis-rank when the retrieval policy makes correct answers harder to reach
than wrong ones: a \emph{positive} gold-unreachability gap would tend to invert
the signal toward below-chance GPS.
This is consistent with both cases examined: on the regenerated HotpotQA
FullWiki KG a negative gap ($-20$ points) goes with near-chance GPS ($0.54$), and
on an earlier 2WikiMultiHopQA build a positive gap ($+26$ points) with
below-chance GPS ($0.38$; cf.\ Appendix Table~\ref{tab:gold_reachability}).

The post-hoc replay on the full $n=230$ answer log raises KG-side coverage from
$97/223$ to $195/223$ and AUROC from $0.47$ to $0.76$
(cf.\ Appendix Table~\ref{tab:realmedqa_replay}); the same replay reaches $0.89$ on
the dense side.
RealMedQA is the calibrated case: it shows what retrieval-state scoring can
deliver on a clinical shared corpus, while the held-out rows above show the
generalisation limit.

\subsection{Concrete cases}
\label{sec:results_cases}

The aggregate AUROCs and route statistics summarise behaviour across questions;
the presence-lock-in case traced in Figure~\ref{fig:lockin_trace} makes the
mechanism concrete in scores. Answer-state uncertainty sits at its floor
($\mathrm{DSE}=0$, $\mathrm{SD\text{-}UQ}\approx 0$) and GPS stays low ($0.44$)
because the wrong entity is genuinely reachable, so only the evidence-state
signal fires ($\mathrm{SEU}=1.0$ from contradicting chunks).
The case shows that the graph can make a wrong reasoning chain visible even when
the scalar retrieval-state score reports low risk, provided the evidence-support
head is read alongside it; the full per-case audit readout is in
Appendix Table~\ref{tab:path_faithfulness_case}.

Absence lock-in is the same logic with a different face: the $48$ strict-clinical
silent errors leave no graph trace to inspect, visible only as the empty-retrieval
route. Presence hides a wrong answer inside a coherent trace; absence leaves no
trace at all, which is why the two variants need different flags.
A consolidated case gallery, with seven verified lock-in failures, four KG
successes, and per-case diagnostic readouts, is in
Appendix~\ref{sec:appendix_cases} (Table~\ref{tab:lockin_gallery}).

\subsection{Finding 4: A conjunctive gate yields high precision but low coverage}
\label{sec:results_composite}

\emph{Requiring answer-, evidence-, and retrieval-state checks to concur
certifies a small slice of answers at precision well above the base rate, at
the cost of low coverage.}
No single non-answer arm proved reliable enough to act alone.
The decision layer has to respect the difference between the objects being
measured, rather than collapse them into one score too early.
Empirically, the scores barely move together; that is the quantitative reason
to keep the decomposition visible.
Across the six-snapshot heatmap suite ($n=803$ pooled non-abstained KG rows),
answer-state and evidence-state dispersion are essentially uncorrelated
(Spearman $\rho=0.03$ for SD-UQ versus SEU; SEU versus GPS $\rho=-0.10$). The one
non-trivial coupling, $\rho=-0.27$ between SD-UQ and GPS, is largely a sign
artefact of GPS's risk orientation, not a universal risk law.
The three scores carry partly separate diagnostic signal, even if low correlation
alone cannot prove distinct latent causes; the pooled family-disagreement scatter
for the adaptive KG setting is in
Appendix Figure~\ref{fig:family_disagreement}.

A low answer-state score should not override a retrieval-state abstention, a
fragile graph path, or high evidence contradiction. A deliberately simple
percentile-rank composite can
capture this family disagreement (Appendix~\ref{sec:appendix_composite},
Table~\ref{tab:composite_audit_score}); here we propose a stricter, more
auditable policy.

\paragraph{A conjunctive low-risk audit rule.}
The most conservative use of the suite turns the decomposition into a single
gate, treating a response as low-risk only when all three checks concur: low
answer-state uncertainty (SD-UQ in the lower half of its dataset distribution),
non-contradictory evidence support ($\mathrm{SEU} \leq 0.5$), and defined,
present retrieval-state support ($\mathrm{GPS} \leq 0.5$).
The $\mathrm{SEU} \leq 0.5$ arm acts as a hallucination filter, not a relevance
filter: the threshold sits at the neutral point to catch \emph{active}
contradiction in the retrieved evidence, leaving provenance (the GPS arm) to
handle missing or neutral evidence. A neutral, all-$0.5$ verdict counts as a
pass.

The gate is narrow, admitting only $7.7\%$ of answers ($86$ of $1{,}117$), but
those it admits are correct $91.9\%$ of the time against a $69.7\%$ base rate.
This pooled figure spans real per-dataset heterogeneity (from $1.000$ on the
clinical calibration domain to $0.808$ on the larger HotpotQA-FullWiki
contribution, Table~\ref{tab:certificate_perds}), and is read here as a summary,
not a single operating point.
This is unlikely to be threshold overfitting: no gate cutoff is fit to
correctness labels, since SEU uses its neutral entailment point ($0.5$) and the
GPS gate the midpoint of the frozen graph-support risk scale ($0.5$). The GPS score itself
still depends on the RealMedQA-calibrated linking and decay hyperparameters
reported above. The only data-dependent gate threshold is the SD-UQ cutoff (the
per-dataset median, set to hold relative answer-state risk constant across
domains). Recomputing that median on a random half of each dataset and applying
the rule to the other half gives $91.8\%$ mean precision ($5$th--$95$th
percentile $86.7$--$97.4\%$ over $200$ splits) at
unchanged coverage (Appendix~\ref{sec:appendix_audit_generalisation}).

On the clinical target domain the rule selects $22\%$ of answers at $100\%$
\emph{automated-judge} precision ($48/48$).
This is an in-domain automated-label upper bound, not evidence of clinical
safety or readiness, and it needs human-expert validation before any clinical
use (Section~\ref{sec:ethics}).
The independent Llama-3.3-70B judge confirms all $48$ of these certified answers
with zero label flips, so the cell is robust to the judge family despite
RealMedQA's low dataset-wide $\kappa$
(Appendix~\ref{sec:appendix_rejudge_certificate}).
The same policy form is applied to every dataset
(Table~\ref{tab:certificate_perds}, the cross-dataset generalisation view), with
recall reported because a high-precision selection rule is trivial without it.
Excluding RealMedQA, its calibration domain, the rule still reaches $81.6\%$
pooled precision ($31/38$) on the five out-of-calibration datasets, so the
high-precision behaviour is not solely an in-domain artefact.

Coverage is structurally limited: the rule cannot select answers with no
linkable answer entity (binary, numeric, or label-space answers, where GPS
abstains by construction), so on those formats it defers rather than selects,
and a deployment would need a separate fallback.
On the adaptive RealMedQA run, SEU is largely at the $0.5$ neutral plateau
($74\%$ of rows), so the conjunctive rule there effectively operates on SD-UQ
and GPS alone.
The conjunction beats matched-coverage scalar gates: SD-UQ alone gives $81.4\%$
precision at the same coverage, and the percentile-rank combined score gives
$88.4\%$.
A dataset-stratified Mantel--Haenszel check gives the same direction (odds
ratio $3.39$, bootstrap 95\% CI $[1.64, 10.99]$; the per-dataset $2\times2$
cells are in Appendix Table~\ref{tab:mh_2x2}).

The rule is exploratory and domain-conditional: strongest on the clinical
target domain (partly an in-domain advantage, since GPS was calibrated there at
$\tau=0.60$, $\gamma=0.4$), thin-coverage on the harder multi-hop sets, and
undefined for label-space answers.
The combined audit score beats the best single-family score on only half of
the six datasets (cf.\ Table~\ref{tab:composite_audit_score}); the advantage is
dataset-dependent and should not be read as universal.

\begin{table}[t]
\centering
\footnotesize
\caption{%
  \textbf{The conjunctive audit rule is high-precision and low-coverage,
  strongest on the clinical domain.}
  Per-dataset behaviour on the adaptive KG runs.
  \emph{Recall} is the fraction of all \emph{correct} answers the rule selects;
  Wilson intervals expose the small denominators.
  Cells with fewer than roughly ten selected answers (HotpotQA, 2WikiMHQA,
  MuSiQue) are trace-level, not reliable per-dataset precision estimates.
  The rule presumes unselected answers are routed to ordinary confidence
  handling or review, not discarded.%
}
\label{tab:certificate_perds}
\setlength{\tabcolsep}{2.5pt}
\begin{tabular}{@{}lrrcr@{}}
\toprule
Dataset & Selected & Prec. & 95\% CI & Recall \\
\midrule
PubMedQA & 0/100 (0\%) & -- & -- & 0.00 \\
RealMedQA & 48/223 (22\%) & 1.000 & [0.926, 1.000] & 0.23 \\
HotpotQA & 6/238 (3\%) & 0.833 & [0.436, 0.970] & 0.04 \\
HotpotQA-FW & 26/218 (12\%) & 0.808 & [0.621, 0.915] & 0.14 \\
2WikiMHQA & 4/244 (2\%) & 1.000 & [0.510, 1.000] & 0.02 \\
\rowcolor{black!6}MuSiQue$^\dagger$ & 2/94 (2\%) & 0.500 & [0.095, 0.905] & 0.03 \\
\midrule
Pooled & 86/1117 (7.7\%) & 0.919 & [0.841, 0.960] & 0.10 \\
\bottomrule
\end{tabular}
\\[2pt]
{\scriptsize $^\dagger$Denominator too small to carry statistical weight; reported for completeness, not as evidence.}
\end{table}
Per-dataset behaviour exposes both cautionary cases: 2WikiMultiHopQA
selects only four answers ($2\%$ coverage) despite weak GPS ranking, while
MuSiQue selects two answers at only $50\%$ precision on a denominator too small
to weigh. Cells with fewer than roughly ten selections (HotpotQA,
2WikiMultiHopQA, MuSiQue) should be read as trace-level, not as reliable
per-dataset precision estimates.
The rule is a \emph{demonstration of certifiability}: the three diagnostic arms
can jointly certify a small, high-precision subset. It is not a
\emph{select-to-answer} policy that discards the rest. Its value is making the
decomposition actionable; coverage is the binding constraint, and per-domain
validation is mandatory before any deployment.
Coverage is bounded mainly by GPS definability and SEU neutrality, so
the refinements above (soft linking already; sentence-level entailment next)
are the most direct route to higher audit-rule coverage.
A simple learned alternative does not transfer better in this fixed-subset test:
leave-one-dataset-out logistic gating selects $7.5\%$ of answers at $85.7\%$
precision. With three raw standardised scalar features (SD-UQ, SEU, and
GPS-risk, not their binarised gate decisions) and substantial cross-dataset
shift, conjunctive gating remains the more auditable policy.
The dense substrate gives the same pattern, although this is only an internal
scalar-gate control, not a reimplementation of SURE-RAG or FRANQ: it asks
whether saved dense-side scalar signals can reproduce the high-precision
rejection frontier. A
learned logistic gate over all seven dense uncertainty scalars never attains the
lowest risk--excess--coverage and inverts below chance on the low-error
biomedical sets, while the conjunctive rule attains the lowest AUREC on three
of six datasets (including the clinical target) and the best single signal wins
the other three
(Appendix~\ref{sec:appendix_dense_frontier}).

\section{Conclusion}
\label{sec:conclusion}

Retrieval-state lock-in turns agreement into the wrong kind of reassurance.
When the retriever repeatedly returns the same empty or wrong state, more
samples need not test more possibilities; they may simply repeat the same
condition.
The confidence question changes from whether the answer is stable to which
object is stable: the answer surface, the retrieved evidence, or the retrieval
state itself.

We find that answer-state uncertainty remains a useful default, but a bounded
one.
In the deployable-policy runs, $42$--$59\%$ of errors already have zero observed
answer dispersion at $N{=}5$, so an answer-only method has no within-question
disagreement signal to recover.
Evidence-state diagnostics catch some of these failures by testing support in
the retrieved text.
GPS adds a graph-specific view that faithfully reports the retrieval state:
strong where retrieval exposes the gold entity (the clinical corpus), weak where
the policy does not (open-domain multi-hop).
That off-domain weakness survives re-calibration
(Appendix~\ref{sec:appendix_gps_sensitivity}), so it reflects the retrieval
policy rather than a tuning gap, and its value there lies in the auditable trace
it exposes, not a universal ranking.
Agreement, support, and retrieval grounding are different observables; a single
confidence scalar hides that difference.

In practice, this means reporting answer-, evidence-, and retrieval-state
signals separately, and calling an answer low-risk only when all three agree:
low answer uncertainty, adequate evidence support, and present graph support.
In \system{}, this conjunctive rule selects a small subset at $91.9\%$ pooled
precision against a $69.7\%$ base rate, reaching near-ceiling automated-label
precision ($100\%$ under the \emph{automated judge}) on the clinical target
domain at $22\%$ coverage.
That clinical cell is an in-domain upper bound, not a safety claim, and still
needs human-expert validation (Section~\ref{sec:ethics}).
A learned gate does not transfer better here
(Section~\ref{sec:results_composite}, Appendix~\ref{sec:appendix_dense_frontier});
a head-to-head against external baselines (SURE-RAG, FRANQ) remains an
informative next test.
Unselected answers are not thereby wrong, only uncertified: the answers outside
the $7.7\%$ certified subset are neither certified nor rejected by this audit
rule, and in the reported runs they retain the system's pooled base accuracy
($69.7\%$).
When overall trust is low (the rule does not certify the answer and the
evidence- or retrieval-state diagnostics also disagree), the case should instead
be escalated to human review or a retrieval-perturbation retry.

The case the diagnostics leave open is the hardest presence lock-in: the system
retrieves a coherent but wrong neighbourhood that locally supports the answer, so
even the evidence- and retrieval-state checks stay calm.
That is the setting where relation-level path faithfulness matters most.
Future systems should test not only whether an answer entity is reachable, but
whether the path uses the relation the question actually asks for.

The evidence is not equally strong throughout.
The observed silent-error rates and the answer-state collapse under strict
stress are the firmest pieces; the route decomposition gives moderate evidence
for absence lock-in; presence lock-in rests on populated-route cases and a
smaller set of confirmed examples; and reliable detection of a coherent wrong
neighbourhood when both SEU and GPS are calm remains open.
The silent-error phenomenon itself is substrate-general (dense retrieval shows
it too), while the full three-family audit is easiest on an inspectable graph,
where the retrieval state is a queryable object.
So the contribution is not that we solve lock-in, but that we make it nameable,
measurable, and diagnosable: retrieval-state lock-in does not make RAG
uncertainty hopeless, it makes confidence object-specific. Once the answer,
evidence, and retrieval state are separated, stable agreement stops being a
blanket certificate and becomes a diagnostic clue.

\subsection{Limitations}
\label{sec:limitations}

The strict graph-only setting is a mechanism probe, not a deployable policy or
a one-component ablation.
It changes fusion, reranking, decomposition, and fallback together, so it
demonstrates a regime in which answer-state uncertainty fails rather than
estimating production prevalence.

Each representative score is local or conditional.
GPS is strongest on the calibrated clinical corpus, weak on several open-domain
multi-hop suites despite the sensitivity replay
(Appendix~\ref{sec:appendix_gps_sensitivity}), and leaves yes/no and other
label-space answers outside its design; the retrieval-state family is best read
as an observable class (anchors, paths, subgraphs, routes) of which GPS is one
scalar instantiation.
SEU tests answer--evidence consistency, not whether the model used the
evidence; its neutral plateau can make empty-context failures easier to rank
than populated presence-lock-in failures, and while the domain-NLI ablation
(Appendix~\ref{sec:appendix_seu_nli_ablation}) shows part of that plateau is a
general-domain-NLI artefact, the neutral rate does not fall under a
biomedical-aware judge, so a retriever-side component remains.
SD-UQ depends on the response embedding model, and all reported numbers use a
single encoder.

The empirical scale supports the central contrasts but not fine leaderboard
claims between neighbouring metrics.
Correctness labels come from an automated semantic judge and survive an
independent Llama-family re-judge, but human or domain-expert adjudication would
be needed before treating silent errors as clinical risk.
The study also uses fixed small subsets, one KG-RAG framework, one generation
model (GPT-4o-mini), and one primary random seed for subset construction.
These constraints bound the paper as a diagnostic methodology and fixed-system
empirical audit; a cross-model survey of lock-in prevalence would require
additional retrievers and generators.

The dense--KG accuracy comparison is measured on curated benchmark corpora,
which flatter dense retrieval: a retrieval-only dilution probe on the shared
HotpotQA-FullWiki corpus shows gold-passage recall@10 falling from $1.00$ at a
curated $10$-candidate pool to $0.55$ at the full $2{,}489$-passage corpus
(Appendix~\ref{sec:appendix_dilution}).
The near-match is a property of the curated regime, not a guarantee at deployment
scale, where an auditable retrieval state matters more rather than less.

The retrieval-state diagnostics are graph-native, though the silent-error
phenomenon they target is not: dense retrieval shows it too.

\subsection{Future Work}

Extending the same decomposition to dense-RAG retrieval states, for example
through citation-path or passage-link plausibility, is the most direct route to
broader generality.
The decomposition itself is not tied to \system{}: the answer- and
evidence-state arms read only sampled answers and retrieved text, so they port
with minimal adaptation to other KG-RAG systems such as GraphRAG or HippoRAG, while only the
retrieval-state arm must be re-targeted to each system's trace.
A second direction is agentic and iterative-retrieval RAG, where the per-step
diagnostics largely carry over and can flag a state that successive queries
deepen rather than break.
The decomposition also points to a control policy: when answer samples agree
but the evidence- or retrieval-state diagnostics disagree, a system could
perturb the retrieval state (for example by masking the dominant intermediate
anchor) and regenerate: if the answer changes, it was anchored to that state;
if not, the evidence was stable across the perturbation.
Which anchor to mask, how many alternatives to test, and when to stop are open
questions, so we report the diagnostics here rather than a validated controller.

\subsection{Ethical and Safety Considerations}
\label{sec:ethics}

Nothing in these results licenses autonomous clinical answering.
The audit rule is deliberately high precision and low coverage; unselected
answers must not be treated as audited.
The three diagnostic arms differ in cost: SD-UQ is embedding-only, while SEU and GPS
add NLI and graph queries (Appendix Table~\ref{tab:compute_cost}), so a
latency-sensitive deployment can cascade, screening with SD-UQ and invoking the
evidence- and retrieval-state arms only on borderline cases.
Lock-in also has an adversarial analogue: presence lock-in is the benign
counterpart of retrieval-corpus poisoning, where a corpus or query is
deliberately shaped to anchor retrieval in a plausible but wrong neighbourhood
\citep{zou2025poisonedrag,zhong2023poisoning}.
Retrieval diversity constraints and anchor-masking controls are plausible
defences, but they remain future work.
Adversarial robustness lies outside this evaluation; the diagnostic
decomposition is still a prerequisite for it, since a failure mode that cannot
be observed cannot be defended against.

\begin{center}\fbox{\begin{minipage}{0.95\columnwidth}\small\textbf{Takeaway for practitioners.} Deployments should not gate reliability on answer-agreement alone. Monitor three distinct objects rather than a single confidence scalar: answer dispersion (SD-UQ), evidence contradiction (SEU), and the retrieval trace (GPS). When answers agree but SEU exceeds its neutral point, GPS abstains, or GPS reports weak support, trigger a retrieval perturbation or route the case to human review.\end{minipage}}\end{center}

\bibliographystyle{plainnat}
\bibliography{references}

\appendix
\onecolumn
\section{Residual-Mass Bound for Stabilised Retrieval}
\label{sec:appendix_residual_bound}

The collapsed-context approximation in Section~\ref{sec:background} can be
read as a simple mixture argument.
Let $\alpha = p(c^\star \mid q)$ and define the residual context mixture
\[
  \pi_{\mathrm{res}}(r)
  =
  \frac{1}{1-\alpha}
  \sum_{c \in \mathcal{C}(q)\setminus\{c^\star\}}
  p(r \mid q,c)\,p(c\mid q),
\]
whenever $\alpha < 1$.
Then
\[
  p(r\mid q)
  =
  \alpha\,p(r\mid q,c^\star)
  +
  (1-\alpha)\,\pi_{\mathrm{res}}(r),
\]
and therefore
\[
  \bigl\|p(r\mid q)-p(r\mid q,c^\star)\bigr\|_{\mathrm{TV}}
  =
  (1-\alpha)
  \bigl\|\pi_{\mathrm{res}}(r)-p(r\mid q,c^\star)\bigr\|_{\mathrm{TV}}
  \leq 1-\alpha.
\]
The bound is not an estimator; it only states when the stylised approximation is
meaningful: the residual routing mass must be small.

\section{KG Construction and System Configuration}
\label{sec:appendix_construction}

\subsection{Knowledge Graph Construction Pipeline}
\label{sec:appendix_kg}

The KG is built passage-wise with chunk-level provenance, so every triple and
path traces back to the text that licensed it.

\paragraph{Text chunking.}
Input documents are split into overlapping passages using a token-aware
recursive splitter (tiktoken-based, with a character-based fallback when the
tokenizer is unavailable) with chunk size $L = 1{,}500$ tokens and overlap
$\delta = 200$ tokens.
Each chunk receives an embedding at construction time using
\texttt{all-MiniLM-L6-v2} (384 dimensions) from Sentence Transformers
\citep{reimers2019sentencebert}.
Chunks are stored as \textsc{Chunk} nodes in Neo4j with a SHA-1 content
hash as identifier and a \textsc{Part\_Of} edge to their parent
\textsc{Document} node.

\paragraph{Entity and relation extraction.}
Entities and typed relations are extracted from each chunk by a prompted
GPT-4o-mini call \citep{openai2024gpt4omini} returning a JSON object with an
\texttt{entities} array (fields \texttt{id}, \texttt{type}, \texttt{name},
\texttt{description}) and a \texttt{relationships} array (\texttt{source},
\texttt{target}, \texttt{type}, \texttt{description}).
Two extraction modes are supported:

\begin{enumerate}
  \item \textbf{Ontology-guided extraction.} A provided domain schema is parsed
    from JSON or OWL/RDF into one typed internal representation; the extraction
    prompt then carries the ontology entity types (with descriptions and typed
    properties) and relationship types (with domain/range/cardinality
    constraints). After extraction, entity types are normalised against ontology
    class labels by embedding cosine similarity (threshold $\tau = 0.50$), with
    substring matching and keyword heuristics as fallback, and relationship
    labels are canonicalised to the closest schema-compatible type by
    domain/range compatibility and fuzzy matching. Biomedical and clinical types
    include \textsc{Disease}, \textsc{Treatment}, \textsc{Symptom},
    \textsc{Biomarker}, and \textsc{Anatomy}.

  \item \textbf{Open extraction.} Without a type schema, the LLM extracts
    entities and relations freely.
    Used for the open-domain datasets (HotpotQA, HotpotQA FullWiki,
    2WikiMultiHopQA, and MuSiQue).
\end{enumerate}

The system does \emph{not} induce an ontology from the documents; it loads a
user-provided schema and uses it to guide extraction, typing, and relationship
normalisation.

\paragraph{Neo4j schema.}
Entities are \textsc{Entity} nodes with compound labels
\texttt{:\_\_Entity\_\_:Type} (e.g.\ \texttt{:\_\_Entity\_\_:Disease}), each
storing an embedding of its name and description and linked to originating
chunks via \textsc{Mentioned\_In} edges.
Extracted relation types (e.g.\ \textsc{Treats}, \textsc{Causes},
\textsc{Has\_Complication}) are typed edges between entity nodes.
Three indexes are maintained: vector indexes over chunk and entity embeddings,
and a full-text keyword index over chunk content.

\paragraph{KG scope.}
Each dataset's KG is a named graph (a \texttt{kgName} attribute on all nodes),
so multiple datasets coexist in one Neo4j instance without interference.
The experiment pipeline supports two build-scope modes selected by
\texttt{--dataset-kg-scope}: \textbf{evaluation\_subset} (default)
builds the KG from the same question slice being evaluated, which
minimises KG size and keeps retrieval and knowledge in the same closed
corpus; \textbf{full\_dataset} builds from all available passages in
the normalised dataset before evaluating the requested subset.
Unless explicitly noted as a diagnostic, the reported fixed-snapshot runs use
\texttt{evaluation\_subset} so that the KG and dense retriever see the same
closed corpus slice.
This improves experimental control, but it also suppresses some of the
graph-incompleteness and corpus-drift failures that would matter more in a
deployment-scale setting.

\paragraph{Relationship verification.}
Each typed relationship is verified against the chunks that contributed at least
one of its endpoint entities (tracked via provenance positions stored at build
time), so verification NLI runs only over the source chunks, not the full
corpus; this reduces false-positive edges from entity-name collisions across
unrelated passages.

\paragraph{Builder profiles and graph repair.}
The runner exposes three KG-builder profiles. \texttt{full} (used for all
full-metrics runs unless stated otherwise) enables anchor-constrained
extraction, self-reflection over missing entities, anchor-coverage
supplementation, cross-passage relation recovery, low-confidence triple
reverification, soft entity linking, fragmentation repair, optional graph
summaries, and claim extraction; on biomedical and clinical datasets it also
attempts UMLS-backed normalisation when the local SciSpaCy/UMLS stack is
available, continuing without UMLS metadata otherwise rather than failing.
\texttt{lightweight} disables the expensive anchor, reflection, and
cross-passage extras for quick accuracy-only sweeps; \texttt{auto} selects it
only for those sweeps and otherwise resolves to \texttt{full}.
Soft linking and fragmentation repair are deliberately conservative: the builder
canonicalises obvious aliases and near-duplicates, then adds bridge records only
for high-similarity fragments rather than freely rewiring the graph.
Claim extraction and graph summaries are stored as additional artefacts that
enrich inspection, but the eight headline uncertainty measures do not depend on
them as separate scoring heads.

\subsection{System Configuration}
\label{sec:appendix_config}

Table~\ref{tab:config} summarises the hyperparameters used across all
experiments.

\begin{table*}[t]
\centering
\small
\caption{System configuration for all experiments.
  \label{tab:config}}
\begin{tabular}{lll}
\toprule
Parameter & Vanilla RAG & KG-RAG \\
\midrule
Chunk size (tokens)           & 1,500 & 1,500 \\
Chunk overlap (tokens)        & 200   & 200   \\
Embedding model               & \multicolumn{2}{c}{\texttt{sentence-transformers/all-MiniLM-L6-v2} (via sentence-transformers, 384-dimensional)} \\
Top-$k$ chunks                & 10    & 10    \\
Adjacent chunk expansion      & yes (pos $\pm 1$) & -- \\
Similarity threshold          & 0.10  & 0.10  \\
Entity matching               & --    & \shortstack[l]{mention ANN ($\geq 0.72$) when query entities are extracted;\\otherwise query ANN ($\geq 0.55$); plus exact/synonym matching} \\
Chunk sufficiency threshold   & --    & 2 chunks \\
Graph scoring                 & --    & \shortstack[l]{local PPR over the retrieved entity subgraph;\\fallback to hop prior when no graph edges are available} \\
Graph expansion hops          & --    & 2 (4 for MuSiQue) \\
Iterative local hop cap      & --    & $\min(h, 3)$ per sub-question \\
Neighbours per seed entity    & --    & 30    \\
Max seed entities in prompt   & --    & 15    \\
Max neighbour entities        & --    & 10    \\
Max direct relationships      & --    & 25    \\
Iterative decomposition       & --    & enabled when the hop target is $h \geq 2$ \\
Fallback cascade              & vector & \shortstack[l]{entity-first $\to$ graph expansion $\to$ vector $\to$ text-keyword} \\
LLM (generation)              & \multicolumn{2}{c}{GPT-4o-mini} \\
Samples per question ($N$)    & \multicolumn{2}{c}{5} \\
\bottomrule
\end{tabular}
\end{table*}

\paragraph{Configuration note.}
All reported runs use the $k=10$ and similarity-threshold $0.10$ entries of
Table~\ref{tab:config}, under the \texttt{final\_pair} retrieval-study profile:
vanilla RAG runs only \texttt{dense\_floor} and KG-RAG only
\texttt{kg\_entity\_first}, avoiding the earlier four-way crossing of dense and
graph variants.
The strict RealMedQA stress test uses the separate \texttt{strict\_entity} profile,
which disables query fusion, late interaction, dense fallback, vector
augmentation, decomposition, and the KG-RAG runtime guardrail.
Because iterative decomposition is itself a source of retrieval variability,
Equation~\eqref{eq:collapsed} is a stylised limit, not a claim that every KG-RAG
call is deterministic; the stabilised regime is strongest when decomposition and
entity anchors converge to the same neighbourhood across repeated calls.
The released manifests record the reranking, fallback, and query-fusion flags,
but this paper does not sweep $k$, similarity threshold, or routing flags as
independent factors.

\section{Implementation Details and Reproducibility}
\label{sec:appendix_reproducibility}

\paragraph{Question sampling.}
PubMedQA and MuSiQue retain the original fixed $n=100$ subsets; RealMedQA is
reported on its full $n=230$ evaluable set (small enough to run in full and the
main shared-corpus clinical diagnostic); HotpotQA, HotpotQA FullWiki, and
2WikiMultiHopQA use fixed $n=250$ subsets drawn with the same seed to reduce
denominator volatility.
Clean-accuracy analyses can have smaller effective denominators after excluding
provider-side generation failures; those answered counts are in the run
artefacts and summarised in the main text.
Subset sampling is deterministic: given a dataset, sample size $n$, and seed $s$,
the pipeline draws a fixed set of question IDs, persists it, and reuses it on
later runs unless a new subset is requested, keeping KG construction and
evaluation aligned. Exact IDs are stored in the run artefacts and subset
manifests.

\paragraph{KG construction LLM.}
Reported KG builds use GPT-4o-mini (via OpenRouter in the latest local runs) for
extraction, temperature-locked at $T=0.0$ even when answer sampling uses a higher
temperature.
Biomedical and clinical datasets (PubMedQA, RealMedQA) use ontology-guided
extraction with a domain type schema (Disease, Treatment, Symptom, Biomarker,
Anatomy); open-domain datasets (HotpotQA, HotpotQA FullWiki, 2WikiMultiHopQA,
MuSiQue) use unconstrained open extraction.

\paragraph{Training and fine-tuning configuration.}
No task-specific training, fine-tuning, or gradient-based calibration is
performed: the generation, judge, NLI, and embedding models are all frozen
off-the-shelf components, and the only object that changes across experiments is
the dataset-specific KG.
Unless a run sets \texttt{--judge-provider} or \texttt{--judge-model}, the
correctness judge uses the same GPT-4o-mini backend as generation.
Uncertainty samples use $T=1.0$ with $N=5$ responses per question; accuracy
generation and KG extraction use $T=0.0$.
Final-stage retrieval selection uses retrieval temperature $0.0$, with a
shortlist factor of $4$ retained in the code for stochastic sweeps.

\paragraph{NLI model.}
DSE and the P(True)-style proxy cluster responses by bidirectional NLI
entailment using \texttt{microsoft/deberta-large-mnli} \citep{he2021deberta}, run
locally; SelfCheckGPT uses \texttt{roberta-large-mnli} for pairwise contradiction
scoring.
Entailment decisions are label-based (argmax over the three classes): two
responses share a cluster when at least one direction predicts entailment and
neither predicts contradiction.

\paragraph{Embeddings for geometric metrics.}
VN-Entropy, SD-UQ, and SRE-UQ embed sampled responses with
\texttt{all-MiniLM-L6-v2} (384-d), the same model used for chunk and entity
indexing; response embeddings are computed at evaluation time, not cached.

\paragraph{Metric hyperparameters.}
GPS: maximum path length $h = 2$ hops (4 for MuSiQue), matching the
retrieval hop depth; soft answer-entity linking with
\texttt{all-MiniLM-L6-v2} cosine threshold $\tau = 0.60$ and distance decay
$\gamma = 0.4$, both selected on the RealMedQA development run over a small
grid and applied frozen to all other datasets.
Reported GPS values come from a post-hoc replay on the saved answer logs and the
persistent dataset KGs: no answer generation, document retrieval, or LLM call is
rerun. For the primary GPS numbers the stored linked entities and path lengths
are reused; the coverage-raising and gold-reachability replays instead recompute
entity alignment and path support against the persistent KG. Replay artefacts are
recorded under \texttt{results/analyses/}.
SD-UQ: $k_{\max} = \min(N{-}1, 8)$ principal components
(effective $k_{\max}=4$ for the reported $N=5$ samples),
$\varepsilon = 10^{-12}$.

\paragraph{Compute environment.}
All experiments ran on a MacBook Pro (Apple~M4, macOS~15.5 ARM64, 14 cores,
52\,GB RAM), with NLI and embeddings on CPU.
Because KG extraction and generation call GPT-4o-mini through an API, wall-clock
time is dominated by remote calls: roughly 8--12 hours for the core-suite pass,
longer for the larger shared-corpus and ablation runs (the HotpotQA FullWiki
$n=250$ stress run took about a day).
These timings are specific to the laptop-and-API setup and would improve with
local GPU serving.

\paragraph{Effective denominators and bootstrap intervals.}
The main text reports point estimates because the analysis is organised around
family-level behaviour and qualitative traces, not hypothesis tests over small
AUROC gaps.
To make the scale explicit, the appendix reports effective answered counts (after
provider-failure filtering), usable GPS denominators (after fallback-abstention
filtering), and percentile-bootstrap 95\% intervals for the headline AUROCs.
All intervals use $B=2000$ resamples over answered questions within each dataset
and system, at a fixed seed of 42.
For HotpotQA and 2WikiMultiHopQA the remaining GPS denominators are small enough
to make the corresponding AUROCs descriptive.

\begin{table*}[t]
\centering
\small
\caption{Summary of reported run artefacts: dataset, subset seed, sample size,
evaluation mode, and KG build scope for each reported snapshot.
All runs use GPT-4o-mini as the generation model and $N=5$ uncertainty samples.
\label{tab:run_artifacts}}
\begin{tabular}{@{}llcp{4.0cm}c@{}}
\toprule
Dataset & Subset seed & $n$ & Evaluation mode & KG build scope \\
\midrule
PubMedQA           & 42 & 100 & full\_metrics & evaluation\_subset \\
RealMedQA (adaptive) & 42 & 230 & full\_metrics & evaluation\_subset \\
RealMedQA (strict)   & 42 & 230 & full\_metrics & evaluation\_subset \\
HotpotQA           & 42 & 250 & full\_metrics & evaluation\_subset \\
HotpotQA FullWiki  & 42 & 250 & full\_metrics & evaluation\_subset \\
2WikiMultiHopQA (adaptive) & 42 & 250 & full\_metrics & evaluation\_subset \\
2WikiMultiHopQA (strict)   & 42 & 250 & full\_metrics & evaluation\_subset \\
MuSiQue            & 42 & 100 & full\_metrics & evaluation\_subset \\
\bottomrule
\end{tabular}
\end{table*}

\begin{table*}[t]
\centering
\small
\caption{Answered and failed rows in the reported saved runs.
Failures are provider- or pipeline-side generation failures and are excluded
from clean accuracy but retained in raw accounting.
The RealMedQA adaptive--strict comparison is analysed on the $196$ questions
answered by both KG policies.
\label{tab:generation_failures}}
\setlength{\tabcolsep}{4.5pt}
\begin{tabular}{@{}lrrrrrr@{}}
\toprule
Dataset & Dense ans. & Dense fail & KG ans. & KG fail & Strict ans. & Strict fail \\
\midrule
PubMedQA & 100/100 & 0 & 100/100 & 0 & -- & -- \\
RealMedQA & 219/230 & 11 & 223/230 & 7 & 200/230 & 30 \\
HotpotQA & 226/250 & 24 & 238/250 & 12 & -- & -- \\
HotpotQA FullWiki & 208/250 & 42 & 218/250 & 32 & -- & -- \\
2WikiMHQA & 212/250 & 38 & 244/250 & 6 & 245/250 & 5 \\
MuSiQue & 90/100 & 10 & 94/100 & 6 & -- & -- \\
\bottomrule
\end{tabular}
\end{table*}

\begin{table*}[t]
\centering
\small
\caption{Effective denominators for the current reported runs.
``Van answered'' and ``KG answered'' count questions remaining after
generation-failure filtering.
``GPS usable'' counts rows remaining after GPS abstention filtering.
The HotpotQA FullWiki stress snapshot is also expanded in
Table~\ref{tab:hotpot_fullwiki_stress}.
GPS columns use the final soft-linked, depth-matched estimator replayed
on the saved runs (Table~\ref{tab:realmedqa_replay} reports the corresponding
coverage before/after); RealMedQA is the GPS calibration domain.
Retrieval-state AUROC is only shown where the resulting denominator leaves an
interpretable ranking problem.
\label{tab:effective_denominators}}
\begin{tabular}{@{}lcccc@{}}
\toprule
Dataset & Van answered & KG answered & GPS usable & GPS AUROC [95\% CI] \\
\midrule
PubMedQA & 100 & 100 & -- & -- \\
RealMedQA & 219 & 223 & 195 & 0.76 [0.59, 0.90] \\
HotpotQA & 226 & 238 & 158 & 0.51 [0.42, 0.61] \\
HotpotQA FullWiki & 208 & 218 & 185 & 0.54 [0.45, 0.63] \\
2WikiMHQA & 212 & 244 & 182 & 0.38 [0.31, 0.47] \\
MuSiQue & 90 & 94 & 83 & 0.68 [0.56, 0.80] \\
\bottomrule
\end{tabular}
\end{table*}

Table~\ref{tab:gps_reliability_bins} gives a descriptive reliability view: the
clinical calibration domain has the expected monotone shape, several open-domain
rows do not, which is why GPS is treated as a conditional graph-support
diagnostic rather than a calibrated probability.

\begin{table*}[t]
\centering
\small
\caption{%
  \textbf{GPS reliability bins.}
  GPS abstention and fixed-bin empirical error rates on adaptive KG runs.
  The three right columns report wrong/total rates within GPS-risk bins; lower
  GPS means stronger local graph support.
  This is a descriptive reliability check, not an expected-calibration-error
  (ECE) calculation, because GPS is a graph-support score rather than a
  calibrated probability.%
}
\label{tab:gps_reliability_bins}
\setlength{\tabcolsep}{4pt}
\begin{tabular}{@{}lrrrrrr@{}}
\toprule
Dataset & Answered & Abstained & Usable & GPS $<.33$ & $.33$--$.67$ & GPS $>.67$ \\
\midrule
PubMedQA & 100 & 100 (100.0\%) & 0 & -- & -- & -- \\
RealMedQA & 223 & 28 (12.6\%) & 195 & 0.0\% (0/39) & 4.4\% (5/114) & 14.3\% (6/42) \\
HotpotQA & 238 & 80 (33.6\%) & 158 & 44.4\% (4/9) & 38.1\% (16/42) & 33.6\% (36/107) \\
HotpotQA FullWiki & 218 & 33 (15.1\%) & 185 & 30.0\% (21/70) & 30.8\% (16/52) & 36.5\% (23/63) \\
2WikiMHQA & 244 & 62 (25.4\%) & 182 & 33.3\% (6/18) & 50.0\% (10/20) & 27.8\% (40/144) \\
MuSiQue & 94 & 11 (11.7\%) & 83 & 33.3\% (3/9) & 50.0\% (10/20) & 70.4\% (38/54) \\
\bottomrule
\end{tabular}
\end{table*}

\begin{table*}[t]
\centering
\small
\caption{Mean pairwise retrieval-overlap summaries from the saved
reported runs.
Overlap is the mean Jaccard similarity of retrieved chunk-text sets across
the $N=5$ uncertainty-sampling calls for each question, averaged over the
run.
The current artefacts log chunk overlap at run level; they do not yet retain
per-question entity-overlap or path-overlap fields.
The strict graph-only rows are mechanism stress tests, not deployment
policies.
\label{tab:retrieval_overlap}}
\setlength{\tabcolsep}{4pt}
\begin{tabular}{@{}lcccp{5.1cm}@{}}
\toprule
Dataset & Dense & Adaptive KG & Strict KG & Reading \\
\midrule
PubMedQA & 1.000 & 0.994 & -- &
Single-abstract control; both retrievers are nearly deterministic. \\
RealMedQA & 0.952 & 0.558 & 0.661 &
Strict KG concentrates the graph context relative to adaptive KG, but dense
retrieval is also highly stable. \\
HotpotQA & 0.904 & 0.677 & -- &
Both systems retrieve stable chunk sets; the KG trace adds provenance but not
an accuracy win. \\
HotpotQA FullWiki & 0.832 & 0.516 & -- &
The shared-corpus stress run introduces more KG route and fallback variation. \\
2WikiMHQA & 0.840 & 0.802 & 0.540 &
Adaptive KG is close to dense at chunk level; strict graph-only failures are
more visible in hop-wise output collapse than in aggregate chunk overlap. \\
MuSiQue & 0.900 & 0.684 & -- &
Deep chains are stable enough for audit, but early wrong anchors still damage
answer quality. \\
\bottomrule
\end{tabular}
\end{table*}

\begin{figure*}[t]
\centering
\includegraphics[width=0.92\linewidth]{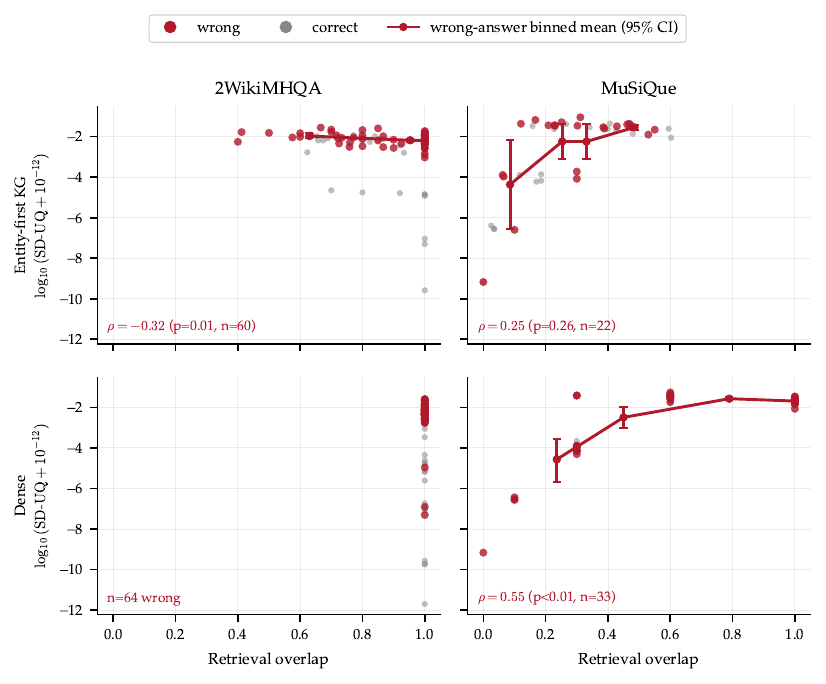}
\caption{%
  \textbf{Full per-dataset, per-system view of the archived
  concentration--dispersion traces}, with per-dataset Spearman
  correlations among wrong answers and binned means with bootstrap $95\%$
  CIs.
  The KG-side coupling is carried by the bridge-style 2WikiMHQA trace, while
  MuSiQue's small wrong set trends positive. These details keep
  the pooled $\rho=-0.33$ from being read as a between-dataset artefact or as a
  universal law.%
}
\label{fig:concentration_facets}
\end{figure*}

\begin{table*}[t]
\centering
\small
\caption{Selective risk at fixed coverage on the KG side: error rate among
the accepted questions when the most uncertain $20\%$ or $10\%$ are
abstained, using a single uncertainty signal as the gate.
``Base'' is the no-abstention error rate.
Under the adaptive policy, gating on SD-UQ alone is the strongest simple
policy; SEU gating is mediocre there because $25$--$74\%$ of adaptive rows
sit at the all-neutral $\mathrm{SEU}=0.5$ plateau, so the $80\%$ threshold
falls inside a tie mass.
Under the strict clinical stress test the ordering reverses: SD-UQ gating is no
better than accepting everything, while SEU gating removes roughly one in
five residual errors at $80\%$ coverage.
The 2WikiMultiHopQA strict run is the regime where no scalar gate helps,
consistent with the dataset-dependent survival reported in
Section~\ref{sec:results_grounding}.
\label{tab:operating_points}}
\setlength{\tabcolsep}{4pt}
\begin{tabular}{@{}lcccccc@{}}
\toprule
Run & $n$ & Base err. & SD@90\% & SD@80\% & SEU@80\% & Combined@80\% \\
\midrule
PubMedQA adaptive & 100 & 0.270 & 0.222 & 0.175 & 0.243 & 0.200 \\
RealMedQA adaptive & 223 & 0.054 & 0.030 & 0.022 & 0.054 & 0.034 \\
HotpotQA adaptive & 238 & 0.395 & 0.393 & 0.384 & 0.405 & 0.395 \\
HotpotQA FullWiki adaptive & 218 & 0.335 & 0.296 & 0.282 & 0.312 & 0.293 \\
2WikiMHQA adaptive & 244 & 0.311 & 0.282 & 0.256 & 0.261 & 0.267 \\
MuSiQue adaptive & 94 & 0.606 & 0.600 & 0.573 & 0.547 & 0.533 \\
\midrule
RealMedQA strict & 200 & 0.330 & 0.328 & 0.338 & 0.269 & 0.300 \\
2WikiMHQA strict & 245 & 0.592 & 0.600 & 0.633 & 0.668 & 0.617 \\
\bottomrule
\end{tabular}
\\[2pt]
{\footnotesize Adaptive SEU cells are means over $200$ randomised tie-breaking
draws; the spread across draws is at most $0.006$.}
\end{table*}

\begin{table}[t]
\centering
\small
\caption{Silent-failure sensitivity ladder: pooled fraction of wrong answers
classified as silent under progressively relaxed definitions of answer-state
dispersion.
With $N=5$ samples DSE is quantised, so $\mathrm{DSE} \leq 0.51$ admits at
most one dissenting sample (a 4-of-5 agreement gives
$\mathrm{DSE}=0.5004$).
The headline definition in Table~\ref{tab:silent_failures} is the most
conservative rung.
\label{tab:silent_sensitivity}}
\setlength{\tabcolsep}{4pt}
\begin{tabular}{@{}lccc@{}}
\toprule
Silent-error definition & Dense & Adaptive KG & Strict KG \\
\midrule
DSE $=0$ and SD-UQ at floor (headline) & 0.59 & 0.42 & 0.84 \\
DSE $=0$, any SD-UQ (unanimous)        & 0.67 & 0.52 & 0.88 \\
DSE $\leq 0.51$ ($\leq$1 dissenting sample) & 0.84 & 0.69 & 0.92 \\
\bottomrule
\end{tabular}
\end{table}

\begin{table}[t]
\centering
\small
\caption{Verbalised-confidence baseline (original P(True) protocol): the
generator backbone is asked once per saved answer for the probability that
the answer is correct, and $1-p$ is scored as uncertainty against the
paper's correctness labels.
No retrieval or generation is rerun.
AUROC treats incorrect answers as the positive class; SD-UQ columns repeat
the sampled-dispersion values for comparison.
The probe is competitive on the HotpotQA variants, weaker than SD-UQ on
RealMedQA, near chance on bridge-style 2WikiMultiHopQA under both policies,
and, unlike sampled dispersion, survives the strict clinical stress test.
\label{tab:verbalized_confidence}}
\setlength{\tabcolsep}{4.5pt}
\begin{tabular}{@{}lcc@{}}
\toprule
Run & Verbalised AUROC & SD-UQ AUROC \\
\midrule
PubMedQA adaptive & 0.55 & 0.70 \\
RealMedQA adaptive & 0.65 & 0.81 \\
HotpotQA adaptive & 0.72 & 0.64 \\
HotpotQA FullWiki adaptive & 0.77 & 0.67 \\
2WikiMHQA adaptive & 0.49 & 0.69 \\
MuSiQue adaptive & 0.65 & 0.63 \\
\midrule
RealMedQA strict & 0.89 & 0.23 \\
2WikiMHQA strict & 0.58 & 0.52 \\
\midrule
Pooled adaptive KG & 0.69 & -- \\
\bottomrule
\end{tabular}
\end{table}

\begin{table}[t]
\centering
\small
\caption{Mean marginal compute time per question for each diagnostic on the
KG side, averaged over the six headline runs (MacBook Pro M4; NLI on CPU).
Times are \emph{marginal}: they exclude the shared $N=5$ answer-sampling
calls that every answer-state estimator requires (the dominant
latency and token cost) and the cached NLI clustering shared by the
entropy-family scores.
GPS varies with graph depth (its maximum, $12.3$\,s, is the $h=4$ MuSiQue
configuration; the clinical graph costs $0.06$\,s).
\label{tab:compute_cost}}
\setlength{\tabcolsep}{5pt}
\begin{tabular}{@{}llr@{}}
\toprule
Diagnostic & Extra inputs & Mean time (s) \\
\midrule
DSE / P(True) proxy & cached NLI clusters & $<10^{-4}$ \\
SelfCheckGPT & $O(N^2)$ NLI calls & 1.68 \\
SRE-UQ & embeddings & 0.040 \\
VN-Entropy & embeddings & 0.007 \\
SD-UQ & embeddings & 0.012 \\
SEU & $K$ NLI calls & 1.65 \\
GPS & graph queries & 2.43 (0.06--12.3) \\
\bottomrule
\end{tabular}
\end{table}

\begin{table*}[t]
\centering
\small
\caption{HotpotQA FullWiki $n=250$ shared-corpus stress run.
Vanilla uses dense retrieval; KG-RAG uses the reported entity-first policy
with fallback.
Clean accuracy excludes provider-side generation failures.
GPS AUROC is reported only for KG-RAG because GPS is a graph-state diagnostic;
the GPS null rate is the fraction of answered KG rows for which the metric is
unavailable or falls back.
\label{tab:hotpot_fullwiki_stress}}
\setlength{\tabcolsep}{4pt}
\begin{tabular}{@{}lccccccccc@{}}
\toprule
Policy & Ans. & Clean acc. & Raw acc. & DSE & SRE-UQ & VN-Ent. & SD-UQ & SEU & GPS / null \\
\midrule
Dense vanilla & 208 & 0.721 & 0.600 & 0.650 & 0.630 & 0.666 & 0.665 & 0.563 & -- \\
Entity-first KG & 218 & 0.665 & 0.580 & 0.666 & 0.688 & 0.701 & 0.672 & 0.565 & 0.543 / 0.151 \\
\bottomrule
\end{tabular}
\end{table*}

\begin{table*}[t]
\centering
\small
\caption{Post-hoc RealMedQA GPS replay on the full $n=230$ answer logs.
No generation, retrieval, or LLM call was rerun; only GPS
was recomputed against the persistent dataset KG.
``Old'' is the original surface-matched, binary-reachability GPS;
``new'' is the final soft-linked, depth-matched estimator of
Section~\ref{sec:struct_metrics}.
The linking threshold and decay ($\tau=0.60$, $\gamma=0.4$) were selected on
this dataset and applied frozen everywhere else, so this table shows
calibrated in-domain behaviour; the open-domain tables provide the held-out
estimate.
\label{tab:realmedqa_replay}}
\begin{tabular}{@{}lccc@{}}
\toprule
Setting & Answered & GPS usable (old$\rightarrow$new) & GPS AUROC (old$\rightarrow$new) \\
\midrule
Dense + vanilla & 219 & 100 $\rightarrow$ 192 & 0.72 $\rightarrow$ 0.89 \\
Entity-first + KG & 223 & 97 $\rightarrow$ 195 & 0.47 $\rightarrow$ 0.76 \\
\bottomrule
\end{tabular}
\end{table*}

\begin{table*}[t]
\centering
\small
\caption{RealMedQA no-rebuild strict graph-only stress test on the full
$n=230$ question set.
The adaptive row is the reported KG policy from the full RealMedQA run.
The strict row reuses the same RealMedQA graph and disables dense fallback,
vector augmentation, and decomposition, testing whether a more
concentrated entity-first retrieval regime compresses answer-state
uncertainty.
All metric columns report AUROC.
\label{tab:realmedqa_strict_entity}}
\setlength{\tabcolsep}{3.5pt}
\begin{tabular}{@{}lcccccccc@{}}
\toprule
KG policy & Ans. & Acc. & Overlap & DSE & VN & SD-UQ & SEU & GPS \\
\midrule
Adaptive + fallback & 223 & 0.946 & 0.558 & 0.699 & 0.815 & 0.808 & 0.501 & 0.76$^\dagger$ \\
Strict graph-only & 200 & 0.670 & 0.661 & 0.463 & 0.248 & 0.233 & 0.721 & 0.746$^\dagger$ \\
\bottomrule
\end{tabular}

\vspace{0.4em}
\raggedright\footnotesize
$^\dagger$ GPS values come from the post-hoc replay
(Table~\ref{tab:realmedqa_replay}): adaptive $195/223$ usable rows, strict
$142/200$.
RealMedQA is the calibration domain for the GPS hyperparameters, so these
cells show calibrated in-domain behaviour; Table~\ref{tab:gps_reliability_bins}
gives the corresponding binned error-rate view.
\end{table*}

\paragraph{Depth-matched GPS scoring.}
GPS uses a depth-matched decay, $\gamma^{|L_e - \hat{L}|}$, where $\hat{L}$ is
the expected reasoning depth: the question's logged hop count where available
(2WikiMultiHopQA, MuSiQue), the dataset's nominal depth by construction
(HotpotQA variants, $\hat{L}=2$), and $\hat{L}=1$ otherwise.
Because $\hat{L}=1$ on RealMedQA, the calibration domain is untouched and the
frozen $(\tau, \gamma)$ carry over; all other rows are single-shot
evaluations of the same declared definition.
This is the GPS definition used in all main tables.

\begin{table*}[t]
\centering
\scriptsize
\caption{Percentile-bootstrap 95\% intervals for the headline answer-state and
evidence-state AUROCs.
These are the main answer-side metrics used in the narrative: SRE-UQ and
SD-UQ as the practical answer-state baselines, and SEU as the evidence-state signal.
The table is intended to calibrate scale, not to claim significance testing
for every pairwise gap.
\label{tab:headline_ci_output}}
\begin{tabular}{@{}lcccccc@{}}
\toprule
Dataset & SRE-UQ Van. & SRE-UQ KG & SD-UQ Van. & SD-UQ KG & SEU Van. & SEU KG \\
\midrule
PubMedQA & 0.56 [0.49, 0.64] & 0.64 [0.55, 0.73] & 0.60 [0.46, 0.74] & 0.70 [0.57, 0.81] & 0.55 [0.44, 0.67] & 0.59 [0.47, 0.71] \\
RealMedQA & 0.43 [0.24, 0.62] & 0.66 [0.51, 0.80] & 0.63 [0.39, 0.86] & 0.81 [0.64, 0.94] & 0.63 [0.47, 0.79] & 0.50 [0.35, 0.63] \\
HotpotQA & 0.61 [0.54, 0.68] & 0.63 [0.57, 0.70] & 0.64 [0.56, 0.72] & 0.64 [0.56, 0.70] & 0.46 [0.39, 0.54] & 0.48 [0.40, 0.55] \\
HotpotQA FullWiki & 0.63 [0.55, 0.71] & 0.69 [0.60, 0.76] & 0.66 [0.58, 0.75] & 0.67 [0.60, 0.75] & 0.56 [0.47, 0.65] & 0.56 [0.49, 0.64] \\
2WikiMHQA & 0.59 [0.52, 0.67] & 0.65 [0.58, 0.72] & 0.61 [0.52, 0.69] & 0.69 [0.61, 0.75] & 0.62 [0.53, 0.70] & 0.64 [0.57, 0.72] \\
MuSiQue & 0.64 [0.53, 0.75] & 0.66 [0.56, 0.77] & 0.67 [0.55, 0.78] & 0.63 [0.52, 0.75] & 0.63 [0.52, 0.75] & 0.63 [0.51, 0.74] \\
\bottomrule
\end{tabular}
\end{table*}

\subsection{Answer-State Baseline Metric Definitions}
\label{sec:appendix_answer_metrics}

These are the five borrowed answer-state estimators summarised in
Section~\ref{sec:gen_metrics}.
Each is applied unchanged; only SD-UQ, the answer-state score introduced here, is
defined in the main text.

\paragraph{Semantic entropy and DSE.}
\citet{kuhn2023semantic} cluster sampled responses into meaning-equivalent
sets $\mathcal{C}=\{C_1,\ldots,C_K\}$ using bidirectional NLI entailment.
The API setting exposes no reliable per-sample likelihoods, so the reported
score is the count-weighted black-box variant,
\begin{equation}
  \mathrm{DSE}(\bm{r})
  = -\sum_{k=1}^{K} \frac{|C_k|}{N}\log\frac{|C_k|}{N}.
\end{equation}
DSE is semantic entropy with uniform sample weights; likelihood-weighted SE is
not reported separately.

\paragraph{P(True).}
\textbf{P(True)}~\citep{kadavath2022language} is implemented as a black-box
cluster-agreement proxy: the fraction of samples in the same semantic cluster
as the first response
($\mathrm{P(True)}_{\mathrm{proxy}} =
|\{i: \mathrm{cl}(r_i)=\mathrm{cl}(r_1)\}|/N$);
uncertainty is $1 - \mathrm{P(True)}_{\mathrm{proxy}}$.
It is monotone in DSE here and is not treated as independent evidence.

\paragraph{SelfCheckGPT.}
\citet{manakul2023selfcheckgpt} measure the fraction of response pairs flagged
as contradictions by NLI:
\begin{equation}
  \mathrm{SCG}
  = \frac{1}{|\mathcal{P}|}
    \sum_{(i,j)\in\mathcal{P}}
    \bigl[\mathrm{NLI}(r_i,r_j)=\text{contradiction}\bigr],
\end{equation}
where $\mathcal{P}$ is the set of ordered response pairs.
For these short-answer tasks, the implementation uses whole-response NLI.

\paragraph{SRE-UQ.}
\citet{vipulanandan2026sreuq} measure perturbation sensitivity of the response
embedding distribution.
Let $\Delta_i$ denote the first-order perturbation score of response embedding
$\bm{\phi}_i$ around the weighted kernel mean embedding, using the published
bandwidth rule.
This analysis uses
$\displaystyle\mathrm{SRE\text{-}UQ}(\bm{r})
  = \frac{1}{M}\sum_{i \in \mathcal{T}} |\Delta_i|$,
where $\mathcal{T}$ indexes the $M$ highest-amplitude modes.
The mode cap is $M=8$, so at the $N=5$ budget all responses are retained, and
the Gaussian kernel bandwidth is $\sigma = \mathrm{std}_i\lVert\bm{\phi}_i -
\bar{\bm{\phi}}\rVert_2$, the standard deviation of response distances to the
kernel mean embedding $\bar{\bm{\phi}}$; both settings are listed in
Appendix~\ref{sec:appendix_reproducibility}.
The estimator is imported as written; the contribution is its KG-RAG
benchmark, not a new perturbation objective.

\paragraph{VN-Entropy (KG-RAG instantiation).}
VN-Entropy measures spectral diversity of the normalised response-embedding
Gram matrix.
With $\bm{G}=\bm{V}\bm{V}^\top$ and $\bm{\rho}=\bm{G}/N$,
\begin{equation}
  \mathrm{VN\text{-}Entropy}(\bm{r}) = S(\bm{\rho})
  = -\operatorname{tr}(\bm{\rho}\log\bm{\rho})
  = -\textstyle\sum_{i=1}^{N} \lambda_i \log \lambda_i,
  \label{eq:vn_entropy}
\end{equation}
where $\{\lambda_i\}_{i=1}^N$ are the eigenvalues of $\bm{\rho}$.
$S(\bm{\rho}) = 0$ when all responses are identical and approaches $\log N$
when responses are mutually orthogonal.
This is a cosine-Gram instantiation of the VNE family introduced by Kernel
Language Entropy \citep{nikitin2024kernellanguageentropy}; the novelty lies in
the KG-RAG benchmark.

\section{Extended Results and Robustness Analyses}
\label{sec:appendix_extended}

This section reports the robustness and mechanism checks referenced from
Section~\ref{sec:results}.
Sections~\ref{sec:appendix_silent_sensitivity}--\ref{sec:appendix_diversity}
test whether silent errors persist under alternative sampling budgets and
retrieval-stability views; Sections~\ref{sec:appendix_verbalized}
and~\ref{sec:appendix_composite} compare auxiliary confidence and
family-disagreement signals; Sections~\ref{sec:appendix_gps_sensitivity}
and~\ref{sec:appendix_residue} check GPS hyperparameter sensitivity and audit the
fully unflagged residue; and Section~\ref{sec:appendix_relocated} holds two
relocated diagnostic tables.

\subsection{Silent-Error Threshold Sensitivity}
\label{sec:appendix_silent_sensitivity}

The exact-floor silent-error definition used in
Table~\ref{tab:silent_failures} is conservative: requiring only unanimous
sampled answers (DSE $=0$, any SD-UQ) raises the pooled rates to
$67\%$/$52\%$/$88\%$ (dense / adaptive KG / strict), and admitting one
dissenting sample out of five raises them to $84\%$/$69\%$/$92\%$
(Table~\ref{tab:silent_sensitivity}), so the headline definition understates
rather than inflates the blind spot.
Unanimity at $N=5$ is already informative: if the modal answer held only
$0.7$ probability, five identical samples would occur with probability
$0.7^5 \approx 0.17$.
Two $20$-sample probes confirm the mass is not a budget artefact in the
analysed snapshots.
On the strict clinical subset, all $24$ empty-route wrong answers stay
perfectly unanimous at $N{=}20$ ($\mathrm{DSE}=0$ for every row; SD-UQ AUROC
$0.17$, unchanged from $N{=}5$).
On \emph{dense} 2WikiMultiHopQA (a pure non-graph retriever), $68\%$ of wrong
answers remain strictly silent and $86\%$ stay within one dissenting sample of
twenty, against $79\%$ strictly silent at $N{=}5$ on the headline run (the
$N{=}20$ probe uses an $n{=}100$ subset and the $N{=}5$ headline an $n{=}250$
subset, so the level at $N{=}20$ is the robust statistic rather than the
precise $79\!\to\!68\%$ delta).
Quadrupling the budget barely moves the silent rate, including on
a dense system with no graph state.
These are narrow probes: they show the headline failure mode persists, not a
full sampling-budget sweep across every KG setting.

\subsection{Interventional Dose-Response}
\label{sec:appendix_doseresponse}

The strict-clinical collapse summarised in
Section~\ref{sec:results_collapse} is not a chunk-level stochasticity
artefact.
An interventional dose-response holds the graph, policy, prompts, and
questions fixed ($n=100$, seed 42) and varies only final-stage retrieval
determinism, sampling the final $k{=}10$ chunks from a $4k$ shortlist at
retrieval temperature $\{0, 0.5, 1.0\}$.
The manipulation works: mean within-question chunk overlap falls from
$0.67$ to $0.29$, a $57\%$ reduction.
Yet the collapse does not move: SD-UQ AUROC is $0.18/0.18/0.16$ across doses
and the silent-error count is identical
($24$ of $29$--$30$ wrong, with the same $24/24$ empty-retrieval rows in each
arm).
Chunk-level stochasticity does not dislodge the lock-in; the route
decomposition in Section~\ref{sec:results_collapse} locates the silent mass in
the empty-retrieval state instead.

\subsection{Archived Concentration--Dispersion Traces}
\label{sec:appendix_concentration}

Direct per-question evidence comes from archived traces that retained
per-question retrieval overlap
(2WikiMultiHopQA $n=100$, MuSiQue $n=66$;
Figure~\ref{fig:concentration_facets}).
Among \emph{wrong} KG answers the pooled coupling is negative
(Spearman $\rho = -0.33$, $p = 0.003$, $n = 82$): the more concentrated the
retrieval state, the lower the wrong-answer dispersion.
The coupling is carried by 2WikiMultiHopQA ($\rho = -0.32$, $p = 0.01$,
$n = 60$), the bridge-style dataset where lock-in is sharpest, while MuSiQue's
small wrong-answer set trends the other way ($\rho = +0.25$, $n = 22$,
not significant).
On the dense side the coupling is absent where overlap is near-deterministic
and positive on MuSiQue ($\rho = +0.55$), nothing like the KG-side bridge
pattern.
These archived traces support, but do not identify, the mechanism.

\subsection{Within-Adaptive Stability Check}
\label{sec:appendix_within_stability}

An archived 2WikiMultiHopQA $n=100$ trace, unlike the newer $n=250$ run,
retained per-question chunk-overlap fields;
Table~\ref{tab:within_adaptive_stability} uses it for a small
within-adaptive stability check.
The high-overlap band does not prove collapse monotonically, but it is a useful
warning case: accuracy falls to $0.294$, DSE AUROC is near chance at $0.467$,
and VN-Entropy AUROC falls to $0.583$.
SD-UQ still carries moderate signal ($0.633$).
High-stability strata thus exist inside an adaptive policy, but the logger must
persist per-question chunk, entity, and path overlap before this becomes a
headline claim.

\begin{table*}[t]
\centering
\small
\caption{%
  \textbf{Within-adaptive retrieval-stability check.}
  Mechanism check from an archived 2WikiMultiHopQA $n=100$ trace that
  retained per-question chunk-overlap fields. The newer $n=250$ run is the
  headline result; this table is a supporting diagnostic.
  Bands are population tertiles of per-question chunk overlap; because a
  large fraction of questions sit at exactly $1.0$ overlap, ties spill
  across the tertile boundary.
  AUROC treats incorrect answers as the positive class.%
}
\label{tab:within_adaptive_stability}
\setlength{\tabcolsep}{3.5pt}
\begin{tabular}{@{}lrrrrrrr@{}}
\toprule
Overlap band & $n$ & Range & Acc. & DSE AUC & SD-UQ AUC & VN AUC & DSE mean \\
\midrule
Low & 33 & 0.400--0.840 & 0.364 & 0.415 & 0.714 & 0.750 & 0.277 \\
Medium & 33 & 0.850--1.000 & 0.545 & 0.667 & 0.800 & 0.752 & 0.103 \\
High & 34 & 1.000--1.000 & 0.294 & 0.467 & 0.633 & 0.583 & 0.157 \\
\bottomrule
\end{tabular}
\end{table*}

\subsection{Dedicated Per-Sample Graph-State Diversity Run}
\label{sec:appendix_diversity}

Table~\ref{tab:diversity_run} comes from a run built to close the logging gap
above, where archived traces persisted chunk overlap alone.
It was run on HotpotQA-FullWiki ($n=221$ answered questions with complete
per-sample traces, $74$ wrong), reusing the existing FullWiki KG with no rebuild.
For every question it records the mean pairwise Jaccard overlap across the
$N=5$ samples for four retrieval-state families (matched seed entities,
traversed paths, assembled subgraph, and retrieved chunks), together with the
per-question SD-UQ used for the coupling test.
It confirms two pieces of the argument, both reported in the main text: a
near-stable retrieval state across samples (the premise
Equation~\eqref{eq:collapsed} assumes) and a negative overlap--dispersion
coupling among wrong answers.
The seed-entity coupling there is null only because seed overlap saturates near
$1.0$, leaving almost no variance to correlate against.
The sign and magnitude of the path, subgraph, and chunk couplings match the
archived 2WikiMultiHopQA trace of Section~\ref{sec:appendix_concentration} on
independent data and across the full set of overlap families, so the coupling no
longer rests on chunk overlap from a single archived run.

\begin{table}[t]
\centering
\scriptsize
\caption{%
  \textbf{Per-sample graph-state diversity on the dedicated HotpotQA-FullWiki
  run} ($n=221$ answered, $74$ wrong).
  \emph{Stability} is the mean (median) pairwise Jaccard overlap across the
  $N=5$ samples per question, by retrieval-state family.
  \emph{Coupling} is the correlation between that per-question overlap
  and answer-state dispersion (SD-UQ) among wrong answers ($n=74$), reported as
  Spearman's $\rho$ with its two-sided $p$-value; a negative
  value means more overlap goes with less dispersion, the lock-in prediction.%
}
\label{tab:diversity_run}
\begin{tabular}{@{}lcccc@{}}
\toprule
 & \multicolumn{2}{c}{Stability (Jaccard)} & \multicolumn{2}{c}{Coupling vs SD-UQ} \\
\cmidrule(lr){2-3}\cmidrule(lr){4-5}
Family & Mean & Median & $\rho$ & $p$ \\
\midrule
Seed entity & $0.99$ & $1.00$ & $+0.09$ & $0.47$ \\
Path        & $0.58$ & $0.58$ & $-0.30$ & $0.009$ \\
Subgraph    & $0.64$ & $0.63$ & $-0.23$ & $0.048$ \\
Chunk       & $0.65$ & $0.65$ & $-0.24$ & $0.037$ \\
\bottomrule
\end{tabular}
\end{table}

\begin{figure}[t]
\centering
\includegraphics[width=0.72\linewidth]{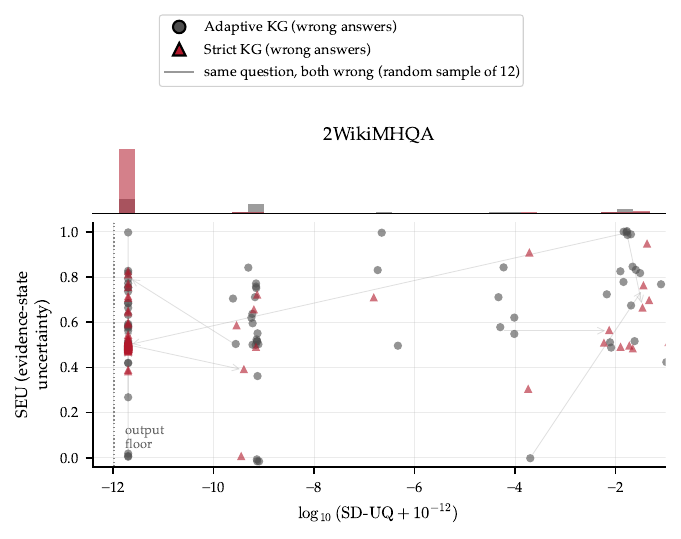}
\caption{%
  \textbf{Lock-in as migration into the low-dispersion corner
  (2WikiMultiHopQA).}
  Each point is a \emph{wrong} KG answer, plotted by answer-state dispersion
  (SD-UQ, log scale) and evidence-state uncertainty (SEU), under the adaptive
  policy (grey circles) and the strict graph-only stress test (red triangles);
  grey arrows connect a random sample of $12$ questions wrong under \emph{both}.
  Under strict retrieval the SD-UQ mass collapses onto the floor while SEU stays
  spread. Small jitter is added for legibility.%
}
\label{fig:lockin_migration}
\end{figure}

\subsection{Verbalised-Confidence Baseline}
\label{sec:appendix_verbalized}

The verbalised-confidence baseline summarised in
Section~\ref{sec:results_output} uses the original P(True) protocol, which
asks the model for a probability rather than counting cluster agreement and is
not blind to silent errors by construction.
Querying the same backbone once per saved answer
(Table~\ref{tab:verbalized_confidence}) yields pooled adaptive-KG
AUROC $0.69$: competitive on the HotpotQA variants ($0.72$/$0.77$, above
SD-UQ there), weaker than SD-UQ on RealMedQA ($0.65$ vs.\ $0.81$), and near
chance precisely where bridge-style lock-in lives (2WikiMultiHopQA: $0.49$
adaptive, $0.58$ strict).
On the strict clinical stress test it reaches $0.89$ while sampled dispersion
collapses to $0.233$, so part of the clinical silent-error mass is
recoverable from the generator's own self-estimate without any retrieval-side
signal.
If a one-call self-estimate can flag clinical silent errors, why decompose at
all?
First, it fails on bridge-style lock-in
($0.49$--$0.58$), the regime the decomposition targets: where the wrong
answer is parametrically plausible, \emph{no} output-side signal tested
here, sampled or verbalised, ranks the errors.
Second, a verbalised probability is an uncalibrated self-report with known
self-preference pathologies, whereas SEU and GPS are grounded in visible
evidence and graph state.
Third, the self-estimate returns a number without a reason: when it
disagrees with the answer there is no provenance to audit, which is precisely
the evidence a clinical review workflow would need.

\subsection{Family Disagreement and the Simple Composite Audit Score}
\label{sec:appendix_composite}

\begin{figure*}[t]
\centering
\includegraphics[width=\linewidth]{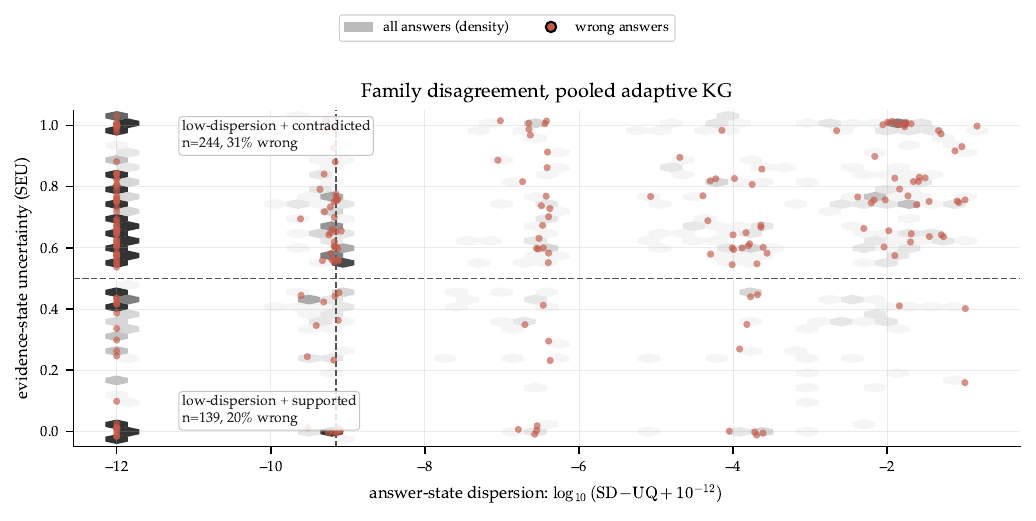}
\caption{%
  \textbf{Family disagreement in the pooled adaptive KG runs.}
  Grey hexagons show the density of all answered questions with a
  non-neutral SEU verdict; red points overlay the wrong answers.
  The $x$-axis is answer-state residual dispersion (SD-UQ, log scale) and
  the $y$-axis is evidence-state uncertainty (SEU).
  The all-neutral $\mathrm{SEU}=0.5$ rows, where the NLI head said nothing
  ($n=464$, $24\%$ wrong), are excluded.
  Dashed lines mark the pooled SD-UQ median and $\mathrm{SEU}=0.5$; the two
  low-answer-uncertainty quadrants carry the decision-relevant contrast:
  low answer uncertainty with contradicted evidence is $31\%$ wrong against
  $20\%$ for low answer uncertainty with support, so answer-surface agreement
  is not a proxy for evidence support.%
}
\label{fig:family_disagreement}
\end{figure*}

Beyond the conjunctive audit rule of Section~\ref{sec:results_composite},
Table~\ref{tab:composite_audit_score} gives a deliberately simple audit score:
the mean of within-dataset percentile ranks for SD-UQ, SEU, and final GPS-risk,
with GPS abstention treated as high risk.
It is uneven and not claimed to dominate the best answer-state metric.
The PubMedQA row shows where the combination earns its keep: GPS abstains on
every yes/no row (flat $0.5$), yet Combined reaches $0.813$ against SD-UQ's
$0.699$, because SEU contributes complementary support-level ranking exactly
where answer dispersion is coarse, not because the abstaining family adds
signal.
The score shows how the reported families can feed a decision layer;
a deployable policy would still need held-out calibration.

\begin{table*}[t]
\centering
\small
\caption{%
  \textbf{Adaptive-KG composite audit score.}
  The combined score is the mean of within-dataset percentile ranks for SD-UQ,
  SEU, and final GPS-risk, with GPS abstention treated as high risk.
  AUROC treats incorrect answers as the positive class; ``Base err.''\ is the
  no-abstention error rate, and lower area under the risk-excess-coverage curve
  (AUREC) is better within a dataset.
  ``Best single'' is the largest per-family AUROC in the row; bold cells mark
  datasets where the combined score exceeds the best single-family metric.%
}
\label{tab:composite_audit_score}
\setlength{\tabcolsep}{4pt}
\begin{tabular}{@{}lrrrrrrrr@{}}
\toprule
Dataset & $n$ & Base err. & SD-UQ AUC & SEU AUC & GPS-risk AUC & Best single & Combined AUC & Combined AUREC \\
\midrule
PubMedQA & 100 & 0.270 & 0.699 & 0.594 & 0.500 & 0.699 & \textbf{0.813} & 0.096 \\
RealMedQA & 223 & 0.054 & 0.808 & 0.501 & 0.703 & \textbf{0.808} & 0.729 & 0.022 \\
HotpotQA & 238 & 0.395 & 0.635 & 0.477 & 0.551 & \textbf{0.635} & 0.573 & 0.326 \\
HotpotQA FullWiki & 218 & 0.335 & 0.672 & 0.565 & 0.549 & \textbf{0.672} & 0.661 & 0.228 \\
2WikiMHQA & 244 & 0.311 & 0.685 & 0.643 & 0.414 & \textbf{0.685} & 0.619 & 0.225 \\
MuSiQue & 94 & 0.606 & 0.631 & 0.626 & 0.640 & 0.640 & \textbf{0.745} & 0.426 \\
\bottomrule
\end{tabular}
\end{table*}

\subsection{GPS Hyperparameter Sensitivity}
\label{sec:appendix_gps_sensitivity}

To test whether the headline GPS AUROCs depend on the two calibrated
hyperparameters, the soft-link threshold $\tau$ and the distance decay $\gamma$
are swept over $\tau \in \{0.50, 0.55, 0.60, 0.65, 0.70\}$ and
$\gamma \in \{0.2, 0.4, 0.6, 0.8, 1.0\}$ ($25$ cells), recomputed by offline
replay from the saved GPS replay stores. No generation, retrieval, linking, or KG
build is rerun.
Table~\ref{tab:gps_sensitivity} reports the calibrated cell (which reproduces
the headline numbers), the AUROC range across the grid, its standard deviation,
and the usable-row range.
The held-out open-domain AUROCs are robust to the hyperparameters (standard
deviation $\leq 0.018$ on HotpotQA, HotpotQA FullWiki, 2WikiMultiHopQA, and
MuSiQue), so the open-domain weakness, including the below-chance
2WikiMultiHopQA score, is a genuine retrieval limitation rather than a tuning
artefact.
The clinical calibration domain is the most sensitive ($0.038$ adaptive,
$0.091$ strict), as expected for the dataset the hyperparameters were tuned on.

\begin{table}[t]
\centering
\small
\caption{GPS AUROC sensitivity to $(\tau, \gamma)$ by offline replay.
``Calib.'' is the paper's frozen $\tau{=}0.60$, $\gamma{=}0.40$ cell; range,
SD, and usable counts are over the $5\times5$ grid.
\label{tab:gps_sensitivity}}
\setlength{\tabcolsep}{4pt}
\begin{tabular}{@{}lcccc@{}}
\toprule
Run & Calib. & Grid range & SD & Usable \\
\midrule
RealMedQA adaptive & 0.76 & 0.60--0.76 & 0.038 & 191--197 \\
RealMedQA strict   & 0.75 & 0.53--0.90 & 0.091 & 139--171 \\
HotpotQA           & 0.51 & 0.50--0.55 & 0.014 & 154--169 \\
HotpotQA FullWiki  & 0.54 & 0.53--0.56 & 0.008 & 182--187 \\
2WikiMHQA adaptive & 0.38 & 0.35--0.40 & 0.012 & 177--188 \\
2WikiMHQA strict   & 0.48 & 0.40--0.53 & 0.035 & 94--104 \\
MuSiQue            & 0.68 & 0.64--0.70 & 0.018 & 80--85 \\
\bottomrule
\end{tabular}
\end{table}

\subsection{Audit of the Fully Unflagged Residue}
\label{sec:appendix_residue}

Of the $141$ pooled adaptive-KG silent errors, $7$ ($5\%$) are unflagged by
every non-answer family (Section~\ref{sec:results_collapse}): not on an empty
route, $\mathrm{SEU}\leq 0.5$, and GPS defined and $\leq 0.5$.
A manual audit of all seven from the saved logs (question, gold answer,
generated answer) characterises them as two benchmark-ambiguous or
contested-label questions (e.g.\ the founding year of an institution with a
disputed predecessor date; a subjective ``more known for'' comparison), two
wrong-answer-slot errors where the model returns a question anchor rather than
the asked attribute (e.g.\ naming a song's performer instead of their cause of
death), and three parametric conflations or specificity mismatches (e.g.\
returning a film character's love interest instead of the lead actress's real
spouse).
SEU sits at its neutral $0.5$ default on four of the seven and below it on the
other three, so it abstains or mildly supports rather than flagging; by
construction none exceed $0.5$.

Crucially, this manual pass reads the saved logs (question, gold answer,
generated answer, scores, route metadata) only; chunk text is re-retrievable
from the KG, but verifying whether the evidence itself \emph{entails} the wrong
answer needs the cross-passage entailment and path-faithfulness replay left as
future work, so none is confirmed as deep presence lock-in, and the residue is
better read as a mixture of label ambiguity, answer-slot errors, and parametric
overconfidence than as evidence that undetectable lock-in dominates.

\subsection{Relocated Diagnostic Tables}
\label{sec:appendix_relocated}

Table~\ref{tab:twowiki_hopwise_diagnostics}, collected here to keep the results
flow tight, gives the full hop-stratified 2WikiMultiHopQA comparison across the
three retrieval policies; its decisive row, the strict graph-only 4-hop slice
with zero clean accuracy and collapsed SD-UQ and VN-Entropy, is discussed in
Section~\ref{sec:results_collapse}.
It uses clean answered rows rather than raw hop-bucket sizes, so $n$ differs
across policies when provider failures or skipped systems remove rows, the same
convention used for clean accuracy.

\begin{table*}[t]
\centering
\small
\caption{%
  \textbf{Hop-wise 2WikiMultiHopQA diagnostics.}
  Computed on the same fixed $n=250$ subset.
  The table uses clean answered rows; $n_{\mathrm{w}}$ is the number of wrong
  answered rows (the AUROC positive class).
  AUROC is omitted when a hop slice has only one correctness class.
  The $\geq$5-hop bucket contains four questions and is kept only in the
  released JSON artefact.%
}
\label{tab:twowiki_hopwise_diagnostics}
\setlength{\tabcolsep}{3.4pt}
\begin{tabular}{@{}llrrrrrrrrrr@{}}
\toprule
Policy & Hop & $n$ & $n_{\mathrm{w}}$ & Acc. & SD mean & VN mean & SEU mean &
SD AUC & VN AUC & SEU AUC & GPS AUC \\
\midrule
Dense vanilla & 2-hop & 150 & 42 & 0.720 & 0.000 & 0.092 & 0.612 & 0.615 & 0.639 & 0.615 & 0.491 \\
Dense vanilla & 4-hop & 58 & 16 & 0.724 & 0.000 & 0.198 & 0.458 & 0.628 & 0.571 & 0.656 & 0.410 \\
Adaptive KG & 2-hop & 182 & 57 & 0.687 & 0.003 & 0.191 & 0.544 & 0.702 & 0.668 & 0.611 & 0.335 \\
Adaptive KG & 4-hop & 58 & 16 & 0.724 & 0.002 & 0.248 & 0.466 & 0.600 & 0.638 & 0.646 & 0.452 \\
Strict KG & 2-hop & 183 & 83 & 0.546 & 0.003 & 0.127 & 0.569 & 0.560 & 0.680 & 0.378 & 0.353 \\
Strict KG & 4-hop & 58 & 58 & 0.000 & 0.000 & 0.000 & 0.500 & -- & -- & -- & -- \\
\bottomrule
\end{tabular}
\end{table*}

Table~\ref{tab:gold_reachability}, also relocated here, reports the
gold-reachability gap behind the falsifiable GPS prediction of
Section~\ref{sec:results_structural}: a positive gap (correct answers harder to
reach than wrong ones) would tend toward below-chance GPS, a diagnostic pattern
consistent with the regenerated HotpotQA FullWiki KG and an earlier
2WikiMultiHopQA build.

\begin{table}[t]
\centering
\footnotesize
\caption{%
  \textbf{Gold-reachability gap and GPS's position relative to chance.}
  ``Unreach.'' is the fraction of rows whose gold answer entity is unreachable
  from the question entities, split by correctness; ``Gap'' is correct minus
  wrong.
  A positive gap would tend to predict below-chance GPS.
  The HotpotQA-FW row regenerates from the released KG; the 2WikiMHQA row
  ($^\dagger$) does not.%
}
\label{tab:gold_reachability}
\setlength{\tabcolsep}{4pt}
\begin{tabular}{@{}lcccc@{}}
\toprule
Dataset & $n$ & Unreach.\ corr. & Unreach.\ wrong & Gap \\
\midrule
2WikiMHQA$^\dagger$ & 244 & $0.71$ & $0.45$ & $+0.26$ \\
HotpotQA-FW    & 177 & $0.31$ & $0.51$ & $-0.20$ \\
\bottomrule
\end{tabular}

\vspace{2pt}
{\footnotesize GPS AUROC: $0.38$ (2WikiMHQA, below chance);
$0.54$ (HotpotQA-FW, near chance).
$^\dagger$From an earlier KG build, not regenerated from the current released
KG.}
\end{table}

\FloatBarrier
\section{Additional Qualitative Cases}
\label{sec:appendix_cases}

\begin{table*}[!t]
\centering
\small
\setlength{\tabcolsep}{4pt}
\renewcommand{\arraystretch}{1.2}
\caption{%
  \textbf{Consolidated case gallery from the reported runs.}
  \texttimes{} rows are verified lock-in failures: the KG answer is wrong yet
  repeated across all samples ($\mathrm{DSE}=0$, SD-UQ at or near the floor);
  the diagnostic columns show which family, if any, still fires.
  \checkmark{} rows are KG successes on bridge questions, where the same
  retrieval stability is helpful because the path reaches the right entity
  and relation, plus one correct answer despite GPS abstention.
  The Baltic Cup row is the hardest failure: the retrieved evidence
  \emph{entails} the wrong answer, so every scalar family reports low risk and
  only the provenance trace remains auditable.
  GPS $=0.5$ denotes abstention.
  The cases are selected for illustration, not sampled; prevalence claims
  rest on Tables~\ref{tab:silent_failures} and
  \ref{tab:failure_taxonomy}, not on this gallery.%
}
\label{tab:lockin_gallery}
\begin{tabular}{@{}c>{\raggedright\arraybackslash}p{0.26\textwidth}
                  l
                  >{\raggedright\arraybackslash}p{0.22\textwidth}
                  l c c@{}}
\toprule
\textbf{KG} & \textbf{Question (dataset)} & \textbf{Gold} & \textbf{Anchor / path}
& \textbf{KG answer} & \textbf{SEU} & \textbf{GPS} \\
\midrule
\texttimes & Osireion is behind the temple of which pharaoh? (HotpotQA)
 & Seti I
 & \emph{Ramesses II} neighbourhood of the Abydos temple
 & Ramesses II & $1.00$ & $0.44$ \\
\texttimes & Who is the paternal grandfather of Uskhal Khan? (2Wiki)
 & Khutughtu Khan
 & \emph{Toghon Tem\"ur} (father, one hop; bridge skipped)
 & Toghon Tem\"ur & $0.60$ & $0.00$ \\
\texttimes & Who created Mickey Mouse's spouse? (MuSiQue)
 & Walt Disney
 & \emph{Ub Iwerks} via Mickey Mouse $\to$ creator
 & Ub Iwerks & $0.75$ & $0.00$ \\
\texttimes & Who is Ieuan ab Owain Glynd\^wr's paternal grandfather? (2Wiki)
 & Gruffudd Fychan II
 & \emph{Owain Glynd\^wr} (father, one hop; bridge skipped)
 & Owain Glynd\^wr & $0.43$ & $0.24$ \\
\texttimes & Which country skipped the 1991 Baltic Cup? (HotpotQA)
 & Belarus
 & \emph{Estonia} via Baltic Cup participants
 & Estonia & $0.00$ & $0.00$ \\
\texttimes & Are TEC-1 and Dubna 48K based on the same processor? (HotpotQA)
 & yes
 & weak graph state; no usable anchor, GPS abstains
 & no & $0.75$ & $0.50$ \\
\texttimes & 2010 population of the city popular with tourists? (MuSiQue)
 & 8.005 million
 & wrong city; only linkable answer string four hops away
 & 2{,}217 & $0.67$ & $0.94$ \\
\midrule
\checkmark & Who is the spouse of the director of \emph{Fire-Eater}? (2Wiki)
 & Pirkko Saisio
 & film $\to$ director $\to$ spouse bridge preserved
 & Pirkko Saisio & $0.60$ & $0.44$ \\
\checkmark & Who is the father of the director of \emph{The Cup} (1999)? (2Wiki)
 & Thinley Norbu
 & film $\to$ director $\to$ father bridge preserved
 & Thinley Norbu & $0.69$ & $0.35$ \\
\checkmark & Birthplace of the director of \emph{Dollar} (1938)? (2Wiki)
 & Helsingfors
 & film $\to$ director $\to$ birthplace bridge preserved
 & Helsingfors & $0.50$ & $0.60$ \\
\checkmark & Dow Jones fall at the highest US unemployment rate? (MuSiQue)
 & 54.7\%
 & chain resolved correctly; GPS still abstains
 & 54.7\% & $0.72$ & $0.50$ \\
\bottomrule
\end{tabular}
\end{table*}

Table~\ref{tab:lockin_gallery} is the canonical case gallery.
The failure rows follow the form wrong anchor $\to$ stable retrieved context
$\to$ stable wrong answer $\to$ silent answer surface, showing per case which
diagnostic family still responds; the success rows show the same retrieval
stability working as intended on 2WikiMHQA bridge questions, where the graph
preserves the intermediate entity that vanilla retrieval loses, plus one
MuSiQue chain resolved correctly despite GPS abstention.
The main text uses one path-faithfulness example
(the Osireion lock-in of Figure~\ref{fig:lockin_trace}) to keep the exposition
focused on the lock-in mechanism; the gallery provides trace evidence, not
additional leaderboard claims.
Table~\ref{tab:path_faithfulness_case} gives the full per-family audit readout
for that Osireion case (question, retrieved graph state, and the
DSE/SD-UQ/GPS/SEU verdicts), expanding the summary in
Section~\ref{sec:results_cases}.

\begin{table*}[t]
\centering
\small
\setlength{\tabcolsep}{5pt}
\renewcommand{\arraystretch}{1.2}
\caption{%
  \textbf{Path-faithfulness failure under retrieval lock-in (HotpotQA).}
  A wrong-but-graph-coherent case from the fixed-$n=250$ run.
  The retriever anchors to a strongly associated but incorrect
  Nineteenth-Dynasty pharaoh, and every sampled answer repeats it.
  Answer-state metrics are silent and GPS reports support; only SEU fires.%
}
\label{tab:path_faithfulness_case}
\begin{tabular}{@{}>{\raggedright\arraybackslash}p{0.30\textwidth}
                  >{\raggedright\arraybackslash}p{0.30\textwidth}
                  >{\raggedright\arraybackslash}p{0.33\textwidth}@{}}
\toprule
\textbf{Question and gold answer} &
\textbf{Retrieved graph state} &
\textbf{Diagnostic reading} \\
\midrule
\emph{Osireion is located to the rear of the temple named after which New
Kingdom Nineteenth Dynasty of Egypt pharaoh?}

\vspace{0.3em}
Gold answer: \emph{Seti~I}.

\vspace{0.4em}
The intended chain attributes the Great Temple of Abydos, directly in front
of the Osireion, to its builder \emph{Seti~I}.
&
The retrieved subgraph anchors instead to \emph{Ramesses~II}, the adjacent
Nineteenth-Dynasty pharaoh who completed the temple and built his own nearby:
\begin{center}
\begin{tabular}{@{}c@{}}
Osireion / Abydos temple \\
$\downarrow$ associated pharaoh \\
Ramesses~II
\end{tabular}
\end{center}

\vspace{0.3em}
The generated KG answer is therefore \emph{Ramesses~II}~(\textbf{\texttimes}),
and dense retrieval makes the same error.
The graph state is locally coherent but attached to the wrong builder.
&
All sampled KG answers repeat the same wrong entity:
\[
  \mathrm{DSE}=0,\qquad \mathrm{SD\text{-}UQ}\approx 0.
\]
Answer-state uncertainty is therefore at its floor.

\vspace{0.3em}
GPS is low ($0.44$), so it does not flag the error:
\emph{Ramesses~II} is reachable in the retrieved graph, and GPS is reported
in risk orientation, where lower means stronger graph support for the
generated answer.

\vspace{0.3em}
SEU is maximal ($1.0$): every retrieved chunk contradicts the generated
answer, the one family that fires. \\
\bottomrule
\end{tabular}
\end{table*}

\FloatBarrier
\section{Prompts}
\label{sec:appendix_prompts}

The generation and judging prompts used in the experiments are reproduced
verbatim below; the entity- and relation-extraction prompts are in the code
release (Appendix~\ref{sec:appendix_code}).

\subsection{Correctness Judge Prompt}
\label{sec:appendix_judge_prompt}

Used to produce the binary correctness labels on which AUROC and AUREC
are computed.

\textit{System message:}
{\small\begin{verbatim}
You are a strict answer evaluator for a factoid question answering
task. Your job is to decide if a model's response is CORRECT.

Rules:
- Reply with exactly one word: 'correct' or 'incorrect'.
- The response is CORRECT only if it contains an answer semantically
  equivalent to the expected answer (minor spelling/accent differences
  are ok).
- The response is INCORRECT if the model says it doesn't know, cannot
  determine, or provides a factually different answer.
\end{verbatim}}

\textit{User message (template):}
{\small\begin{verbatim}
Question: {question}
Expected answer: {expected_answer}
Model response: {model_response}

Is the model response correct? Reply with one word only:
correct or incorrect.
\end{verbatim}}

\subsection{Vanilla RAG Generation Prompt}
\label{sec:appendix_vanilla_prompt}

Used by the vanilla RAG system to generate all $N = 5$ sampled
responses per question.

\textit{System message (template):}
{\small\begin{verbatim}
You are an AI assistant that provides accurate, factual answers
based on the provided context.

Context Information:
{context}

Guidelines:
- Read all context passages carefully -- the answer is often present
  but may require connecting two passages.
- For multi-hop questions: explicitly chain your reasoning step by
  step (e.g. "The film starred X -> X later held position Y").
- Base your answer on the provided context; do not invent facts.
- If the answer is not directly stated but can be inferred by
  connecting two pieces of evidence, make the inference explicitly
  and state your reasoning chain.
- Only say the context is insufficient if you genuinely cannot find
  any relevant evidence after carefully reading all passages.
- Be concise but comprehensive; include specific facts to support
  your answer.
- IMPORTANT: For yes/no questions (questions starting with: Is, Are,
  Does, Do, Can, Should, Was, Were, Has, Have), you MUST begin your
  response with either "Yes" or "No" as the very first word,
  followed by your explanation.

User Query: {question}
\end{verbatim}}

\subsection{KG-RAG Generation Prompt}
\label{sec:appendix_kg_prompt}

Used by the KG-RAG system to generate all $N = 5$ sampled responses
per question.
The instruction to prefer evidence appearing in \emph{both} text chunks
and graph paths is a plausible contributor to consistent, confident generation
under KG context, and may amplify graph-induced overconfidence.

\textit{System message (template):}
{\small\begin{verbatim}
You are a knowledgeable AI assistant. Answer the question using
the provided context.

The context has three parts:

1. TEXT CHUNKS -- document passages retrieved by semantic similarity.
2. GRAPH TRAVERSAL PATHS -- multi-hop chains discovered by walking
   the knowledge graph from seed entities.
   Each path shows how concepts connect:
   Entity A --RELATIONSHIP--> Entity B --RELATIONSHIP--> Entity C
3. ENTITIES -- individual concepts found in the graph.

HOW TO REASON:
- Read all text chunks carefully -- the answer is often present but
  requires connecting two passages.
- Start from the entities most relevant to the question, then follow
  graph paths to discover indirect relationships.
- For multi-hop questions: explicitly chain your reasoning step by
  step (e.g. "The film starred X -> X later held position Y").
- Prefer evidence that appears in BOTH a text chunk AND a graph path
  -- that is the strongest signal.

IMPORTANT:
- For yes/no questions, begin with "Yes" or "No" followed by your
  explanation.
- Ground every claim in the provided context; do not invent facts.
- If the answer is not directly stated but can be inferred by
  connecting two pieces of evidence in the context, make the
  inference explicitly and state your reasoning chain.
- Only say the context is insufficient if you genuinely cannot find
  any relevant evidence after carefully reading all passages.

Text Chunks:
{context}

Knowledge Graph Traversal Paths:
{graph_paths}

Entities:
{entities}

Question: {question}
\end{verbatim}}

\subsection{Entity Extraction Prompts}

The prompts used for LLM-based entity and relation extraction during KG
construction (open and ontology-guided modes) follow the JSON schema
described in Appendix~\ref{sec:appendix_kg}.
They are available in the supplementary code release
(Appendix~\ref{sec:appendix_code}).

\section{\system{} System Description}
\label{sec:appendix_ontographrag}

\system{} (v1.0.0) is the open-source platform used for all reported
experiments. It exposes a FastAPI application on port 8004 (\texttt{uvicorn
ontographrag.api.app:app --host 0.0.0.0 --port 8004}, or the console script
\texttt{ontograph}); Neo4j is checked at startup, and endpoints requiring the
graph database return \texttt{HTTP 503} if unavailable.

\subsection{Package Structure}

\begin{description}
  \item[\texttt{ontographrag.kg}]
    KG construction. \texttt{builders/} contains the open extraction
    and ontology-guided extraction pipelines.
    \texttt{loaders/} handles Neo4j serialisation, and \texttt{utils/}
    contains chunking, source-node, and graph-query helpers.

  \item[\texttt{ontographrag.rag}]
    \texttt{EnhancedRAGSystem} (KG-RAG pipeline) and
    \texttt{VanillaRAGSystem} (dense-retrieval baseline), with auxiliary
    modules for reranking, retrieval sampling, and answer guardrails.

  \item[\texttt{ontographrag.providers}]
    Unified LLM interface supporting OpenAI, Anthropic, Google Gemini,
    Vertex AI, OpenRouter, and Ollama.
\end{description}

Persistence: Neo4j $\geq$5.25 for all graph, vector, and full-text indexes.
Chunk and entity embeddings are stored as Neo4j native vector indexes;
no external vector store is required.
Install with \texttt{pip install .}; deployment requires Python $\geq$3.11
and Neo4j connection environment variables.

\subsection{Serving Interface}

The full REST API is documented in the software release. The relevant interface
here is the question-answering endpoint, which returns the generated answer,
retrieved text chunks, matched entities, graph paths, route labels, relation
anchors, and per-query diagnostic scores; the manuscript treats the API as
infrastructure for trace logging, not a separate systems contribution.

\FloatBarrier
\section{Additional Robustness Checks}
\label{sec:appendix_robustness}

\subsection{Corpus Dilution: How Curation Flatters Dense Retrieval}
\label{sec:appendix_dilution}

The accuracy comparisons in the main text run on curated benchmark corpora,
where each question's answer-bearing passage sits in a small candidate pool.
This probe isolates how much that curation helps dense retrieval, on the
retrieval step alone (no generation, no API). Using the persisted
HotpotQA-FullWiki chunk corpus ($2{,}489$ passages, one per document), we take the
$143$ free-text questions whose gold answer string is locatable in some chunk and
treat that chunk as the gold target. For a pool of size $P$ we score the gold
chunk against $P-1$ distractor chunks sampled from the rest of the shared corpus
(seed $42$), rank by question--chunk cosine under the deployed MiniLM encoder, and
record gold-passage recall@$k$. Growing $P$ from a curated $10$ to the full
corpus simulates moving from benchmark to deployment retrieval.

The effect in Table~\ref{tab:dilution} is large and monotone: recall@$10$
falls from $1.00$ at $P{=}10$ to $0.55$ at $P{=}2489$, and recall@$1$ from $0.69$
to $0.27$. Since the questions, encoder, and gold passages are held fixed, the
drop estimates the cost of removing curated candidate sets under a primary
string-match recall criterion (a stricter all-gold-chunk criterion would bound
it differently). The dense--KG near-match reported in
Section~\ref{sec:results_accuracy} is measured in the regime most
favourable to dense retrieval; at deployment scale, retrieval misses are far more
frequent, and a retrieval layer that exposes what was retrieved and whether it
supports the answer matters more, not less. The probe bounds the retrieval
step only; it does not re-estimate end-to-end accuracy, which would require
rerunning generation over each diluted pool.

\begin{table}[t]
\centering
\footnotesize
\caption{%
  \textbf{Gold-passage retrieval recall as the candidate corpus grows}
  (HotpotQA-FullWiki, $143$ locatable free-text questions, MiniLM cosine, seed
  $42$).
  $P$ is the candidate-pool size (gold passage plus $P-1$ sampled distractors).
  Recall collapses as curation is removed, with questions and encoder fixed.%
}
\label{tab:dilution}
\setlength{\tabcolsep}{6pt}
\begin{tabular}{@{}rccc@{}}
\toprule
Pool $P$ & Recall@1 & Recall@5 & Recall@10 \\
\midrule
10    & $0.69$ & $0.92$ & $1.00$ \\
25    & $0.66$ & $0.83$ & $0.88$ \\
50    & $0.62$ & $0.76$ & $0.79$ \\
100   & $0.56$ & $0.69$ & $0.74$ \\
250   & $0.49$ & $0.65$ & $0.68$ \\
500   & $0.43$ & $0.63$ & $0.65$ \\
1000  & $0.34$ & $0.56$ & $0.61$ \\
2489  & $0.27$ & $0.49$ & $0.55$ \\
\bottomrule
\end{tabular}
\end{table}

\subsection{Accuracy: Paired McNemar Test}
\label{sec:appendix_mcnemar}

Table~\ref{tab:mcnemar_accuracy} gives the exact paired McNemar test behind the
no-detectable-gap claim in Section~\ref{sec:results_accuracy}.
For each snapshot the test is computed on the rows answered by both systems,
counting the discordant pairs (dense correct while KG wrong, and the reverse);
the exact two-sided binomial $p$-value on those discordant pairs is the
appropriate test because the two systems see the same questions.
No snapshot reaches significance: the smallest $p$ is $0.12$ (HotpotQA FullWiki),
and even MuSiQue, whose $-0.081$ point gap is the largest, has only $14$ versus
$7$ discordant pairs ($p=0.19$). The point deltas are within paired
sampling noise, not statistically resolved by this fixed-subset test, so the
diagnostic comparison is made without a measured accuracy penalty.
Numbers regenerate from the saved per-question labels
(\texttt{accuracy\_parity\_mcnemar.json}); no generation or retrieval is rerun.

\begin{table}[t]
\centering
\footnotesize
\caption{%
  \textbf{Paired dense-vs-adaptive-KG accuracy with exact McNemar test.}
  $n$ is rows answered by both systems; ``D-only''/``KG-only'' are discordant
  pairs (one system correct, the other wrong); $\Delta$ is KG $-$ dense clean
  accuracy on the paired set; $p$ is the exact two-sided McNemar value.
  No snapshot shows a significant accuracy difference.%
}
\label{tab:mcnemar_accuracy}
\setlength{\tabcolsep}{4pt}
\begin{tabular}{@{}lrrrrr@{}}
\toprule
Dataset & $n$ & D-only & KG-only & $\Delta$ & McNemar $p$ \\
\midrule
PubMedQA    & 100 &  5 &  3 & $-0.020$ & $0.73$ \\
RealMedQA   & 218 &  4 &  4 & $+0.000$ & $1.00$ \\
HotpotQA    & 225 & 14 &  7 & $-0.031$ & $0.19$ \\
HotpotQA-FW & 196 & 18 &  9 & $-0.046$ & $0.12$ \\
2WikiMHQA   & 212 & 29 & 26 & $-0.014$ & $0.79$ \\
MuSiQue     &  86 & 14 &  7 & $-0.081$ & $0.19$ \\
\bottomrule
\end{tabular}
\end{table}

\subsection{Audit-Rule Generalisation: Stratified Effect and Split-Half}
\label{sec:appendix_audit_generalisation}

Table~\ref{tab:mh_2x2} reproduces the per-dataset $2\times2$ contingency cells
behind the Mantel--Haenszel analysis cited in
Section~\ref{sec:results_composite}, so the stratified odds ratio can be
recomputed from the manuscript rather than only from the released JSON.
Each row counts answered KG rows as certified-or-not (the conjunctive rule)
crossed with correct-or-wrong.
The Mantel--Haenszel common odds ratio across the six strata is $3.39$
(bootstrap $95\%$ CI $[1.64, 10.99]$, $B{=}5000$ resampling questions within
dataset), against a pooled (unstratified) odds ratio of $5.04$; the stratified
value is the honest one because it controls for the easy-domain confound (the
rule selects more, and more accurately, on the high-accuracy clinical domain).
The single data-dependent threshold (the per-dataset SD-UQ median) survives
held-out evaluation: recomputing it on a random half of each dataset and
applying the rule to the other half over $B{=}200$ splits gives mean precision
$0.918$ (5th--95th percentile $[0.867, 0.974]$) at mean coverage $0.078$,
essentially unchanged from the in-sample $0.919$ at $0.077$, so the headline
precision is not threshold overfitting.

\begin{table}[t]
\centering
\footnotesize
\caption{%
  \textbf{Per-dataset $2\times2$ cells for the conjunctive audit rule}
  (adaptive KG runs), feeding the Mantel--Haenszel odds ratio of
  Section~\ref{sec:results_composite}.
  ``Cert.'' is certified (selected as low-risk by the rule); ``Corr.''/``Wr.''
  are correct/wrong answered rows.
  PubMedQA selects nothing (binary answers expose no linkable answer entity), so
  it contributes no stratum.%
}
\label{tab:mh_2x2}
\setlength{\tabcolsep}{4pt}
\begin{tabular}{@{}lrrrr@{}}
\toprule
 & Cert.\ Corr. & Cert.\ Wr. & Uncert.\ Corr. & Uncert.\ Wr. \\
\midrule
PubMedQA        &  0 & 0 &  73 & 27 \\
RealMedQA       & 48 & 0 & 163 & 12 \\
HotpotQA        &  5 & 1 & 139 & 93 \\
HotpotQA-FW     & 21 & 5 & 124 & 68 \\
2WikiMHQA       &  4 & 0 & 164 & 76 \\
MuSiQue         &  1 & 1 &  36 & 56 \\
\midrule
Pooled          & 79 & 7 & 699 & 332 \\
\bottomrule
\end{tabular}
\end{table}

\subsection{Dense-Side Selective-Prediction Frontier}
\label{sec:appendix_dense_frontier}

The dense-side frontier comparison referenced in
Section~\ref{sec:results_composite} reports the AUREC for four gating
strategies, computed as a pure post-hoc replay from the saved dense-run
uncertainty scalars (no retrieval, generation, or API calls).
The four gates are: \emph{best\_single} (the single signal with the highest
dense-side AUROC on that dataset), \emph{logistic} (a leave-one-fold-out
logistic gate fit on all seven dense uncertainty scalars, standardised and
sanitised against non-finite values), \emph{conjunctive} (the dense-analogue
of the KG audit rule: SD-UQ $\leq$ dataset median \emph{and} SEU $\leq 0.5$;
GPS is KG-only so it is dropped), and \emph{mean\_percentile} (the paper's
composite score applied to the dense run).
AUREC is the area under the risk-excess-coverage curve (lower is better);
the logistic gate's AUROC is reported to expose where it overfits the
low-error biomedical sets.

\begin{table}[t]
\centering
\scriptsize
\caption{%
  \textbf{Dense-side selective-prediction frontier.}
  AUREC (area under risk-excess-coverage; lower is better) for the four gates,
  with base error and the learned gate's out-of-fold AUROC; the lowest AUREC in
  each row is bold.
  ``Best'' is the best single signal per dataset; the winning signal is, in row
  order: SD-UQ, DSE, SD-UQ, VN-Entropy, SEU, VN-Entropy.
  The logistic gate collapses below chance (AUROC $<0.5$) on the two low-error
  biomedical sets and never attains the lowest AUREC on any dataset.%
}
\label{tab:dense_selective_frontier}
\setlength{\tabcolsep}{4pt}
\begin{tabular}{@{}lcccccc@{}}
\toprule
 & Base & \multicolumn{4}{c}{AUREC (lower better)} & Log. \\
\cmidrule(lr){3-6}
Dataset & err. & Best & Log. & Conj. & Mean & AUROC \\
\midrule
PubMedQA        & .250 & \textbf{.057} & .185 & .078 & .193 & .453 \\
RealMedQA       & .050 & .006 & .076 & \textbf{.003} & .018 & .312 \\
HotpotQA        & .345 & \textbf{.085} & .107 & .113 & .118 & .615 \\
HotpotQA-FW     & .279 & \textbf{.047} & .097 & .073 & .071 & .626 \\
2WikiMHQA       & .288 & .084 & .093 & \textbf{.048} & .100 & .628 \\
MuSiQue         & .522 & .194 & .159 & \textbf{.084} & .128 & .684 \\
\bottomrule
\end{tabular}
\end{table}

The logistic gate's below-chance AUROC on RealMedQA ($0.31$) and PubMedQA
($0.45$) is not a numerical artefact: features are standardised and
non-finite values are median-imputed before fitting. It reflects genuine
overfitting in the low-positive-fraction regime ($5\%$ and $25\%$ error),
where a five-fold logistic model with seven features has too few positives
per fold to generalise and inverts the ranking on held-out data. On the
larger, higher-error multi-hop sets (HotpotQA, 2WikiMHQA, MuSiQue) the
logistic gate recovers to $0.61$--$0.68$ AUROC but still does not attain the
lowest AUREC on any dataset (Table~\ref{tab:dense_selective_frontier}); the
conjunctive rule and the best single signal split the six datasets between
them.

\subsection{SEU Domain-NLI Ablation}
\label{sec:appendix_seu_nli_ablation}

One remaining question is whether SEU's neutral plateau ($74\%$ of RealMedQA rows at
the all-neutral $\mathrm{SEU}=0.5$ default) is an artefact of the
general-domain NLI model (\texttt{microsoft/deberta-large-mnli}) rather than
a property of the retrieved evidence.
We test this by recomputing SEU on PubMedQA under two NLI conditions
(Table~\ref{tab:seu_nli_ablation}): (A) the
paper's \texttt{deberta-large-mnli}, and (C) a biomedical-aware LLM-NLI judge
(\texttt{gpt-4o-mini}, prompted as a domain expert; zero-shot, temperature
$0$).
Because the run logs keep chunk identifiers rather than the SEU per-chunk
inputs, retrieval is re-run against the existing KG (no KG rebuild) to recover
the per-question chunks; the saved answer is used as the
SEU hypothesis, matching the production path.
A reproduction check confirms that the re-retrieved chunks reproduce the
paper's SEU: recomputed \texttt{deberta-large} AUROC matches the saved value
to within $0.001$ on both policies, so there is no retrieval drift.

\begin{table}[t]
\centering
\scriptsize
\caption{%
  \textbf{SEU domain-NLI ablation on PubMedQA} ($n{=}100$).
  Condition (A) reproduces the paper; (C) replaces the NLI head with a
  biomedical-aware LLM judge.
  ``Neutral rate'' is the fraction of rows where no chunk entails or
  contradicts the answer; ``Plateau@0.5'' is the fraction sitting at the
  all-neutral $\mathrm{SEU}=0.5$ default.
  The LLM judge improves AUROC on both policies, confirming that part of the
  neutral plateau is a general-domain-NLI artefact, but the neutral rate does
  not fall, so the plateau is not entirely an artefact.%
}
\label{tab:seu_nli_ablation}
\setlength{\tabcolsep}{4pt}
\begin{tabular}{@{}llcccc@{}}
\toprule
Policy & NLI condition & AUROC & Mean SEU & Neutral rate & Plateau@0.5 \\
\midrule
\multirow{2}{*}{KG}
  & (A) deberta-large  & 0.595 & 0.619 & 0.732 & 0.59 \\
  & (C) gpt-4o-mini    & 0.625 & 0.460 & 0.887 & 0.71 \\
\midrule
\multirow{2}{*}{Dense}
  & (A) deberta-large  & 0.551 & 0.591 & 0.788 & 0.62 \\
  & (C) gpt-4o-mini    & 0.641 & 0.467 & 0.902 & 0.72 \\
\bottomrule
\end{tabular}
\end{table}

The LLM judge improves AUROC by $+0.030$ (KG) and $+0.090$ (dense), so part
of the neutral plateau is indeed a general-domain-NLI artefact: the LLM judge
resolves some biomedical paraphrase that \texttt{deberta-large} labels
neutral.
The improvement is modest, however, and the neutral rate paradoxically
\emph{rises} under the LLM judge ($0.73\to 0.89$ on KG), because
\texttt{gpt-4o-mini} is more conservative about committing to entailment on
specialist passages than the MNLI-fine-tuned model.
The plateau reflects two effects: (i) a general-domain-NLI labelling
gap that a domain-aware judge partially closes, and (ii) genuinely
non-committal retrieved evidence that neither entails nor contradicts the
answer.
RealMedQA was not included because its KG was not loaded for the same
re-retrieval here (PubMedQA's was), so the result is PubMedQA-only; the direction is consistent with the RealMedQA plateau being
partly artefactual, but the magnitude on the clinical target domain remains to
be confirmed.

\subsection{Independent Re-Judge on the Certified Subset}
\label{sec:appendix_rejudge_certificate}

The clinical headline (the conjunctive rule's $48/48$ selection on adaptive
RealMedQA) is the cell most exposed to same-model judging, and RealMedQA also
carries the lowest inter-judge $\kappa$ ($0.52$) because near-ceiling accuracy
leaves few wrong answers for $\kappa$ to score (a marginal-imbalance effect, not
broad disagreement: raw agreement is $92.8\%$).
We re-judged the $48$ certified answers with the independent
Llama-3.3-70B judge: \textbf{all $48$ are confirmed correct, with zero label
flips} (\texttt{certificate\_rejudge.json}).
The certified cell is robust to the judge family even though dataset-wide
$\kappa$ is low, because the rule selects high-agreement, near-ceiling-confidence
clinical answers rather than the contested wrong-answer rows that depress
$\kappa$.
More broadly, re-ranking every free-text run under the independent labels leaves
each central contrast intact: on adaptive RealMedQA SD-UQ AUROC is $0.86$ under
the independent judge (versus $0.81$ originally), while on the strict clinical
run SD-UQ stays collapsed ($0.31$) and SEU and GPS continue to rank the errors
($0.71$ and $0.67$); the lock-in signature is not an artefact of
same-model self-preference, which would have inflated answer-state agreement, not
the evidence- and retrieval-state signals.

\FloatBarrier
\section{Code and Data Availability}
\label{sec:appendix_code}

The system implementation, experiment harness, and scripts that generate every
table and figure are available at
\url{https://github.com/julka01/OntoGraphRAG}, with the run artefacts in the same
repository.
Each run log is a per-query JSON record storing the verbatim model responses for
all $N{=}5$ uncertainty samples (and the single accuracy-generation response),
retrieved chunk identifiers and graph paths, route labels, both judges'
correctness labels, the raw per-query metric scores and uncertainty estimates,
and the fixed sampled question IDs for each snapshot.
The GPS replay stores for the post-hoc recomputations are also released (the
RealMedQA replay, the $\tau\times\gamma$ sensitivity sweep, and the
HotpotQA-FullWiki diversity run).
The release bundle is \texttt{reproducibility/arxiv\_v1/}, whose
\texttt{MANIFEST.json} records the exact repository commit and file hashes for
the cached logs.
Every fixed-subset number and post-hoc replay is computed from these artefacts,
so the results reproduce without rerunning generation or retrieval; the
verification bundle replays cached JSONL logs (metrics, confidence intervals,
audit-rule cells) but excludes the main experiment runner, so end-to-end
regeneration of the per-query logs requires cloning the full repository and
incurring API costs.
None of these tiers stores the full retrieved chunk \emph{text}: the run logs
keep chunk identifiers, the verification bundle holds cached metric inputs, and
the chunk text itself lives in the persistent dataset KG. The few analyses that
need chunk text (the SEU domain-NLI ablation,
Appendix~\ref{sec:appendix_seu_nli_ablation}, and confirmation of deep presence
lock-in) re-retrieve it from the KG.

\paragraph{Reproducibility checklist.}
Table~\ref{tab:repro_checklist} consolidates the exact configuration values
scattered through this appendix.

\begin{table*}[t]
\centering
\small
\caption{Reproducibility checklist (verified against the released run
artefacts and code).
\label{tab:repro_checklist}}
\setlength{\tabcolsep}{5pt}
\begin{tabular}{@{}lp{9.2cm}@{}}
\toprule
Item & Value \\
\midrule
Generation model & \texttt{gpt-4o-mini} via OpenRouter (\texttt{openai/gpt-4o-mini-2024-07-18}; OpenAI snapshot 2024-07-18, accessed via OpenRouter; all runs use this fixed snapshot); $T{=}1.0$ for the $N{=}5$ uncertainty samples, $T{=}0.0$ for accuracy generation and KG extraction \\
Judge model (main) & same \texttt{gpt-4o-mini} backbone as generation \\
Judge model (re-judge) & \texttt{meta-llama/llama-3.3-70b-instruct} via OpenRouter, $T{=}0.0$ \\
Embedding model & \texttt{all-MiniLM-L6-v2} (384-d, Sentence Transformers), shared by chunks, entities, and response embeddings \\
NLI models & \texttt{microsoft/deberta-large-mnli} (DSE / P(True) clustering, SEU); \texttt{roberta-large-mnli} (SelfCheckGPT pairwise) \\
Subset seeds & $42$ for every dataset and run (including dose-response and $N{=}20$ probes) \\
Sampling budget & $N{=}5$ headline; $N{=}20$ resampling probe \\
GPS calibration and sensitivity & $\tau{=}0.60$, $\gamma{=}0.40$ selected on the RealMedQA replay and applied frozen elsewhere; Appendix~\ref{sec:appendix_gps_sensitivity} replays $\tau \in \{0.50,\ldots,0.70\}$ and $\gamma \in \{0.2,\ldots,1.0\}$ \\
Cached reproduction & headline tables and figures regenerate from released saved artefacts; no API calls, KG rebuild, or generation rerun is required for the reported analyses \\
Retrieval temperature & $0.0$ in all reported runs; $\{0, 0.5, 1.0\}$ in the dose-response arms (shortlist factor $4$) \\
Provider failures & per-run generation failures excluded from clean accuracy: KG side $0$--$32$ and dense side $0$--$42$ per run (Table~\ref{tab:generation_failures}) \\
Code version & \texttt{OntoGraphRAG} v1.0.0, public repository; exact commit and artefact hashes recorded in the released manifest \\
\bottomrule
\end{tabular}
\end{table*}

\paragraph{Mechanism-probe artefacts.}
The additional mechanism analyses reported in
Section~\ref{sec:results_collapse} have their own artefacts, released
alongside the run logs: the retrieval-temperature dose-response runs
(RealMedQA, strict profile, $n=100$, seed 42, retrieval temperatures
$\{0, 0.5, 1.0\}$, with per-question chunk overlap retained), the
$N{=}20$ resampling probe on the same strict subset, the independent
re-judge of all free-text runs
(\texttt{rejudge\_independent.json}, \texttt{rejudge\_wave1.json}), the
verbalised-confidence baseline (\texttt{verbalized\_confidence.json}),
the audit-rule evaluation with matched-coverage baselines and split-half
validation (\texttt{certificate\_eval.json}), the silent-rate sensitivity
and route-decomposition numbers (\texttt{wave1\_analysis.json}), the
reporting-robustness bookkeeping for failure counts and the audit rule
odds ratio (\texttt{reporting\_robustness.json}), and the
KG scale statistics queried from the persistent Neo4j stores
(\texttt{kg\_scale\_stats.json}).

\end{document}